\documentclass{article}
\usepackage{caption}

\usepackage[preprint]{neurips_2025}

\usepackage[utf8]{inputenc} 
\usepackage[T1]{fontenc}    
\usepackage{hyperref}       
\usepackage{url}            
\usepackage{booktabs}       
\usepackage{amsfonts}       
\usepackage{nicefrac}       
\usepackage{microtype}      

\usepackage{microtype}
\usepackage{multirow}
\usepackage{multicol}
\usepackage{graphicx}
\usepackage{subfigure}
\usepackage{booktabs} 
\usepackage{algorithmic}
\usepackage{algorithm}
\usepackage{lipsum}

\usepackage[table, svgnames, dvipsnames]{xcolor}
\usepackage{makecell, cellspace, caption}
\usepackage{amsmath}
\usepackage{amssymb}
\usepackage{mathtools}
\usepackage{amsthm}

 \usepackage{tabularray}
\theoremstyle{plain}
\newtheorem{theorem}{Theorem}[section]
\newtheorem{proposition}[theorem]{Proposition}

\theoremstyle{definition}

\theoremstyle{remark}

\newcommand{\cB}{{\mathcal{B}}}

\newcommand{\cD}{{\mathcal{B}}}

\newcommand{\bbE}{\mathbb{E}}

\newcommand{\Mix}{\text{\tiny\textsc{Mix}}}

\newcommand{\KL}{KL}
\def\cmt#1{{\color{blue}{#1}}}

\newcommand{\showImage}[7][0.25]{
\begin{minipage}{#1\textwidth}
    \caption*{\detokenize{#7}}
    \captionsetup{justification=centering}
    \vskip -0.05in
    \centering
    \includegraphics[width=1.0\textwidth,trim={#2 #3 #4 #5},clip]{#6}
\end{minipage}
}

\newcommand{\showinstance}[4][0.25]{
\begin{minipage}{#1\textwidth}
    \centering
    \includegraphics[width=1.0\textwidth,trim={#2 0 0 0},clip]{#3}
    \ifx #4\empty \else \parbox{\linewidth}{\centering #4} \fi
    \captionsetup{justification=centering}
\end{minipage}
}

\usepackage{wrapfig}

\newcommand{\showLegend}[6][0.25]{
\begin{minipage}{#1\textwidth}
    \centering
    \includegraphics[width=1.0\textwidth,trim={#2 #3 #4 #5},clip]{#6}
\end{minipage}
}

\newcommand{\highlight}[1]{\textcolor{RoyalBlue}{\textbf{#1}}}

\title{Learning What to Do and What Not To Do: Offline Imitation from Expert and Undesirable Demonstrations}

\author{%
    Huy Hoang \\
  Singapore Management University\\
  \texttt{mh.hoang.2024@phdcs.smu.edu.sg} \\
  \And 
    Tien Mai \\
  Singapore Management University\\
  \texttt{atmai@smu.edu.sg} \\
  \And 
    Pradeep Varakantham \\
  Singapore Management University\\
  \texttt{pradeepv@smu.edu.sg} \\
  \And 
  Tanvi Verma \\
  Institute of High Performance Computing \\
  Agency for Science, Technology and Research, Singapore \\
  \texttt{Tanvi$\_$Verma@ihpc.a-star.edu.sg}
}

\begin{document}

\maketitle

\begin{abstract}
Offline imitation learning typically learns from expert and unlabeled demonstrations, yet often overlooks the valuable signal in explicitly undesirable behaviors. In this work, we study offline imitation learning from contrasting behaviors, where the dataset contains both expert and undesirable demonstrations. We propose a novel formulation that optimizes a difference of KL divergences over the state-action visitation distributions of expert and undesirable (or bad) data. Although the resulting objective is a DC (Difference-of-Convex) program, we prove that it becomes \textit{convex} when expert demonstrations outweigh  undesirable demonstrations, enabling a practical and stable non-adversarial training objective. Our method avoids adversarial training and handles both positive and negative demonstrations in a unified framework. Extensive experiments on standard offline imitation learning benchmarks demonstrate that our approach consistently outperforms state-of-the-art baselines.


\end{abstract}

\section{Introduction}

Imitation learning~\cite{garg2021iq,kim2021demodice,li2023imitation,hoang2024imitate,xu2022discriminator} offers a compelling alternative to Reinforcement Learning (RL)~\cite{Sutton1998,puterman2014markov,mnih2015human} by enabling agents to learn directly from expert demonstrations without the need for explicit reward signals. This paradigm has been successfully applied in various domains, even with limited expert data, and is particularly effective in capturing complex human behaviors and preferences.

Traditional imitation learning typically assumes access to high-quality expert demonstrations, which can be expensive and difficult to obtain~\cite{ross2011reduction,torabi2018behavioral,Zhu2020The}. In practice, datasets often contain a mixture of expert and sub-optimal demonstrations. Recent advances in imitation learning have begun to address this more realistic setting, aiming to develop algorithms that can leverage informative signals from both expert and non-expert data~\cite{brown2019extrapolating,myers2022learning,hoang2024imitate}.

While existing imitation learning approaches in the mixed-quality setting typically assume that mixed-quality demonstrations are not drastically different from expert behavior, they often frame learning as mimicking both expert and  sub-optimal trajectories—albeit with different weights~\cite{kim2021demodice,kim2022lobsdice,xu2022discriminator}. However, in practice, mixed-quality data may contain poor or undesirable demonstrations that the agent should explicitly avoid. For example, in autonomous driving, undesirable demonstrations may include unsafe lane changes or traffic violations, which should not be imitated under any circumstances. Another example can be found in healthcare applications, where undesirable demonstrations may correspond to incorrect diagnoses or unsafe treatment plans that could harm patients if imitated.  Existing imitation learning approaches are limited in their ability to handle contrasting demonstrations. Most methods are either not explicitly designed to avoid undesirable behaviors, or are ill-equipped to deal with scenarios where both expert and undesirable demonstrations coexist within the dataset~\cite{wu2019imitation,zhang2021confidence,hoang2024imitate}. It is important to note that learning by mimicking expert or mildly sup-optimal demonstrations is often tractable, as the corresponding objective—typically framed as divergence minimization—is convex~\cite{kim2021demodice,kim2022lobsdice}. However, incorporating objectives that explicitly avoid bad (or undesirable) demonstrations can introduce non-convexities, making the optimization significantly more challenging. In this paper, we propose a unified framework that addresses these challenges, aiming to bridge this gap in the current imitation learning literature.

Specifically, we focus on the setting of \textit{offline imitation learning}, where interaction with the environment is not available, and assume that the dataset contains both \textit{expert} and \textit{undesirable} demonstrations. We make the following contributions:
\begin{itemize}
    \item We formulate the learning problem with the goal of matching expert behavior while explicitly avoiding undesirable demonstrations. Although the resulting training objective is expressed as the difference between two KL divergences (and is therefore difference-convex), we prove that it becomes \textit{convex} when the expert component outweighs the undesirable one. This convexity is critical, as it enables us to reformulate the learning problem over the state-action visitation distribution as an more tractable unconstrained optimization via Lagrangian duality. Our objective stands in contrast to most existing distribution-matching imitation learning approaches, which typically rely solely on divergence minimization and naturally yield convex objectives. By introducing a divergence maximization term to account for undesirable behavior, we demonstrate that the overall objective \textit{remains convex and manageable.}
    \item We further enhance the learning objective by proposing a surrogate objective that lower-bounds the original one, offering the advantage of a non-adversarial and convex optimization problem in the Q-function space. In addition, we introduce a novel Q-weighted behavior cloning (BC) approach, supported by theoretical guarantees, for efficient policy extraction.
    \item Extensive experiments on standard imitation learning benchmarks show that our method consistently outperforms existing approaches, both in conventional settings where datasets contain expert and unlabeled demonstrations, and in more realistic scenarios where explicitly undesirable demonstrations are included.
\end{itemize}

\section{Related Works}
\noindent \textbf{Imitation Learning.} Imitation learning trains agents to mimic expert behavior from demonstrations, with Behavioral Cloning (BC) serving as a foundational method by maximizing the likelihood of expert actions. However, BC often suffers from distributional shift~\cite{ross2011reduction}. Recent work addresses this issue by leveraging the strong generalization capabilities of generative models~\cite{zhao2023learning,chi2023diffusion}. Inspired by GANs~\cite{goodfellow2014generative}, methods like GAIL~\cite{ho2016generative} and AIRL~\cite{fu2018learning} use a discriminator to align the learner’s policy with the expert’s, while SQIL~\cite{reddy2019sqil} simplifies reward assignment by distinguishing expert and non-expert behaviors. Although effective, these approaches typically require online interaction, which may be impractical in many real-world scenarios.

To address this, offline methods such as AlgaeDICE~\cite{nachum2019algaedice} and ValueDICE~\cite{kostrikovimitation} employ Stationary Distribution Correction Estimation (DICE), though they often encounter stability issues. Building on ValueDICE, O-NAIL~\cite{arenz2020non} avoids adversarial training, enabling stable offline imitation. More recently, several approaches have extended the DICE framework with stronger theoretical foundations and improved empirical performance~\cite{lee2021optidice,mao2024odice}. In parallel, IQ-Learn~\cite{garg2021iq} has emerged as a unified framework for both online and offline imitation learning, inspiring a range of follow-up works~\cite{alhafez2023,hoang2024sprinql}. However, all these approaches rely on the presence of many expert demonstrations, which may not always be available. 

\noindent\textbf{Offline imitation learning from suboptimal demonstrations:} Several approaches have been developed to tackle the challenges of offline imitation learning from suboptimal data, which is common in real-world scenarios. A notable direction involves preference-based methods, where algorithms infer reward functions by leveraging ranked or pairwise-compared trajectories to guide learning~\cite{kimpreference, kang2023beyond, hejna2024inverse}. Recent works, such as SPRINQL~\cite{hoang2024imitate}, take advantage of demonstrations that exhibit varying levels of suboptimality, enabling the learner to better generalize beyond near-optimal behaviors. Another important line of research explores the use of unlabeled demonstrations in conjunction with a limited number of expert trajectories. Techniques like DemoDICE~\cite{kim2021demodice}, SMODICE~\cite{ma2022versatile}, and ReCOIL~\cite{sikchi2024dual} apply Distribution Correction Estimation (DICE)~\cite{sunehag2017value,lee2021optidice,mao2024odice} to re-weight trajectories and align the state or state-action distributions with those of the expert. In parallel, classifier-based methods, such as DWBC~\cite{xu2022discriminator}, ISW-BC~\cite{li2023imitation}, and ILID~\cite{yue2024how}, use discriminators to distinguish expert-like behaviors within mixed-quality data and assign them greater importance. Collectively, these strategies aim to enhance policy robustness and performance in offline settings where high-quality expert data is scarce or expensive to obtain. However, all of these approaches are primarily focused on imitating and are unable to avoid undesirable or bad demonstrations, which is crucial in domains such as self driving where there are many unsafe behaviors that would need to be avoided. 

SafeDICE~\cite{jang2024safedice} was introduced to address this problem of avoiding undesirable or bad demonstrations. However, SafeDICE is not designed to handle scenarios where both expert and undesirable datasets are available. Moreover, their approach still relies on minimizing a divergence between the learning policy and a mixture of unlabeled and undesirable data—an approach that is vulnerable to the quality of the unlabeled dataset and may degrade when such data is of low quality.

In this paper, we aim to optimize on the principle of "Imitate the Good and Avoid the Bad", which has recently gained attention in reference and safe reinforcement learning~\cite{abdolmaleki2025learning, hoang2024imitate,gong2025offline} and large language model training~\cite{lu2025semantic}. We extend this idea to the offline imitation setting by proposing a novel and efficient method that learns from expert demonstrations while avoiding undesirable ones. To  our knowledge, this is the first offline imitation learning approach to efficiently learn policies by jointly utilizing both expert and undesirable demonstrations.

\section{{Preliminaries}}\label{sec:background}

\paragraph{Markov Decision Process (MDP).} We consider a MDP defined by the following tuple  $\mathcal{M} = \left\langle S, A, r, P, \gamma, s_0 \right\rangle$, where  $S$ denotes the set of states, $s_0$ represents the initial state set, $A$ is the set of actions, $r: S \times A \rightarrow \mathbb{R}$ defines the reward function for each state-action pair, and $P: S \times A \rightarrow S$ is the transition function, i.e., $P(s'|s,a)$ is the probability of reaching state $s'\in S$ when action $a\in A$ is made at state $s\in S$,  and $\gamma$ is the discount factor. In reinforcement  learning (RL), the aim is to find a policy that maximizes the expected long-term accumulated reward:
$\max_{\pi} \left\{\bbE_{(s,a)\sim d^\pi}[r(s,a)]\right\}$, where $d^\pi$ is the occupancy measure (or state-action visitation distribution) of  policy $\pi$:
$d^\pi(s,a) = (1-\gamma)\pi(a|s) \sum_{t=1}^\infty \gamma^t P(s_t = s|\pi).$

\paragraph{Offline Imitation Leaning.}
Recent imitation learning (IL) approaches have adopted a distribution-matching formulation, where the objective is to minimize the divergence between the occupancy measures (i.e., state-action visitation distributions) of the learning policy and the expert policy:
$\min_{d^{\pi}} \left\{ D_f\left(d^{\pi} \,\|\, d^E\right) \right\},$ where $D_f$ denotes an $f$-divergence between the occupancy distributions $d^{\pi}$ (induced by the learning policy $\pi$) and $d^E$ (induced by the expert policy). In particular, when the Kullback–Leibler (KL) divergence is used, the learning objective becomes:
$\min_{d^{\pi}} \; \mathbb{E}_{(s,a) \sim d^{\pi}} \left[ \log\left( \frac{d^{\pi}(s,a)}{d^E(s,a)} \right) \right].$
In the space of state-action visitation distributions ($d^\pi$), the training can be formulated as a convex constrained optimization problem. To enable efficient training, Lagrangian duality is typically employed to recast the problem into an unconstrained form~\cite{lee2021optidice,kim2021demodice}. 

\paragraph{Offline IL with unlabeled data.} In offline imitation learning with unlabeled data, it is typically assumed that a limited set of expert demonstrations $\cD^E$ is available, along with a larger set of unlabeled demonstrations $\cD^{\Mix}$. Distribution-matching approaches have been widely adopted to handle this setting. Prior methods often formulate the objective as a weighted sum of divergences between the learning policy and both expert and unlabeled data:
$
\min_{d^{\pi}} \left\{ D_f\left(d^{\pi} \,\|\, d^E\right) + \alpha D_f\left(d^{\pi} \,\|\, d^{\Mix}\right) \right\}, \text{ where } \alpha \geq 0.$
Other approaches construct mixtures of occupancy distributions, such as $d^{\pi,\Mix} = \alpha d^{\pi} + (1 - \alpha) d^{\Mix} \text{ and } d^{E,\Mix} = \alpha d^E + (1 - \alpha) d^{\Mix},$ and minimize the divergence between $d^{\pi,\Mix}$ and $d^{E,\Mix}$~\cite{kim2021demodice,kim2022lobsdice,ma2022versatile,sikchi2024dual}. In most existing approaches along this line of research, the convexity of the objective with respect to $d^\pi$ has been heavily leveraged to derive tractable learning objectives. However, when a divergence \emph{maximization} term is introduced—as in our approach—this convexity may no longer hold, rendering many existing methods inapplicable.

\section{ContraDICE: Offline Imitation Learning from Contrasting Behaviors}
We begin by introducing a novel learning objective based on the difference between two KL divergences. Leveraging the convexity of this formulation, we derive a tractable and unconstrained optimization problem. Given that the resulting objective includes exponential terms that may lead to numerical instability, we enhance this by proposing a lower-bound approximation. This approximation enables us to reformulate the learning process as a more tractable, non-adversarial Q-learning objective, which remains convex in the space of Q-functions.

\subsection{Dual KL-Based Formulation}
{Assume that we have access to three sets of demonstrations: good dataset $\cD^G$ contains \textit{good} or \textit{expert} demonstrations, bad dataset $\cD^B$ contains \textit{bad} or \textit{undesirable} demonstrations that the agent should avoid, and the unlabeled dataset $\cD^{\Mix}$ is a large set of unlabeled demonstrations used to support offline training. We consider the realistic scenario where the identified datasets $\cD^G$ and $\cD^B$ are limited in size, while $\cD^{\Mix}$ is significantly larger—an assumption that aligns with typical settings in offline imitation learning from unlabeled demonstrations.}

Let $d^\pi(s, a)$, $d^G(s, a)$, and $d^B(s, a)$ denote the state-action visitation distributions induced by the learned policy $\pi$, the good policy, and the bad policy, respectively. Following the DICE framework~\cite{nachum2019algaedice,kostrikovimitation}, we propose to optimize the following training objective:
\begin{align}
\min_{d^\pi} \quad f(d^{\pi})  = D_{\text{KL}}(d^\pi \,\|\, d^G) - \alpha \, D_{\text{KL}}(d^\pi \,\|\, d^B),\label{eq:main-obj}
\end{align}
where $\alpha > 0$ is a tunable hyperparameter. The goal of this objective is twofold: (1) to minimize the divergence between the learned policy and the good policy, and (2) to \emph{maximize} the divergence from the bad policy, thereby avoiding undesirable behavior.

This formulation differs from all existing DICE-based approaches in the literature, which primarily focus on minimizing KL divergence—even when dealing with undesirable or unsafe demonstrations. By contrast, our approach introduces a principled mechanism to explicitly repel the learned policy from undesirable behavior while still aligning it with good data.

While the presence of a KL divergence maximization term in the objective may raise concerns about the convexity of the training problem, we observe that the objective in \eqref{eq:main-obj} takes the form of a difference between two convex functions. This is, in general,  not convex and can be challenging to optimize. Fortunately, we show that under a mild condition, the overall objective remains convex. Specifically, if the weight on the bad policy divergence term is smaller than that on the good policy (i.e., $\alpha < 1$), then the objective becomes convex in $d^\pi$.

\begin{proposition}\label{prop:convex}
If $\alpha \leq 1$, then the objective function
$f(d^\pi) = D_{\text{KL}}(d^\pi \,\|\, d^G) - \alpha \, D_{\text{KL}}(d^\pi \,\|\, d^B)$
is convex in $d^\pi$.
\end{proposition}
Convexity is essential in most DICE-based frameworks, as it enables the use of Lagrangian duality to construct well-behaved and tractable training objectives. Our goal is to develop a Q-learning method that recovers a policy minimizing the objective in \eqref{eq:main-obj}. To this end, we formulate the problem as the following constrained optimization:
\begin{align}
\min_{d, \pi} \quad & f(d, \pi) = D_{\text{KL}}(d \,\|\, d^G) - \alpha \, D_{\text{KL}}(d \,\|\, d^B)\label{eq:obj-f-d-pi}  \\
\text{s.t.} \quad & d(s, a) = (1 - \gamma)p_0(s)\pi(a \mid s) + \gamma \pi(a \mid s) \sum_{s', a'} d(s', a') {T}(s \mid s', a'),\nonumber
\end{align}
where $d(s, a)$ is the state-action visitation distribution, and ${T}$ is the environment transition function.
{Let $\cD^U=\cD^G\cup\cD^{\Mix}$ denote the union dataset}, and let $d^U$ be the state-action visitation distribution derived from it. The following proposition gives an another formulation for the objective in \eqref{eq:main-obj}:
\begin{proposition}\label{prop:new-form}
    The objective function in \eqref{eq:obj-f-d-pi} can be written as: $f(d, \pi) =   (1-\alpha)D_{\KL}(d||d^U) - \mathbb{E}_{(s, a) \sim d} \left[ \Psi(s, a) \right]$, where $\Psi(s, a) = \log \frac{d^G(s, a)}{d^U(s, a)} - \alpha \log \frac{d^B(s, a)}{d^U(s, a)}.$
\end{proposition}
This formulation introduces a KL-based regularization centered on the reference distribution $d^U$, with $\Psi(s, a)$ acting as a correction term that incorporates information from the labeled good and bad demonstrations. The reformulated objective in Proposition \ref{prop:new-form} further confirms that the function $f(d, \pi)$ remains convex in $d$ when $\alpha \leq 1$. Here we note that, under the same condition $\alpha \leq 1$, convexity may not hold for other $f$-divergences (a detailed discussion is provided in the appendix).

Given the convexity of the objective in \eqref{eq:main-obj}, we can equivalently move the constraints into the objective using Lagrangian duality, leading to the following Q-learning formulation (details of the derivation are given in the appendix):
\begin{align*}
\max_{\pi} \min_{Q} \Big\{ 
&(1 - \gamma) \, \mathbb{E}_{(s, a) \sim p_0, \pi} \left[ Q(s, a) \right] \\
&+ (1-\alpha) \mathbb{E}_{(s, a) \sim d^U} \left[ \exp\left( \frac{\Psi(s, a)+ \gamma \, \mathbb{E}_{(s', a') \sim {T}, \pi} [Q(s', a')] - Q(s, a) }{1 - \alpha} \right) \right] 
\Big\}
\end{align*}
To further enhance the efficiency of Q-learning, we adopt the well-known Maximum Entropy (MaxEnt) reinforcement learning framework by incorporating an entropy term into the training objective~\cite{garg2021iq,haarnoja2018soft}. This leads to the following objective:
\begin{align}
  &  L(Q, \pi) = 
(1 - \gamma) \, \mathbb{E}_{(s, a) \sim p_0, \pi} \left[ Q(s, a) - \beta \log \frac{\pi(a \mid s)}{\mu^U(a|s)}  \right]
\nonumber\\
&+ (1-\alpha) \mathbb{E}_{(s, a) \sim d^U} \left[ \exp\left( \frac{\Psi(s, a) + \gamma \, \mathbb{E}_{(s', a') \sim {T}, \pi} \left[ Q(s', a') - \beta \log \frac{\pi(a' \mid s')}{\mu^U(a'|s')}  \right] - Q(s, a) }{1 - \alpha}\right) \right].\nonumber
\end{align}
where $\mu^U(a|s)$ is the behavior policy representing the union dataset $\cD^U$. We now define the soft value function and the soft Bellman operator  as follows:
\[
V_Q^\pi(s) = \mathbb{E}_{a \sim \pi(\cdot \mid s)} \left[ Q(s, a) -\beta \log \frac{\pi(a \mid s)}{\mu^U(a|s)} \right], \quad
\mathcal{T}^\pi[Q](s, a) = Q(s, a) - \gamma \, \mathbb{E}_{s' \sim \mathcal{T}(\cdot \mid s, a)} \left[ V_Q^\pi(s') \right].
\]
Using these definitions, the training objective can be rewritten as:
\begin{equation}\label{eq:main-obj-Q-pi}
    L(Q, \pi) = (1 - \gamma) \, \mathbb{E}_{s \sim p_0} \left[ V_Q^\pi(s) \right]
+ (1-\alpha)\mathbb{E}_{(s, a) \sim d^U} \left[ \exp\left( \frac{\Psi(s, a) - \mathcal{T}^\pi[Q](s, a)}{1 - \alpha} \right) \right].
\end{equation}
This formulation shares structural similarities with IQ-Learn, where \( \mathcal{T}^\pi[Q](s, a) \) is referred to as the \emph{inverse Bellman operator} and is often interpreted as a reward function expressed in terms of the Q-function itself.

\paragraph{Remark.} The objective in Equation~\eqref{eq:main-obj-Q-pi} is valid only when \( \alpha < 1 \). In the special case where \( \alpha = 1 \), i.e., when the bad demonstrations are weighted equally to the expert demonstrations—the training objective simplifies significantly. According to Proposition \ref{prop:new-form}, the training objective reduces to a standard offline RL problem with reward function \( \Psi(s, a) \): 
$\max_{d} \; \mathbb{E}_{(s, a) \sim d} \left[ \Psi(s, a) \right] = \max \bbE\left[\sum_{t=0}^\infty \gamma^t \Psi(s_t,a_t)\right].$

\subsection{Tractable Lower Bounded Objective}
In this section, we propose an additional step to improve the stability and tractability of the learning objective introduced above. We first observe that the exponential term in Equation~\eqref{eq:main-obj-Q-pi} may lead to instability during training. To address this issue, we propose to approximate the exponential using a linear lower bound, which not only improves stability but also preserves a similar optimization objective.
\begin{proposition}\label{prop:4.3}
Let the surrogate objective be defined as:
\begin{align}
    \widetilde{L}(Q, \pi) &= (1 - \gamma) \, \mathbb{E}_{s \sim p_0} \left[ V_Q^\pi(s) \right]
- \mathbb{E}_{d^U} \left[ \delta(s,a) \mathcal{T}^\pi[Q](s, a) \right] +(1-\alpha)\mathbb{E}_{d^U} \left[ \delta(s,a) \right].\label{eq:obj-tL-Q-pi}
\end{align}
where $ \delta(s,a)=  \exp\left( \frac{\Psi(s, a)}{1 - \alpha} \right).$
Then \( \widetilde{L}(Q, \pi) \) is a lower bound of \( L(Q, \pi) \), with equality when \( \mathcal{T}^\pi[Q](s, a) = 0 \) for all \( (s, a) \).
\end{proposition}

The lower-bound approximation \( \widetilde{L}(Q, \pi) \) offers several benefits. First, as a valid lower bound of \( L(Q, \pi) \), maximizing \( \widetilde{L}(Q, \pi) \) promotes the original objective. Second, its structure—linear in \( Q \) and concave in \( \pi \)—leads to a simplified, non-adversarial training procedure (see Proposition~\ref{prop:L(Q-pi)}). Finally, its optimization goals remain aligned with those of \( L(Q, \pi) \), encouraging high expected soft value under the initial state distribution and consistency between the soft Bellman residual and the guidance signal \( \Psi(s,a) \).

\paragraph{Remark.} We note that the training objective in Equation~\eqref{eq:obj-tL-Q-pi} generalizes the IQ-Learn objective~\cite{garg2021iq} as a special case. In particular, \( \widetilde{L}(Q, \pi) \) reduces exactly to the IQ-Learn objective when \( \alpha = 0 \) (i.e., the undesirable dataset is ignored) and \( \cD^G \equiv \cD^U \) (i.e., the good dataset coincides with the union dataset). 
To see this, observe that when \( \alpha = 0 \) and \( d^G = d^U \), the  term \( \Psi(s, a) \) becomes zero for all \( (s, a) \). As a result, the surrogate objective simplifies to:
$\widetilde{L}(Q, \pi) = (1 - \gamma) \, \mathbb{E}_{s \sim p_0} \left[ V_Q^\pi(s) \right]
- \mathbb{E}_{(s, a) \sim d^G} \left[ \mathcal{T}^\pi[Q](s, a) \right],$
which is exactly the training objective proposed in IQ-Learn. Thus, our formulation can be viewed as a principled extension of IQ-Learn that explicitly accounts for and contrasts between good and bad behaviors.

We now present several key properties of the training objective \( \widetilde{L}(Q, \pi) \) that make it particularly convenient and tractable for use, as formalized in Proposition~\ref{prop:L(Q-pi)} below.

\begin{proposition}\label{prop:L(Q-pi)}
The following properties hold:
\begin{itemize}
    \item[(i)] \( \widetilde{L}(Q, \pi) \) is linear in \( Q \) and concave in \( \pi \). As a result, the max–min optimization can be equivalently reformulated as a min–max problem:
   $ \max_{\pi} \min_{Q} \widetilde{L}(Q, \pi) = \min_{Q} \max_{\pi} \widetilde{L}(Q, \pi).$
    \item[(ii)] The min–max problem \( \min_{Q} \max_{\pi} \widetilde{L}(Q, \pi) \) reduces to the following non-adversarial problem:
    \[
    \min_{Q} \left\{ \widetilde{L}(Q) = 
    (1 - \gamma) \, \mathbb{E}_{s \sim p_0} \left[ V_Q(s) \right]
    - \mathbb{E}_{(s, a) \sim d^U} \left[\exp\left( \frac{\Psi(s, a)}{1 - \alpha}\right) \mathcal{T}[Q](s, a) \right]
    \right\},
    \]
    where the soft value function \( V_Q(s) \) is defined as:
  $  V_Q(s) = \beta\log \left( \sum_{a}\mu^U(a|s) \exp(Q(s, a)/\beta) \right),$
    and the soft Bellman residual operator is given by:
  $  \mathcal{T}[Q](s, a) = Q(s, a) - \gamma V_Q(s).$ Moreover $\widetilde{L}(Q)$ is convex in Q.
\end{itemize}
\end{proposition}

 \section{Practical Algorithm}
 \label{sec:practial_algo} 
\paragraph{Estimating Occupancy Ratios.} 
The training objective involves several ratios between state-action visitation distributions, which are not directly observable. These quantities can be estimated by solving corresponding discriminator problems.
Specifically, to estimate the ratio \( \frac{d^G(s, a)}{d^U(s, a)} \), we train a binary classifier \( c^G : \mathcal{S} \times \mathcal{A} \rightarrow [0, 1] \) by solving the following standard logistic regression objective:
\begin{align}
\max_{c^G} \left\{ 
\mathbb{E}_{(s, a) \sim d^G} \left[ \log c^G(s, a) \right] 
+ \mathbb{E}_{(s, a) \sim d^U} \left[ \log (1 - c^G(s, a)) \right] 
\right\}.
\label{eq:disc_objective}
\end{align}
Let $c^{G*}(s, a)$ be optimal solution to this problem, then the ratio can be computed as:
$\frac{d^G(s, a)}{d^U(s, a)} = \frac{c^{G*}(s, a)}{1 - c^{G*}(s, a)}.$
Similar discriminators can be trained to estimate other ratios such as \( \frac{d^B(s, a)}{d^U(s, a)} \).

\paragraph{Implicit \texorpdfstring{$V$}{V}-Update and Regularizers.} 
In the surrogate objective $\widetilde{L}(Q)$, the value function $V_Q$ is typically computed via a log-sum-exp over $Q$, which becomes intractable in large or continuous action spaces. To address this, we adopt Extreme Q-Learning (XQL)~\citep{garg2023extreme}, which avoids the log-sum-exp by introducing an auxiliary optimization over $V$, jointly updated with $Q$.
Specifically, $V$ is optimized using the \emph{Extreme-V} objective:
$J(V \mid Q) = \mathbb{E}_{(s,a)\sim d^U} \left[ e^{t(s,a)} - t(s,a) - 1 \right], \quad \text{where} \quad t(s,a) = \frac{Q(s,a) - V(s)}{\beta}.$
The main training objective with fixed $V$ is:
\begin{align}
\widetilde{L}(Q \mid V) &=
    (1 - \gamma) \, \mathbb{E}_{s \sim p_0} \left[ V(s) \right]
    - \mathbb{E}_{(s, a) \sim d^U} \left[\exp\left( \frac{\Psi(s, a)}{1 - \alpha}\right) \left(Q(s,a) - \gamma \mathbb{E}_{s'}[V(s')]\right) \right].
    \label{eq:L(Q|V)}
\end{align}
The overall optimization proceeds by alternating: (i) updating $Q$ via minimizing $\widetilde{L}(Q \mid V)$, and (ii) updating $V$ via minimizing $J(V \mid Q)$. Both sub-problems are convex, enabling efficient and stable training. To further enhance stability, we follow~\citep{garg2021iq,garg2023extreme} and add a convex regularizer $\phi(\mathcal{T}[Q](s,a))$ to prevent reward divergence. We use the $\chi^2$-divergence, $\phi(t) = t^2/2$, a common choice in Q-learning.

\paragraph{Policy Extraction}

Once the $Q$ and $V$ functions are obtained, a common approach for expert policy extraction is to apply advantage-weighted behavior cloning (AW-BC)~\cite{kostrikov2021offline,garg2023extreme,hejna2024inverse,sikchi2024dual}:
\begin{equation}\label{eq:A-weighted-BC}
    \max_{\pi} \sum_{(s,a)\sim \cD^U} \exp\left(\frac{1}{\beta} \left(Q(s,a) - V(s)\right)\right) \log \pi(a \mid s).
\end{equation}

A key limitation of this formulation is that the value function $V(s)$ is only an approximate estimate from the Extreme-V objective, potentially introducing noise and bias into advantage computation and degrading policy quality. To address this, we propose a $Q$-only alternative that avoids reliance on $V(s)$. The following proposition shows that this $Q$-based objective can, in theory, recover the same optimal policy as the original advantage-weighted BC formulation.
\begin{proposition}\label{prop:Q-learning}
The following $Q$-weighted behavior cloning (BC) objective yields the same optimal policy as the original advantage-weighted BC formulation in~\eqref{eq:A-weighted-BC}:
\begin{equation}
     \max_{\pi} \sum_{(s,a)\sim \cD^U} \exp\left(\frac{1}{\beta} Q(s,a)\right) \log \pi(a \mid s).
     \label{eq:qwbc}
\end{equation}
\end{proposition}
\begin{wrapfigure}{r}{0.57\textwidth}
\vspace{-1.1cm}
\begin{minipage}{0.57\textwidth}
\begin{algorithm}[H]
\caption{\textbf{ContraDICE}}
\label{algo:ContraDICE}
\begin{algorithmic}[1]
\REQUIRE Datasets $\cD^G$, $\cD^B$, $\cD^{\Mix}$; training steps $N_\mu$, $N$;\\ models: $c^G_{w_G}$, $c^B_{w_B}$, $\pi_\theta$, $Q_{w_q}$, $V_{w_v}$
\STATE Assign $\cD^U=\cD^G\cup\cD^{\Mix}$
\STATE \cmt{\textit{$\#$ Train discriminator $c^G_{w_G}$ and $c^B_{w_B}$}}
\FOR{$i = 1$ to $N_\mu$}
   \STATE Update $(w_G,w_B)$ to minimize Objective~\ref{eq:disc_objective}.
\ENDFOR
\STATE \cmt{\textit{$\#$ Train $Q_{w_q}$ and $V_{w_v}$, and policy $\pi_\theta$}}
\FOR{$i = 1$ to $N$}
   \STATE Update $w_q$ to minimize $\widetilde{F}(Q_{w_q}|V_{w_v})$
   \STATE Update $w_v$ to minimize $J(V_{w_v}|Q_{w_q})$
   \STATE Update $\theta$ via QW-BC: \\
      $\max_\pi \left\{\sum_{(s,a)\sim \cD^U} e^{Q(s,a)/\beta} \log \pi(a|s) \right\}$
\ENDFOR
\end{algorithmic}
\end{algorithm}
\vspace{-1.7cm}
\end{minipage}
\end{wrapfigure}

While the $Q$-weighted BC objective is theoretically equivalent to the advantage-weighted BC objective in terms of the optimal policy it recovers, it provides a simpler and more practical formulation. This simplification can lead to more stable and accurate optimization in practice. Our experimental results further demonstrate that the $Q$-weighted formulation consistently yields significantly better training outcomes compared to the advantage-weighted BC baseline. 
Bringing all components together, we present our \textsc{ContraDICE} algorithm in Algorithm~\ref{algo:ContraDICE}.

\section{Experiments}
In this section, we conduct extensive experiments to evaluate our method, focusing on the following key questions: \textbf{(Q1)} Can ContraDICE effectively leverage both labeled good and bad data to outperform existing baselines? 
\textbf{(Q2)} How does the size of the bad dataset $\cD^B$ affect the performance of ContraDICE? 
\textbf{(Q3)} ContraDICE relies on an important parameter $\alpha$ to balance the objectives for good and bad data—how does this parameter affect overall performance? Moreover, we also provide some additional experiments in the Appendix.

\subsection{Experiment setting}
\noindent\textbf{Environments and Dataset Generation.} We evaluate our method in the context of learning from the good dataset $\cD^G$ and avoid the bad dataset $\cD^B$ with a support from an additional unlabeled dataset $\cD^{\Mix}$. The use of such unlabeled data is common in offline imitation learning from mixed-quality demonstrations. Our experiments span four MuJoCo locomotion tasks: \textsc{cheetah, ant, hopper, walker}, as well as four hand manipulation tasks from Adroit: \textsc{pen, hammer, door, relocate}, and one task from FrankaKitchen: \textsc{kitchen}—all sourced from the official D4RL benchmark~\cite{fu2020d4rl}.
For each MuJoCo task from D4RL, we have three types of datasets: \textsc{random}, \textsc{medium}, and \textsc{expert}. The good dataset $\cD^G$ is constructed using a single trajectory from the \textsc{expert} dataset. The bad dataset $\cD^B$ consists of 10 trajectories selected from either the \textsc{random} or \textsc{medium} dataset. To construct the unlabeled dataset $\cD^{\Mix}$, we combine the entire \textsc{random} or \textsc{medium} dataset (i.e., the same source as $\cD^B$) with 30 additional trajectories from the \textsc{expert} dataset. This setup mirrors the challenging \textsc{random+few-expert} and \textsc{medium+few-expert} scenarios introduced in ReCOIL~\cite{sikchi2024dual}. These three datasets—$\cD^G$, $\cD^B$, and $\cD^{\Mix}$—form the foundation of our training pipeline.
We use the same dataset construction strategy for Adroit and FrankaKitchen tasks, yielding 18 distinct dataset combinations. Please refer to the Appendix for detailed descriptions of all dataset combinations.

\noindent\textbf{Baselines.} We compare our method against several baselines. First, we evaluate two naive Behavioral Cloning approaches: one that learns directly from the large unlabeled dataset $\cD^{\Mix}$ (BC-MIX), and one that learns solely from the good dataset $\cD^G$ (BC-G). Next, we include comparisons with state-of-the-art methods designed to leverage both expert (or good) data $\cD^G$ and unlabeled data $\cD^{\Mix}$, including SMODICE~\cite{ma2022versatile}, ILID~\cite{yue2024how}, and ReCOIL~\cite{sikchi2024dual}. We exclude DWBC~\cite{xu2022discriminator} from this experiment since both DWBC and ILID use discriminator-based objectives, and ILID has been shown to outperform DWBC. In addition, based on our proposed {objective in\eqref{eq:obj-tL-Q-pi}}, we include a variant of our method that only learns from  $\cD^G$ and $\cD^{\Mix}$ (i.e., $\alpha = 0$), called as ContraDICE-G.
For methods that incorporate support from bad data $\cD^B$, we evaluate our approach against SafeDICE~\cite{jang2024safedice}. Given the limited number of existing baselines that effectively utilize poor-quality data in offline imitation learning, we also propose a simple adaptation of DWBC, which is called as DWBC-GB to jointly learn from $\cD^G$, $\cD^B$, and $\cD^{\Mix}$. Detailed implementation of these baselines are provided in the Appendix.

\noindent\textbf{Evaluation Metrics.} We evaluate all methods using five training seeds. For each seed, we collect the results from the last 10 evaluations (each evaluation consist 10 different environment seeds), then aggregate all evaluations across seeds to compute the mean and standard deviation, which reflect the converged performance of each method. Across all experiments, we report the normalized score commonly used in D4RL tasks $\left(\text{Normalized Score} = \frac{\text{Score} - \text{Random Score}}{\text{Expert Score} - \text{Random Score}}\right)$. This normalization provides a consistent performance measure across different environments.

\noindent\textbf{Reproducibility.} We provide detailed hyperparameters and network architectures for each task in the Appendix. 
To ensure reproducibility and comparison, the source code is publicly available at: \href{https://github.com/hmhuy0/ContraDICE}{https://github.com/hmhuy0/ContraDICE} .
\subsection{Main Comparison}
\begin{table*}[htbp]\small
\begin{center}
\begin{small}
\resizebox{\textwidth}{!}{
\begin{tabular}{lllllllllll|c}
\toprule
\multirow{2}{*}{Task}&\multirow{2}{*}{unlabeled $\cD^{\Mix}$}&&&\multicolumn{4}{c}{\textbf{learning from $\cD^G$ and $\cD^{\Mix}$ only}} & \multicolumn{3}{c}{\textbf{learning with $\cD^B$}}&\\
 
\cmidrule(lr){5-8}\cmidrule(lr){9-11} 
&&    BC-MIX &BC-G&SMODICE&ILID& ReCOIL& ContraDICE-G&SafeDICE &DWBC-GB&ContraDICE &Expert\\
\midrule
\multirow{2}{*}{\textsc{cheetah}}&\textsc{random+expert}&    $2.3_{\pm0.0}$&$-0.6_{\pm0.7}$&$4.6_{\pm 2.7}$&$21.1_{\pm 7.6}$& $2.0_{\pm 0.6}$& $84.4_{\pm 5.3}$&$-0.0_{\pm 0.0}$&$2.8_{\pm 1.1}$&\highlight{86.7$_{\pm 5.0}$}&$90.6$\\
&\textsc{medium+expert}&    $42.5_{\pm0.5}$&$-0.6_{\pm0.7}$&$42.4_{\pm 3.5}$&$40.3_{\pm 15.6}$& $42.5_{\pm 0.6}$& $48.6_{\pm 4.4}$&$37.7_{\pm 0.3}$&$5.6_{\pm 4.3}$&\highlight{77.6$_{\pm 8.1}$}&$90.6$\\
\midrule
\multirow{2}{*}{\textsc{ant}}&\textsc{random+expert}&    $30.9_{\pm0.1}$&$-7.2_{\pm10.3}$&$4.6_{\pm 21.6}$&$71.8_{\pm 19.4}$& $56.2_{\pm 11.2}$& $100.6_{\pm 22.1}$&$-2.6_{\pm 0.0}$&$6.5_{\pm 7.5}$&\highlight{112.7$_{\pm 12.9}$}&$117.5$\\
&\textsc{medium+expert}&    $91.2_{\pm1.9}$&$-7.2_{\pm10.3}$&$88.5_{\pm 9.3}$&$39.6_{\pm 25.7}$& $100.8_{\pm 9.0}$& $102.4_{\pm 7.8}$&$88.1_{\pm 0.9}$&$-4.3_{\pm 5.3}$&\highlight{107.4$_{\pm 11.0}$}&$117.5$\\
\midrule
\multirow{2}{*}{\textsc{hopper}}&\textsc{random+expert}&    $4.9_{\pm0.2}$&$17.9_{\pm6.1}$&$56.4_{\pm 20.6}$&$81.6_{\pm 32.0}$& $81.0_{\pm 32.8}$& $79.4_{\pm 33.1}$&$41.1_{\pm 3.1}$&$40.8_{\pm 21.3}$&\highlight{93.6$_{\pm 20.5}$}&$109.6$\\
&\textsc{medium+expert}&    $52.2_{\pm1.3}$&$17.9_{\pm6.1}$&$53.0_{\pm 3.7}$&$87.9_{\pm 11.9}$& $46.1_{\pm 18.5}$& $70.6_{\pm 17.9}$&$55.8_{\pm 3.7}$&$21.6_{\pm 8.9}$&\highlight{103.7$_{\pm 16.3}$}&$109.6$\\
\midrule
\multirow{2}{*}{\textsc{walker}}&\textsc{random+expert}&    $1.5_{\pm0.1}$&$3.8_{\pm3.3}$&$106.6_{\pm 1.5}$&$100.1_{\pm 9.8}$& $29.8_{\pm 33.4}$& $97.5_{\pm 24.0}$&$23.0_{\pm 1.8}$&$17.4_{\pm 16.7}$&\highlight{107.4$_{\pm 3.7}$}&$107.7$\\
&\textsc{medium+expert}&    $70.8_{\pm0.7}$&$3.8_{\pm3.3}$&$6.0_{\pm 5.0}$&$89.7_{\pm 23.7}$& $72.1_{\pm 12.1}$& $99.8_{\pm 15.5}$&$60.2_{\pm 2.9}$&$25.6_{\pm 16.6}$&\highlight{108.2$_{\pm 0.9}$}&$107.7$\\
\midrule
\multirow{2}{*}{\textsc{pen}}&\textsc{cloned+expert}&    $56.0_{\pm1.1}$&$8.8_{\pm3.1}$&$10.9_{\pm 14.6}$&$1.9_{\pm 4.7}$& $79.2_{\pm 21.4}$& $66.3_{\pm 21.5}$&$19.9_{\pm 4.6}$&$9.5_{\pm 8.8}$&\highlight{96.4$_{\pm 19.4}$}&$107.0$\\
&\textsc{human+expert}&    $18.3_{\pm1.4}$&$8.8_{\pm3.1}$&$-2.5_{\pm 0.5}$&$5.1_{\pm 4.8}$& $99.9_{\pm 18.9}$& $95.5_{\pm 19.7}$&$21.8_{\pm 5.7}$&$6.5_{\pm 5.3}$&\highlight{101.5$_{\pm 18.7}$}&$107.0$\\
\midrule
\multirow{2}{*}{\textsc{hammer}}&\textsc{cloned+expert}&    $0.4_{\pm0.8}$&$1.4_{\pm0.7}$&$0.8_{\pm 0.9}$&$0.4_{\pm 1.3}$& $3.4_{\pm 4.6}$& $66.5_{\pm 26.3}$&$0.0_{\pm 0.2}$&$2.8_{\pm 5.6}$&\highlight{74.3$_{\pm 17.8}$}&$119.0$\\
&\textsc{human+expert}&    $12.8_{\pm7.3}$&$1.4_{\pm0.7}$&$1.9_{\pm 4.6}$&$1.2_{\pm 3.1}$& $113.2_{\pm 12.4}$& $113.2_{\pm 16.1}$&$0.6_{\pm 0.8}$&$3.4_{\pm 4.2}$&\highlight{120.0$_{\pm 8.3}$}&$119.0$\\
\midrule
\multirow{2}{*}{\textsc{door}}&\textsc{cloned+expert}&    $0.4_{\pm0.7}$&$-0.1_{\pm0.1}$&$-0.1_{\pm 0.1}$&$-0.1_{\pm 0.2}$& $19.3_{\pm 16.7}$& $92.6_{\pm 11.3}$&$-0.0_{\pm 0.0}$&$-0.1_{\pm 0.1}$&\highlight{102.4$_{\pm 3.8}$}&$105.3$\\
&\textsc{human+expert}&    $4.0_{\pm2.6}$&$-0.1_{\pm0.1}$&$-0.1_{\pm 0.7}$&$0.2_{\pm 1.6}$& $100.3_{\pm 6.4}$& $104.7_{\pm 1.5}$&$0.9_{\pm 0.9}$&$1.1_{\pm 1.1}$&\highlight{105.0$_{\pm 1.2}$}&$105.3$\\
\midrule
\multirow{2}{*}{\textsc{relocate}}&\textsc{cloned+expert}&    $-0.1_{\pm0.1}$&$-0.1_{\pm0.1}$&$0.1_{\pm 0.2}$&$-0.1_{\pm 0.1}$& $1.4_{\pm 2.4}$& $34.5_{\pm 13.9}$&$-0.1_{\pm 0.0}$&$-0.2_{\pm 0.1}$&\highlight{92.1$_{\pm 11.1}$}&$100.9$\\
&\textsc{human+expert}&    $0.0_{\pm0.1}$&$-0.1_{\pm0.1}$&$-0.2_{\pm 0.1}$&$-0.2_{\pm 0.2}$& $72.3_{\pm 12.6}$& $99.1_{\pm 6.9}$&$0.0_{\pm 0.1}$&$-0.1_{\pm 0.0}$&\highlight{102.6$_{\pm 5.3}$}&$100.9$\\
\midrule
\multirow{2}{*}{\textsc{kitchen}}&\textsc{partial+complete}&    $45.5_{\pm1.9}$&$2.5_{\pm5.0}$&$5.5_{\pm 8.2}$&$27.3_{\pm 5.4}$& $48.8_{\pm 8.9}$& $45.8_{\pm 14.8}$&$2.8_{\pm 1.1}$&$19.4_{\pm 4.6}$&\highlight{53.1$_{\pm 13.1}$}&75.0\\
&\textsc{mixed+complete}&    $42.1_{\pm1.1}$&$2.2_{\pm3.8}$&$3.1_{\pm 5.8}$&$13.3_{\pm 3.1}$& \highlight{50.6$_{\pm 3.8}$}& $20.3_{\pm 14.1}$&$1.5_{\pm 1.9}$&$6.7_{\pm 4.4}$&$48.9_{\pm 16.4}$&75.0\\
\midrule
\rowcolor{Gainsboro!60}
\textbf{Average}&  & 26.4& 2.9& 21.2& 32.4& 56.6& 78.8& 19.5& 9.2& \highlight{94.1}&\\
\bottomrule
\end{tabular}
}
\end{small}
\end{center}
\caption{Comparison with other baselines in MuJoCo, Adroit, and FrankaKitchen. The results are normalized score in mean and standard deviation.}
\label{tab:main_tab}
\end{table*}

To answer Question \textbf{(Q1)}, we present a comprehensive comparison between our method and existing baselines across 18 different datasets, as shown in Table~\ref{tab:main_tab}. First, both BC-MIX and BC-G fail to achieve satisfactory performance across tasks. When learning from the good dataset $\cD^G$ and the unlabeled dataset $\cD^{\Mix}$, methods like SMODICE and ILID perform reasonably well on the four MuJoCo locomotion tasks (\textsc{cheetah}, \textsc{ant}, \textsc{hopper}, \textsc{walker}) but completely fail on the five hand manipulation tasks. In contrast, ReCOIL and our method variant (ContraDICE-G) are able to successfully learn in both locomotion and manipulation tasks, demonstrating more robust generalization.

In the setting that incorporates additional low-quality data $\cD^B$, SafeDICE shows similar performance to SMODICE and ILID—again failing on the manipulation tasks. Furthermore, DWBC-GB fails to learn entirely, highlighting that a naive adaptation for leveraging poor-quality data can harm the learning process. These results suggest that incorporating bad data $\cD^B$ introduces new challenges, and that effectively utilizing such data requires a carefully designed algorithm grounded in strong theoretical principles.
Overall, our method successfully leverages the bad dataset $\cD^B$ and consistently outperforms all other baselines across both locomotion and manipulation tasks.

\subsection{Effect of Number of Bad Demonstrations}
\label{sec:main_dif_bad}
\begin{figure}[htbp]
    \centering    
    \rotatebox[origin=c]{90}{\centering Score}
    \showImage[0.175]{0}{0}{0}{0}{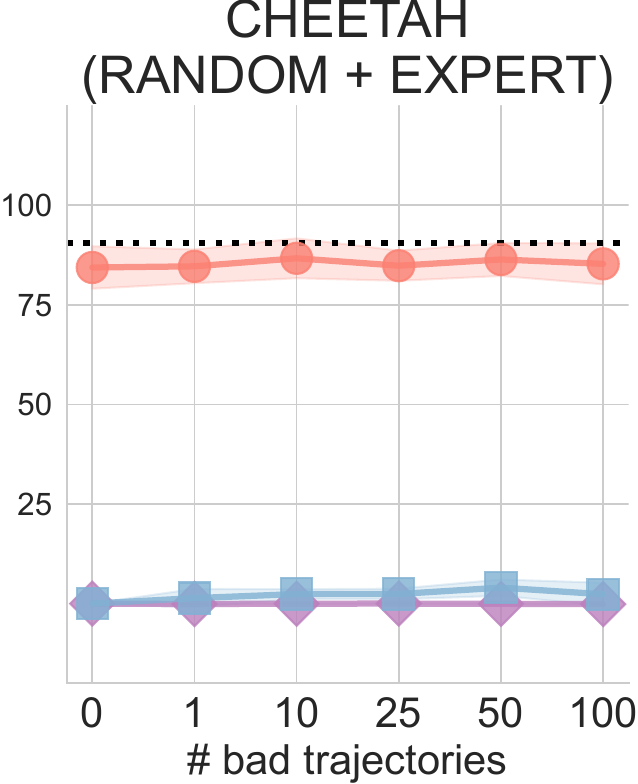}{}
    \showImage[0.175]{0}{0}{0}{0}{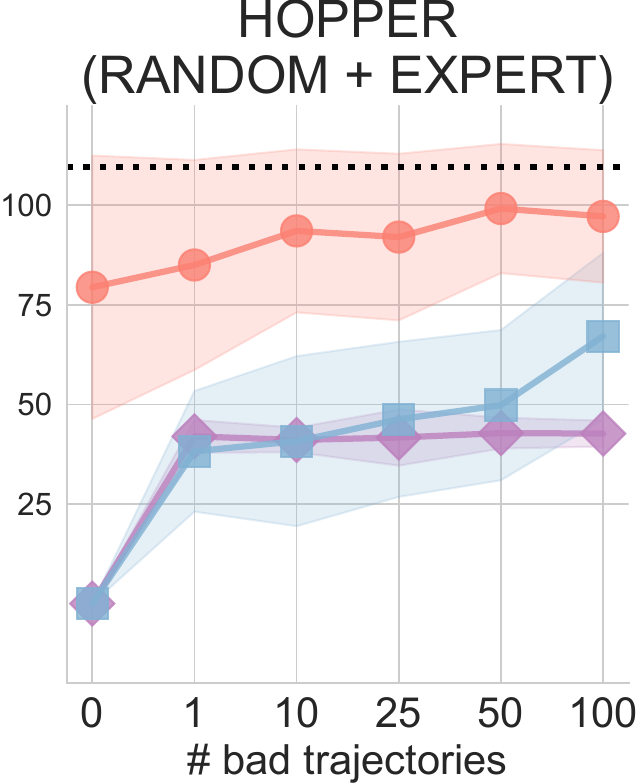}{}
    \showImage[0.175]{0}{0}{0}{0}{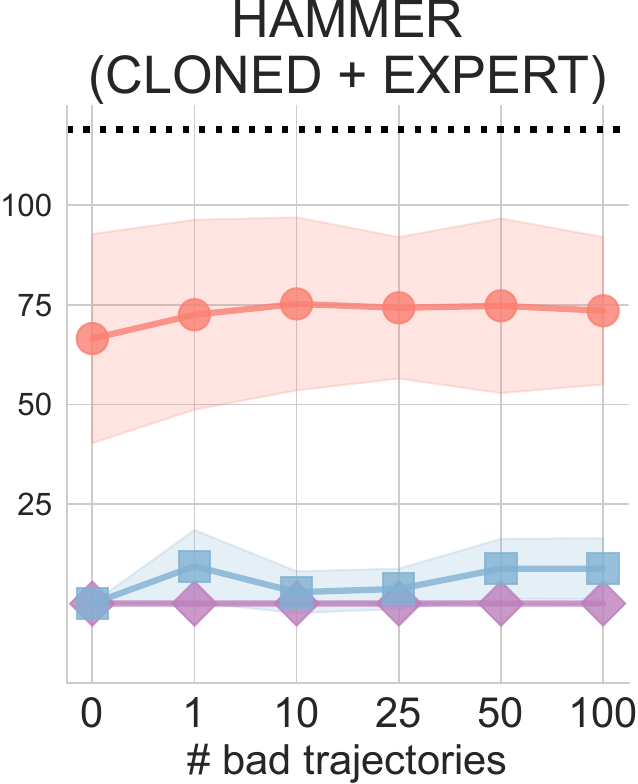}{}
    \showImage[0.175]{0}{0}{0}{0}{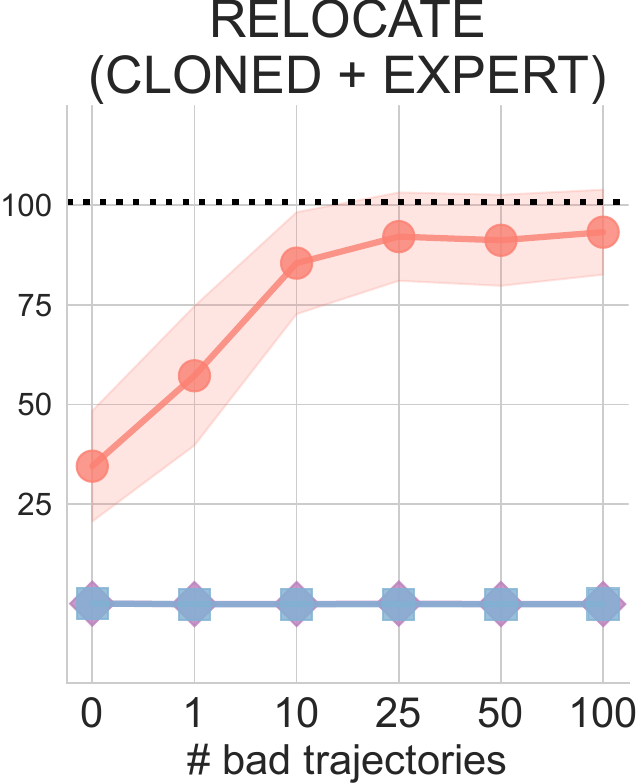}{}
    \showImage[0.175]{0}{0}{0}{0}{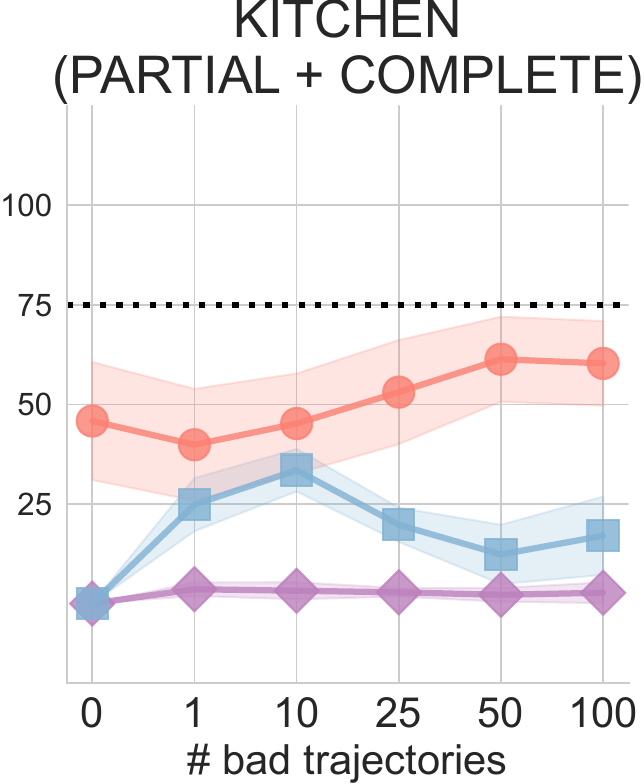}{}
    \showLegend[0.4]{0}{0}{0}{0}{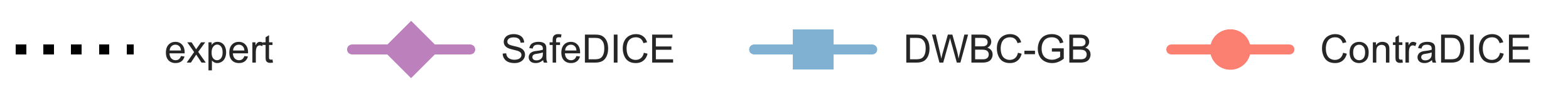}
    
    \caption{Effect of the size of the bad dataset $\cD^B$ on learning performance: The results are averaged over five different training seeds and reported using normalized scores. As the number of bad trajectories increases, our method demonstrates a strong ability to leverage this data. In contrast, baseline methods such as SafeDICE and DWBC-GB struggle to make effective use of bad demonstrations.}
    \label{fig:dif_bad}
\end{figure}
To answer question \textbf{(Q2)}, we investigate the impact of the size of the undesirable (bad) dataset on methods designed to learn from bad data. Specifically, we gradually increase the size of the bad dataset $\cD^B$ and evaluate how the performance of each algorithm is affected. The experimental results are presented in Figure~\ref{fig:dif_bad}. Overall, SafeDICE fails to effectively utilize the bad demonstrations, while DWBC-GB is only able to learn in the \textsc{hopper} task. In contrast, our method demonstrates strong scalability with respect to the size of the bad dataset, maintaining good performance even when provided with as few as a single bad trajectory.

\subsection{Sensitivity Analysis of \texorpdfstring{$\alpha$}{alpha}}
\label{sec:alpha_ablation}
\begin{wrapfigure}{r}{0.4\textwidth}
    \centering    

    \vspace{-2.0cm}
    \showImage[0.35]{0}{0}{0}{0}{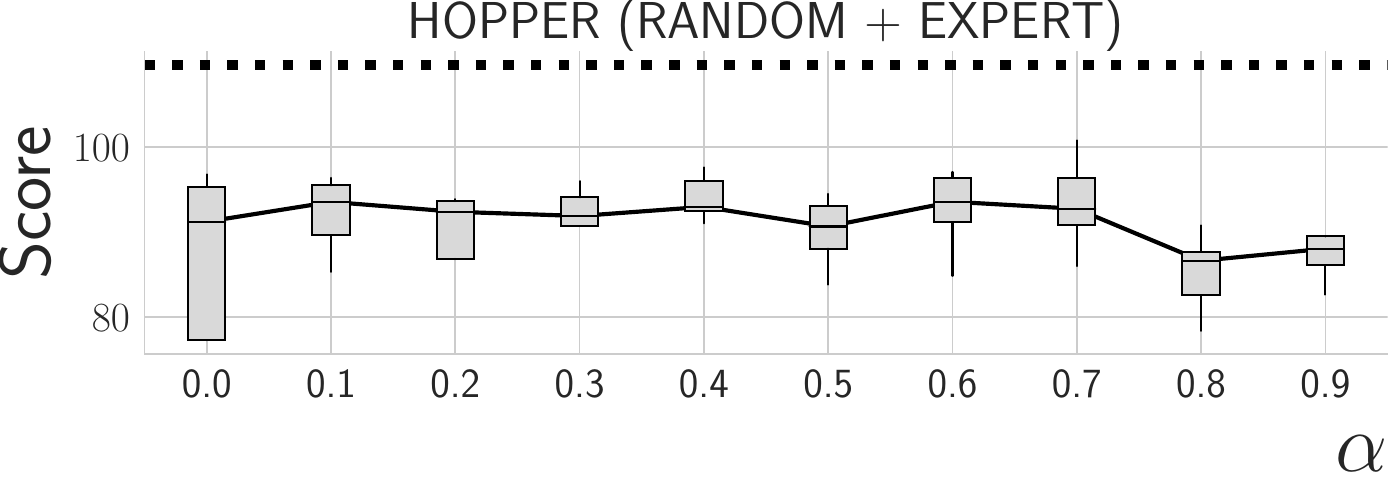}{}
    \vspace{-0.6cm}
    
    \showImage[0.35]{0}{0}{0}{0}{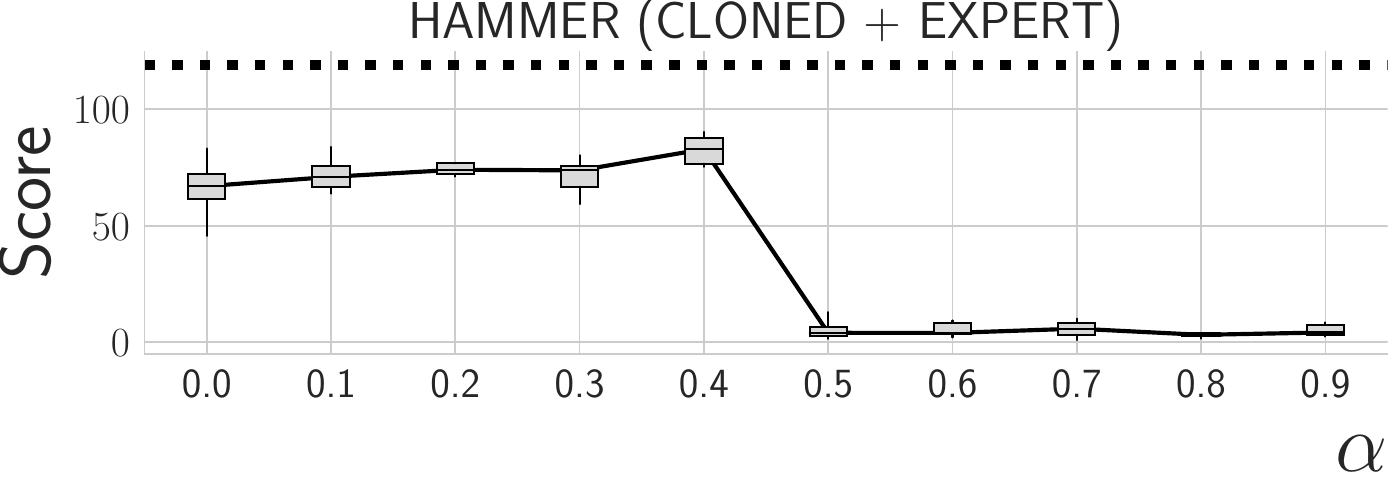}{}
    \vspace{-0.6cm}
    
    \showImage[0.35]{0}{0}{0}{0}{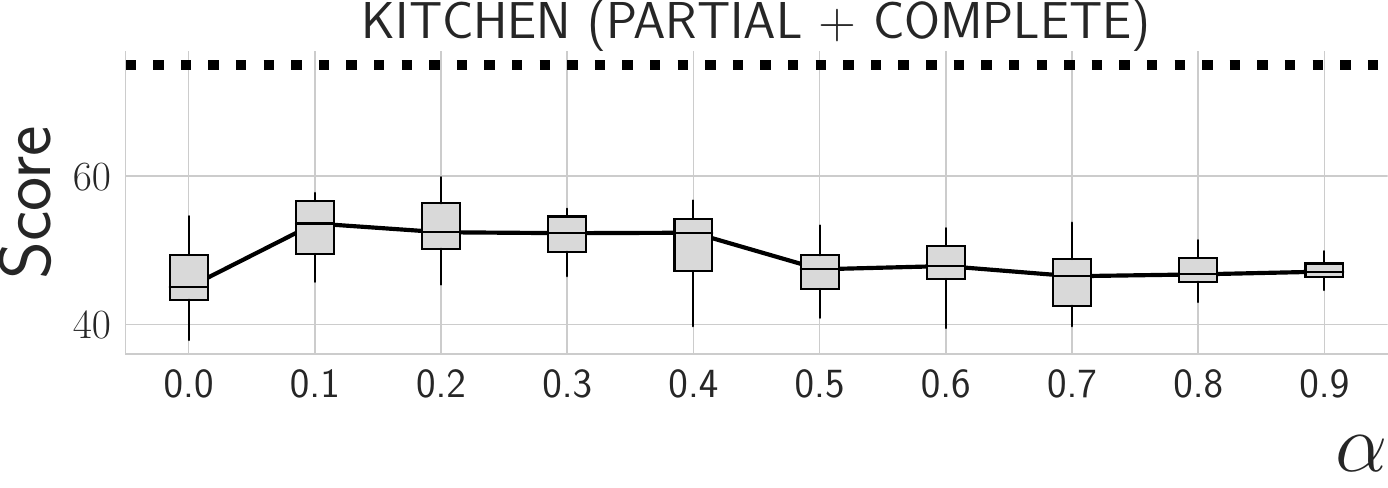}{}
    
    
    \caption{Sensitivity analysis on the trade-off parameter $\alpha$.}
    \label{fig:dif_alpha}
    \vspace{-1.2cm}

\end{wrapfigure}

From our objective function \eqref{eq:main-obj}, we introduce a hyperparameter $0 \leq \alpha < 1$, which controls the weighting of the bad data objective—this relates to question \textbf{(Q3)}. To evaluate the sensitivity of our method to $\alpha$, we conduct experiments by varying its value and observing the effect on final performance, as shown in Figure~\ref{fig:dif_alpha}. While $\alpha$ does have a noticeable impact, our method remains robust across a broad range of values, with optimal performance observed within this range. The specific $\alpha$ values used for each task are provided in the Appendix.

    

\section{Conclusion}

We introduced a new offline imitation learning framework that leverages both expert and explicitly undesirable demonstrations. By formulating the learning objective as the difference of KL divergences over visitation distributions, we capture informative contrasts between good and bad behaviors. While the resulting DC program is generally non-convex, we establish conditions under which it becomes convex—specifically, when expert data dominates—leading to a practical, stable, and non-adversarial training procedure. Our unified approach to handling both expert and  undesirable demonstrations yields superior performance across a range of offline imitation learning benchmarks, setting a new standard for learning from contrasting behaviors.
\vspace{-10pt}
\paragraph{Limitations and Future Work.} 
While our method shows strong empirical performance, it is currently limited to settings where \( \alpha \leq 1 \). Relaxing this constraint would make the learning objective more challenging to optimize, but represents a promising direction for future research. Additionally, we assume access to well-labeled expert and undesirable demonstrations, which may not hold in practice. Developing robust methods that can learn effectively from noisy or weakly labeled data would be a valuable extension of this work.

\bibliographystyle{plain}
\bibliography{references}

\newpage
\appendix
\begin{center}
    {\huge Appendix}
\end{center}

\section{Missing Proofs}

\paragraph{Proposition \eqref{prop:convex}:} \textit{ If $\alpha \leq 1$, then the objective function
$f(d^\pi) = D_{\text{KL}}(d^\pi \,\|\, d^G) - \alpha \, D_{\text{KL}}(d^\pi \,\|\, d^B)$
is convex in $d^\pi$.}
\begin{proof}
 We write the objective function as:
\begin{align}
    f(d^\pi) &= \sum_{(s,a)\sim d^\pi }\log\frac{d^\pi(s,a)}{d^G(s,a)}  
    - \alpha\sum_{(s,a)\sim d^\pi }\log\frac{d^\pi(s,a)}{d^B(s,a)} \nonumber \\
    &=  \sum_{s,a}(1-\alpha) d^{\pi}(s,a) \log p^\pi(s,a)  
    + d^\pi(s,a)(\alpha d^B(s,a) - d^G(s,a))
\end{align}
We can see that the first term is convex in $d^\pi$ since $\alpha \leq 1$ and $d^\pi(s,a)\log d^\pi(s,a)$ is convex in $d^\pi$. Moreover, the second term is linear in $d^\pi$. This implies that $f(d^\pi)$ is convex in $\pi$ if $\alpha \leq 1$, as desired.

\end{proof}

\paragraph{Proposition \ref{prop:new-form}:}
\textit{    The objective function in \eqref{eq:obj-f-d-pi} can be written as: $f(d, \pi) =   (1-\alpha)D_{\KL}(d||d^U) - \mathbb{E}_{(s, a) \sim d} \left[ \Psi(s, a) \right]$, where $\Psi(s, a) = \log \frac{d^G(s, a)}{d^U(s, a)} - \alpha \log \frac{d^B(s, a)}{d^U(s, a)}.$}
\begin{proof}
   We can expand the objective function as:
\begin{align}
f(d, \pi) &= \mathbb{E}_{(s, a) \sim d} \left[ \log \frac{d(s, a)}{d^G(s, a)} \right] - \alpha \, \mathbb{E}_{(s, a) \sim d} \left[ \log \frac{d(s, a)}{d^B(s, a)} \right].\nonumber
\end{align}
We can rewrite the objective using $d^U$ as an intermediate distribution:
\begin{align}
f(d, \pi) &= \mathbb{E}_{(s, a) \sim d} \left[ \log \frac{d(s, a)}{d^G(s, a)} \right] - \alpha \, \mathbb{E}_{(s, a) \sim d} \left[ \log \frac{d(s, a)}{d^B(s, a)} \right] \nonumber \\
&= \mathbb{E}_{(s, a) \sim d} \left[ \log \frac{d(s, a)}{d^U(s, a)} + \log \frac{d^U(s, a)}{d^G(s, a)} \right] 
- \alpha \, \mathbb{E}_{(s, a) \sim d} \left[ \log \frac{d(s, a)}{d^U(s, a)} + \log \frac{d^U(s, a)}{d^B(s, a)} \right] \nonumber \\
&= (1 - \alpha) \, \mathbb{E}_{(s, a) \sim d} \left[ \log \frac{d(s, a)}{d^U(s, a)} \right] -\mathbb{E}_{(s, a) \sim d} \left[ \Psi(s, a) \right],\nonumber\\
&= (1-\alpha)D_{\KL}(d||d^U) - \mathbb{E}_{(s, a) \sim d} \left[ \Psi(s, a) \right]\nonumber
\end{align}
where
$\Psi(s, a) = \log \frac{d^G(s, a)}{d^U(s, a)} - \alpha \log \frac{d^B(s, a)}{d^U(s, a)}.$ 
\end{proof}

\paragraph{Proposition \ref{prop:4.3}:}
\textit{Let the surrogate objective be defined as:
\begin{align}
    \widetilde{L}(Q, \pi) &= (1 - \gamma) \, \mathbb{E}_{s \sim p_0} \left[ V_Q^\pi(s) \right]
- \mathbb{E}_{d^U} \left[ \delta(s,a) \mathcal{T}^\pi[Q](s, a) \right] +(1-\alpha)\mathbb{E}_{d^U} \left[ \delta(s,a) \right].
\end{align}
where $ \delta(s,a)=  \exp\left( \frac{\Psi(s, a)}{1 - \alpha} \right).$
Then \( \widetilde{L}(Q, \pi) \) is a lower bound of \( L(Q, \pi) \), with equality when \( \mathcal{T}^\pi[Q](s, a) = 0 \) for all \( (s, a) \).}
\begin{proof}
We first write $L(Q,\pi)$ as:
\begin{align*}
L(Q, \pi) &= (1 - \gamma) \, \mathbb{E}_{s \sim p_0} \left[ V_Q^\pi(s) \right] \\
&\quad + (1-\alpha)\mathbb{E}_{(s, a) \sim d^U} \left[ \exp\left( \frac{\Psi(s, a) - \mathcal{T}^\pi[Q](s, a)}{1 - \alpha} \right) \right] \\
&= (1 - \gamma) \, \mathbb{E}_{s \sim p_0} \left[ V_Q^\pi(s) \right] \\
&\quad + (1-\alpha)\mathbb{E}_{(s, a) \sim d^U} \left[ \exp\left( \frac{\Psi(s, a)}{1 - \alpha} \right) \exp\left(\frac{- \mathcal{T}^\pi[Q](s, a)}{1 - \alpha} \right) \right] \\
&= (1 - \gamma) \, \mathbb{E}_{s \sim p_0} \left[ V_Q^\pi(s) \right] \\
&\quad + (1-\alpha)\mathbb{E}_{(s, a) \sim d^U} \left[\delta(s,a) \exp\left(\frac{- \mathcal{T}^\pi[Q](s, a)}{1 - \alpha} \right) \right],
\end{align*}
where we define $\delta(s,a) := \exp\left( \frac{\Psi(s,a)}{1 - \alpha} \right)$.

Now, we use the inequality $e^t \geq t + 1$ (which follows from the convexity of $e^t$ and is tight at $t = 0$), to obtain:
\[
\exp\left(\frac{- \mathcal{T}^\pi[Q](s, a)}{1 - \alpha} \right) \geq -\frac{ \mathcal{T}^\pi[Q](s, a)}{1 - \alpha} + 1.
\]
Substituting this into the expression for $L(Q, \pi)$, we get:
\[
L(Q, \pi) \geq (1 - \gamma) \, \mathbb{E}_{s \sim p_0} \left[ V_Q^\pi(s) \right]
+ (1-\alpha)\mathbb{E}_{(s, a) \sim d^U} \left[\delta(s,a) \left(-\frac{ \mathcal{T}^\pi[Q](s, a)}{1 - \alpha} + 1 \right) \right] =: \widetilde{L}(Q,\pi).
\]
Equality holds in the inequality $e^t \geq t + 1$ when $t = 0$, which corresponds to $\mathcal{T}^\pi[Q](s, a) = 0$. That is, the equality $L(Q,\pi) = \widetilde{L}(Q,\pi)$ holds when the rewards represented by the $Q$-function are zero everywhere.
This completes the proof.
\end{proof}

\paragraph{Proposition \ref{prop:L(Q-pi)}:}
\textit{The following properties hold:
\begin{itemize}
    \item[(i)] \( \widetilde{L}(Q, \pi) \) is linear in \( Q \) and concave in \( \pi \). As a result, the max–min optimization can be equivalently reformulated as a min–max problem:
   $ \max_{\pi} \min_{Q} \widetilde{L}(Q, \pi) = \min_{Q} \max_{\pi} \widetilde{L}(Q, \pi).$
    \item[(ii)] The min–max problem \( \min_{Q} \max_{\pi} \widetilde{L}(Q, \pi) \) reduces to the following non-adversarial problem:
    \[
    \min_{Q} \left\{ \widetilde{L}(Q) = 
    (1 - \gamma) \, \mathbb{E}_{s \sim p_0} \left[ V_Q(s) \right]
    - \mathbb{E}_{(s, a) \sim d^U} \left[\exp\left( \frac{\Psi(s, a)}{1 - \alpha}\right) \mathcal{T}[Q](s, a) \right]
    \right\},
    \]
    where the soft value function \( V_Q(s) \) is defined as:
  $  V_Q(s) = \beta\log \left( \sum_{a}\mu^U(a|s) \exp(Q(s, a)/\beta) \right),$
    and the soft Bellman residual operator is given by:
  $  \mathcal{T}[Q](s, a) = Q(s, a) - \gamma V_Q(s).$ Moreover $\widetilde{L}(Q)$ is convex in Q.
\end{itemize}}
\begin{proof}
  We first write $\widetilde{L}(Q,\pi)$ as:
\begin{align}
\widetilde{L}(Q, \pi) &= (1 - \gamma) \, \mathbb{E}_{s \sim p_0} \left[ V_Q^\pi(s) \right]
- \mathbb{E}_{(s,a) \sim d^U} \left[ \delta(s,a) \left( Q(s,a) - \gamma \mathbb{E}_{s'} \left[ V_Q^\pi(s') \right] \right) \right] \nonumber \\
&\quad + (1-\alpha)\mathbb{E}_{(s,a) \sim d^U} \left[ \delta(s,a) \right], \nonumber
\end{align}
where we recall that
\[
V_Q^\pi(s) = \mathbb{E}_{a \sim \pi(\cdot \mid s)} \left[ Q(s, a) - \beta \log \frac{\pi(a \mid s)}{\mu^U(a \mid s)} \right].
\]
Thus, we can observe that $\widetilde{L}(Q, \pi)$ is linear in $Q$.

Moreover, the function $V_Q^\pi(s)$ is concave in $\pi$, since it is composed of the expectation over a linear function of $\pi$ (through $Q(s,a)$) and the negative entropy-regularized KL-divergence term, which is convex in $\pi$ and thus its negative is concave. That is, 
\[
V^\pi_Q(s) =  \mathbb{E}_{a \sim \pi(\cdot \mid s)} \left[ Q(s, a) - \beta \log \frac{\pi(a \mid s)}{\mu^U(a \mid s)} \right]
\]
is concave in $\pi$.

Furthermore, since $\delta(s,a) > 0$, the coefficients associated with $V_Q^\pi(s)$ in $\widetilde{L}(Q, \pi)$ are non-negative. This implies that the entire function $\widetilde{L}(Q,\pi)$ is concave in $\pi$.

Now, since $\widetilde{L}(Q,\pi)$ is concave in $\pi$ and linear in $Q$, we can apply the minimax theorem to swap the order of the max and min:
\[
\max_\pi \min_Q \widetilde{L}(Q,\pi) = \min_Q \max_\pi \widetilde{L}(Q,\pi).
\]
This holds because the function $\widetilde{L}(Q,\pi)$ satisfies the standard conditions of the minimax theorem: it is concave in $\pi$, convex (in fact, linear) in $Q$, and the optimization domains are convex.

Next, observe that in $\widetilde{L}(Q,\pi)$, the variable $\pi$ only appears through the term $V_Q^\pi(s)$, and all coefficients multiplying $V_Q^\pi(s)$ are non-negative. Therefore, maximizing $\widetilde{L}(Q,\pi)$ over $\pi$ is equivalent to maximizing $V_Q^\pi(s)$ for each state $s$ independently. That is,
\[
\max_\pi \widetilde{L}(Q,\pi) \equiv \max_\pi \sum_s c(s) V_Q^\pi(s),
\]
for some non-negative coefficients $c(s) \geq 0$, which implies it suffices to solve $\max_\pi V_Q^\pi(s)$ pointwise.

Recall the definition:
\[
V_Q^\pi(s) = \mathbb{E}_{a \sim \pi(\cdot \mid s)} \left[ Q(s, a) - \beta \log \frac{\pi(a \mid s)}{\mu^U(a \mid s)} \right].
\]
The inner maximization over $\pi(\cdot \mid s)$ is a standard entropy-regularized problem, and the optimal policy has the closed-form solution:
\[
\pi^*(a \mid s) = \frac{ \mu^U(a \mid s) \exp\left( \frac{Q(s,a)}{\beta} \right) }{ \sum_{a'} \mu^U(a' \mid s) \exp\left( \frac{Q(s,a')}{\beta} \right) }.
\]
This is a weighted softmax over $Q(s,a)$ values, using the baseline distribution $\mu^U(a \mid s)$ as the reference. Substituting this back into $V_Q^\pi(s)$ yields the closed-form maximized value:
\[
\max_\pi V_Q^\pi(s) = \beta \log \left(\sum_{a} \mu^U(a \mid s) \exp\left( \frac{Q(s,a)}{\beta} \right)\right).
\]
Thus:
\[
 \min_Q \max_\pi \widetilde{L}(Q,\pi) = \min_Q \widetilde{L}(Q)
\]
where
\[
\widetilde{L}(Q) = 
    (1 - \gamma) \, \mathbb{E}_{s \sim p_0} \left[ V_Q(s) \right]
    - \mathbb{E}_{(s, a) \sim d^U} \left[\exp\left( \frac{\Psi(s, a)}{1 - \alpha} \right) \left( Q(s, a) - \gamma \, \mathbb{E}_{s'}[V_Q(s')] \right) \right],
\]
and
\[
V_Q(s) = \beta \log \sum_a \mu^U(a \mid s) \exp\left( \frac{Q(s,a)}{\beta} \right).
\]
We can now see that \(\widetilde{L}(Q)\) is convex in \(Q\), due to the following reasons:
\begin{itemize}
\setlength{\itemindent}{0pt}
\setlength{\leftmargini}{0pt}
    \item The function \(Q(s,a) \mapsto \log \sum_a \mu^U(a \mid s) \exp\left( \frac{Q(s,a)}{\beta} \right)\) is a softmax (log-sum-exp), which is convex.
    \item \(V_Q(s)\), being a composition of a convex function with an affine transformation, is convex in \(Q\).
    \item Expectations over convex functions (e.g., \(\mathbb{E}_{s \sim p_0}[V_Q(s)]\), \(\mathbb{E}_{s'}[V_Q(s')]\)) preserve convexity.
    \item The remaining terms in \(\widetilde{L}(Q)\), such as \(Q(s,a)\), appear linearly and thus preserve convexity.
\end{itemize}
Hence, the overall objective \(\widetilde{L}(Q)\) is convex in \(Q\), which completes the proof.

\end{proof}

\paragraph{Proposition \ref{prop:Q-learning}}
\textit{The following $Q$-weighted behavior cloning (BC) objective yields the same optimal policy as the original advantage-weighted BC formulation in~\eqref{eq:A-weighted-BC}:
\begin{equation}
     \max_{\pi} \sum_{(s,a)\sim \cD^U} \exp\left(\frac{1}{\beta} Q(s,a)\right) \log \pi(a \mid s).
\end{equation}
}

\begin{proof}
The Q-weighted BC objective can be written as:
\[
\max_{\pi} \sum_{(s,a)} \mu^U(s,a) \exp\left( \frac{1}{\beta} Q(s,a) \right) \log \pi(a \mid s).
\]
This represents a weighted maximum likelihood objective, where the weights are shaped by the exponential of the Q-values. For each state \(s\), the optimal solution \(\pi^*(a \mid s)\) is given by:
\[
\pi^*(a \mid s) = \frac{ \mu^U(s,a) \exp\left( \frac{1}{\beta} Q(s,a) \right) }{ \sum_{a'} \mu^U(s,a') \exp\left( \frac{1}{\beta} Q(s,a') \right) }.
\]
Moreover, we recall that:
\[
V^Q(s) = \beta \log \left( \sum_{a'} \mu^U(s,a') \exp\left( \frac{1}{\beta} Q(s,a') \right) \right),
\]
which allows us to express the optimal policy in terms of the advantage \(Q(s,a) - V^Q(s)\) as:
\[
\pi^*(a \mid s) = { \mu^U(s,a) \exp\left( \frac{1}{\beta} (Q(s,a) - V^Q(s)) \right) }.
\]
This is precisely the optimal policy corresponding to the advantage-weighted BC objective defined in Equation~\eqref{eq:A-weighted-BC}. This completes the proof.

\end{proof}

\section{A Note on ContraDICE under \texorpdfstring{$f$}{f}-Divergence}
 We note that the convexity stated in Proposition~\ref{prop:convex} does not hold under arbitrary $f$-divergences, even under the same assumptions. To illustrate this, consider the following objective defined using an $f$-divergence:
\[
F(d^\pi) = D_{f}(d^\pi \,\|\, d^G) - \alpha \, D_{f}(d^\pi \,\|\, d^B),
\]
which can be written as:
\[
F(d^\pi) = \sum_{(s,a)} d^G(s,a) f\left(\frac{d^\pi(s,a)}{d^G(s,a)}\right) - \alpha \, d^B(s,a) f\left(\frac{d^\pi(s,a)}{d^B(s,a)}\right).
\]
Observe that each term 
\[
d^G(s,a) f\left(\frac{d^\pi(s,a)}{d^G(s,a)}\right) - \alpha \, d^B(s,a) f\left(\frac{d^\pi(s,a)}{d^B(s,a)}\right)
\]
is not necessarily convex for any $\alpha > 0$. Whether this expression is convex depends on the values of $d^G(s,a)$ and $d^B(s,a)$. In particular, if $d^G(s,a) = 0$—i.e., the state-action pair $(s,a)$ is never visited by the expert policy—then the term may become concave. Therefore, in general, the objective $F(d^\pi)$ defined under an $f$-divergence is not convex in $d^\pi$ for arbitrary choices of $\alpha$. Thus, the standard Lagrangian duality cannot be applied. For this reason, the KL divergence appears to be an ideal choice for our problem of learning from both expert and undesirable demonstrations.

\newpage
\section{Experiment Settings}

\subsection{Full Pseudo Code}
The detailed implementation are provided in Algorithm~\ref{algo:full_ContraDICE}.

\begin{algorithm}[htbp]
\caption{ContraDICE: Offline Imitation Learning from Contrasting Behaviors (full)}
\label{algo:full_ContraDICE}
\begin{algorithmic}[1]
\REQUIRE Good dataset $\cD_G$, Bad dataset $\cD_B$, unlabeled dataset $\cD_U$
\REQUIRE Hyperparameters: $\alpha \in [0,1)$, $\beta$, $\gamma$, $N_\mu$, $N$, target update rate $\tau$, batch size $B$
\STATE Initialize networks: $Q_{w_q}(s,a)$, $V_{w_v}(s)$, $\pi_\theta(a|s)$, classifiers $c^G_{w_G}(s,a)$, $c^B_{w_B}(s,a)$
\STATE Initialize target Q-network: $Q_{\text{target}} \leftarrow Q_{w_q}$
\STATE
\STATE \cmt{Step 1: Estimate occupancy ratios}
\FOR{$i = 1$ to $N_\mu$}
\STATE Sample batch $\{(s^G_i)'\}_{i=1}^B \sim \cD_G$; $\{(s^B_i)'\}_{i=1}^B \sim \cD_B$; $\{(s^U_i)'\}_{i=1}^B \sim \cD_U$
\STATE Update $c^G_{w_G}$ by maximizing the objective in Equation~\eqref{eq:disc_objective}.
\STATE Update $c^B_{w_B}$ by maximizing an analogous objective to Equation~\eqref{eq:disc_objective} for the bad dataset.
\ENDFOR
\STATE
\STATE \cmt{Step 2: Calculate $\Psi$ function}
\STATE Calculate $\Psi(s,a) = \log \left( \frac{c^G_{w_G}(s')}{1 - c^G_{w_G}(s')} \right) - \alpha \log \left( \frac{c^B_{w_B}(s')}{1 - c^B_{w_B}(s')} \right)$.
\STATE
\STATE \cmt{Step 3: Train Q, V, and Policy}
\FOR{$i = 1$ to $N$}
\STATE Sample batch $\{(s_i, a_i, s'_i,\Psi_i)\}_{i=1}^B \sim \cD_U$

\STATE \textbf{Q-Update:} Minimize the objective $\tilde{L}(Q_{w_q}|V_{w_v})+\frac{1}{2} (Q_{w_q}(s_i, a_i) - \gamma V_{w_v}(s'_i))^2$ . 
\STATE \hfill \cmt{(reference: $\tilde{L}(Q|V)$ from Sec~\ref{sec:practial_algo}/ Eq~\eqref{eq:L(Q|V)})}

\STATE \textbf{V-Update:} Minimize the Extreme-V objective:
\[
\min_{w_v} \frac{1}{B} \sum_{i=1}^B \left[
\exp\left( \frac{Q_{\text{target}}(s_i, a_i) - V_{w_v}(s_i)}{\beta} \right)
- \frac{Q_{\text{target}}(s_i, a_i) - V_{w_v}(s_i)}{\beta}
- 1
\right].
\]

\STATE \textbf{Policy Update:} Maximize the policy by using Q-weighted Behavior Cloning. 
\STATE \hfill \cmt{(reference: Sec~\ref{sec:practial_algo}/ Eq~\eqref{eq:qwbc})}

\STATE \textbf{Target Q-Update:} Soft update:$Q_{\text{target}} \leftarrow \tau Q_{w_q} + (1 - \tau) Q_{\text{target}}$
\ENDFOR

\STATE \RETURN Trained policy $\pi_\theta$
\end{algorithmic}
\end{algorithm}

\newpage
\subsection{Dataset Construction}
From the official D4RL dataset we use three different domains:
\begin{itemize}
    \item MuJoCo Locomotion[\textsc{cheetah,ant,hopper,walker}] with three types of dataset:
    \begin{itemize}
        \item \textsc{expert}
        \item \textsc{medium}
        \item \textsc{random}
    \end{itemize}
    \item Adroit [\textsc{pen,hammer,door,relocate}] with three types of dataset: 
        \begin{itemize}
        \item \textsc{expert}
        \item \textsc{human}
        \item \textsc{cloned}
    \end{itemize}
    \item FrankaKitchen [\textsc{kitchen}] with three types of dataset: \begin{itemize}
        \item \textsc{complete}
        \item \textsc{mixed}
        \item \textsc{partial}
    \end{itemize}
\end{itemize}

Following the approach of~\cite{sikchi2024dual}, we also provide several combinations across all three domains, as shown in Table~\ref{tab:dataset_construction}. Notably, the unlabeled dataset $\cD^\Mix$ is constructed by combining the entire suboptimal dataset with the expert dataset, resulting in an overlap between $\cD^B$ and $\cD^\Mix$. Nevertheless, this setup is practical: given an good dataset $\cD^G$ and an unlabeled dataset $\cD^\Mix$, users can randomly sample trajectories and assign them to either $\cD^G$ or $\cD^B$ without the need for any additional external data.

\begin{table}[htbp]
\vskip 0.15in
\begin{center}
\begin{small}
\begin{tabular}{cc|lll}
\toprule
Task& Unlabeled  name& $\cD^G$& $\cD^B$& $\cD^{\Mix}$\\
\midrule
 \multirow{2}{*}{\textsc{cheetah}}& \textsc{random}+\textsc{expert}& 1 \textsc{expert}& 10 \textsc{random}& Full \textsc{random}+30 \textsc{expert}\\
 & \textsc{medium}+\textsc{expert}& 1 \textsc{expert}& 10 \textsc{medium}& Full \textsc{medium}+30 \textsc{expert}\\
\midrule
\multirow{2}{*}{\textsc{ant}}& \textsc{random}+\textsc{expert}& 1 \textsc{expert}& 10 \textsc{random}& Full \textsc{random}+30 \textsc{expert}\\
 & \textsc{medium}+\textsc{expert}& 1 \textsc{expert}& 10 \textsc{medium}&Full \textsc{medium}+30 \textsc{expert}\\
\midrule
 \multirow{2}{*}{\textsc{hopper}}& \textsc{random}+\textsc{expert}& 1 \textsc{expert}& 10 \textsc{random}&Full \textsc{random}+30 \textsc{expert}\\
 & \textsc{medium}+\textsc{expert}& 1 \textsc{expert}& 10 \textsc{medium}&Full \textsc{medium}+30 \textsc{expert}\\
\midrule
\multirow{2}{*}{\textsc{walker}}& \textsc{random}+\textsc{expert}&  1 \textsc{expert}& 10 \textsc{random}& Full \textsc{random}+30 \textsc{expert}\\
& \textsc{medium}+\textsc{expert}&  1 \textsc{expert}& 10 \textsc{medium}& Full \textsc{medium}+30 \textsc{expert}\\
\midrule
 \multirow{2}{*}{\textsc{pen}}& \textsc{cloned}+\textsc{expert}& 1 \textsc{expert}& 25 \textsc{cloned}&Full \textsc{cloned}+100 \textsc{expert}\\
 & \textsc{human}+\textsc{expert}& 1 \textsc{expert}& 25 \textsc{human}&Full \textsc{human}+100 \textsc{expert}\\
\midrule
 \multirow{2}{*}{\textsc{hammer}}& \textsc{cloned}+\textsc{expert}& 1 \textsc{expert}& 25 \textsc{cloned}&Full \textsc{cloned}+100 \textsc{expert}\\
 & \textsc{human}+\textsc{expert}& 1 \textsc{expert}& 25 \textsc{human}&Full \textsc{human}+100 \textsc{expert}\\
\midrule
 \multirow{2}{*}{\textsc{door}}& \textsc{cloned}+\textsc{expert}& 1 \textsc{expert}& 25 \textsc{cloned}&Full \textsc{cloned}+100 \textsc{expert}\\
 & \textsc{human}+\textsc{expert}& 1 \textsc{expert}& 25 \textsc{human}&Full \textsc{human}+100 \textsc{expert}\\
\midrule
 \multirow{2}{*}{\textsc{relocate}}& \textsc{cloned}+\textsc{expert}& 1 \textsc{expert}& 25 \textsc{cloned}&Full \textsc{cloned}+100 \textsc{expert}\\
 & \textsc{human}+\textsc{expert}& 1 \textsc{expert}& 25 \textsc{human}&Full \textsc{human}+100 \textsc{expert}\\
\midrule
 \multirow{2}{*}{\textsc{kitchen}}& \textsc{partial}+\textsc{complete}& 1 \textsc{complete}& 25 \textsc{partial}&Full \textsc{partial}+1 \textsc{complete}\\
 & \textsc{mixed}+\textsc{complete}& 1 \textsc{complete}& 25 \textsc{mixed}&Full \textsc{mixed}+1 \textsc{complete}\\
\bottomrule
\end{tabular}
\end{small}
\end{center}
\caption{\textbf{Dataset Construction.} The numbers in Table~\ref{tab:dataset_construction} indicate the number of trajectories drawn from each corresponding dataset. For the \textsc{Kitchen} task, we follow the setting of~\cite{sikchi2024dual}, where only a single trajectory from the \textsc{complete} dataset is included in $\cD^\Mix$.}
\label{tab:dataset_construction}
\vskip -0.1in
\end{table}

\newpage
\subsection{Baselines Implementation}
We compare our method against several established baselines. For methods with publicly available code, we utilized their official implementations without algorithmic modifications.

\subsubsection{Behavior Cloning (BC)}
We employ the standard Behavior Cloning (BC) objective, which aims to minimize the negative log-likelihood of the demonstrated actions under the learned policy:
\begin{equation}
\min_\pi -\mathbb{E}_{(s,a) \sim \cD} \log \pi(a \mid s),
\end{equation}
where $\cD$ denotes the dataset of state-action pairs. Specifically, $\cD$ corresponds to $\cD^\Mix$ in the case of BC-MIX, or $\cD^G$ for BC-G.

\subsubsection{Other Baselines with Official Implementations}
For the following baselines, we used their official, unmodified implementations:
\begin{itemize}
    \item \textbf{SMODICE}~\cite{ma2022versatile}: Applied to both the good dataset ($\cD^G$) and the mixed dataset ($\cD^\Mix$). The official code is available at \href{https://github.com/JasonMa2016/SMODICE}{[GitHub]}.
    \item \textbf{ILID}~\cite{yue2024how}: Applied to $\cD^G$ and $\cD^\Mix$. The official code is available at \href{https://github.com/HansenHua/ILID-ICML24}{[GitHub]}.
    \item \textbf{ReCOIL}~\cite{sikchi2024dual}: Applied to $\cD^G$ and $\cD^\Mix$. The official code is available at \href{https://github.com/hari-sikchi/ReCOIL}{[GitHub]}.
    \item \textbf{SafeDICE}~\cite{jang2024safedice}: Applied to the bad dataset ($\cD^B$) and the mixed dataset ($\cD^\Mix$). The official code is available at \href{https://github.com/jys5609/SafeDICE}{[GitHub]}.
\end{itemize}

\subsubsection{DWBC-GB}
DWBC-GB is our adaptation of DWBC~\cite{xu2022discriminator} (original official implementation: \href{https://github.com/ryanxhr/DWBC}{[GitHub]}). While the original DWBC is designed for scenarios involving $\cD^G$ and $\cD^\Mix$, our modified version, DWBC-GB, is extended to handle all three dataset types: $\cD^G$, $\cD^B$, and $\cD^\Mix$.

This adaptation involves training two discriminators: $c^G$ for good data and $c^B$ for bad data. Their respective loss functions are:
\begin{align}
L_{c^G} = \eta \, \mathbb{E}_{(s,a) \sim \cD^G} &[-\log c^G(s, a, \log \pi(a|s))] \nonumber \\
&+ \mathbb{E}_{(s,a) \sim \cD^\Mix} [-\log(1 - c^G(s, a, \log \pi(a|s)))] \nonumber \\
&- \eta \, \mathbb{E}_{(s,a) \sim \cD^G} [-\log(1 - c^G(s, a, \log \pi(a|s)))], \\
L_{c^B} = \eta \, \mathbb{E}_{(s,a) \sim \cD^B} &[-\log c^B(s, a, \log \pi(a|s))] \nonumber \\
&+ \mathbb{E}_{(s,a) \sim \cD^\Mix} [-\log(1 - c^B(s, a, \log \pi(a|s)))] \nonumber \\
&- \eta \, \mathbb{E}_{(s,a) \sim \cD^B} [-\log(1 - c^B(s, a, \log \pi(a|s)))].
\end{align}
The policy $\pi$ is then learned by minimizing the objective:
\begin{align}
\min_{\pi} \Biggl( & \mathbb{E}_{(s,a) \sim \cD^G} \left[ -\log \pi(a|s) \cdot \left(\alpha-\frac{\eta}{c(s,a)\left(1 - c(s,a)\right)}\right) \right] \nonumber \\
                 & + \mathbb{E}_{(s,a) \sim \cD^\Mix} \left[ -\log \pi(a|s) \cdot \frac{1}{1 - c(s,a)} \right] \Biggr),
\end{align}
where $c(s,a) = c^G(s,a) - c^B(s,a)$. (Note: $\eta$ and $\alpha$ are hyperparameters.)

\newpage
\subsection{Hyper Parameters}

Our method features two primary hyperparameters: $\alpha$ (weighting for balancing positive and negative samples) and $\beta$ (Extreme-V update). Sections~\ref{sec:alpha_ablation},~\ref{sec:large_alpha_ablation}, and~\ref{sec:beta_ablation} present ablation studies detailing the sensitivity to these parameters.

Specific parameters for all tasks are provided in Table~\ref{tab:alpha_lambda_beta_tables} below:

\begin{table}[htbp]
\vskip 0.15in
\begin{center}
\begin{small}
\begin{tabular}{ll|ll}
\toprule
 Task&Unlabeled  name&  $\alpha$&$\beta$\\
\midrule
\multirow{2}{*}{\textsc{cheetah}}&\textsc{random}+\textsc{expert}&  0.6&20.0\\
 & \textsc{medium}+\textsc{expert}& 0.6& 15.0\\
 \multirow{2}{*}{\textsc{ant}}& \textsc{random}+\textsc{expert}& 0.6& 15.0\\
 & \textsc{medium}+\textsc{expert}& 0.6& 15.0\\
 \multirow{2}{*}{\textsc{hopper}}& \textsc{random}+\textsc{expert}& 0.4& 30.0\\
 & \textsc{medium}+\textsc{expert}& 0.4& 30.0\\
 \multirow{2}{*}{\textsc{walker}}& \textsc{random}+\textsc{expert}& 0.6& 20.0\\
 & \textsc{medium}+\textsc{expert}& 0.6& 20.0\\
\multirow{2}{*}{\textsc{pen}}&\textsc{cloned}+\textsc{expert}&  0.4&15.0\\
 &\textsc{human}+\textsc{expert}& 0.4& 10.0\\
\multirow{2}{*}{\textsc{hammer}}&\textsc{cloned}+\textsc{expert}&  0.2&10.0\\
&\textsc{human}+\textsc{expert}&  0.6&20.0\\
\multirow{2}{*}{\textsc{door}}&\textsc{cloned}+\textsc{expert}&  0.4&15.0\\
&\textsc{human}+\textsc{expert}&  0.4&10.0\\
 \multirow{2}{*}{\textsc{relocate}}&\textsc{cloned}+\textsc{expert}& 0.4& 30.0\\
&\textsc{human}+\textsc{expert}&  0.8&3.0\\
\multirow{2}{*}{\textsc{kitchen}}&\textsc{partial}+\textsc{complete}&  0.3&10.0\\
&\textsc{mixed}+\textsc{complete}&  0.3&30.0\\
\bottomrule
\end{tabular}
\end{small}
\end{center}
\caption{Hyper parameters.}
\label{tab:alpha_lambda_beta_tables}
\vskip -0.1in
\end{table}

Beyond these, all other hyperparameters are consistently applied across all benchmarks and settings. The policy, Q-function, V-function, and discriminator all utilize a 2-layer feedforward neural network architecture with 256 hidden units and ReLU activation functions. For the policy, Tanh Gaussian outputs are used. The Adam optimizer is configured with a weight decay of $1 \times 10^{-3}$, all learning rates are set to $3 \times 10^{-4}$, mini batch size is 1024, and a soft critic update parameter $\tau=0.005$ is used. These hyperparameters are
summarized in Table~\ref{tab:consistent_hyperparams}:

\begin{table}[htbp]
\centering

\begin{tabular}{@{}ll@{}} 
\toprule 
\textbf{Hyperparameter} & \textbf{Value} \\
\midrule 
Network Architecture & \multirow{2}{*}{2-layer Neural Network}\\
\quad (Policy, Q-func, V-func, Discriminator) &  \\
Hidden Units per Layer & 256 \\
 Batch size&1024\\
Activation Function (Hidden Layers) & ReLU \\
Policy Output Activation & Tanh Gaussian \\
Optimizer & Adam \\
Learning Rate (all networks) & $3 \times 10^{-4}$ \\
Weight Decay (Adam) & $1 \times 10^{-3}$\\
Soft Critic Update Rate ($\tau$) & 0.005\\
\bottomrule 

\end{tabular}
\caption{Consistent hyperparameters used across all benchmarks and settings.}
\label{tab:consistent_hyperparams}
\end{table}

\subsection{Computational Resource}
Our experiments were conducted using a pool of 12 NVIDIA GPUs, including L40, A5000, and RTX 3090 models. For each experimental configuration, five training seeds were executed in parallel, sharing a single GPU, eight CPU cores, and 64 GB of RAM. Under these shared conditions, completing 1 million training steps across all five seeds took approximately 30 minutes. The software environment was based on JAX version 0.4.28 (with CUDA 12 support), running on CUDA version 12.3.2 and cuDNN version 8.9.7.29.

\newpage
\section{Additional Experiments}
\subsection{Impact of the Size of the Bad Dataset: Full Details}
\label{apdx:increase_bad_full}
\begin{figure}[htbp]
    \centering
  \showLegend[0.6]{0}{0}{0}{0}{figures/dif_bad_size_legend}
    \vspace{-0.3cm}
    
    \rotatebox[origin=c]{90}{\centering Score}
    \showImage[0.225]{0}{0}{0}{0}{figures/dif_bad_size_halfcheetah_random_expert}{}
    \showImage[0.225]{0}{0}{0}{0}{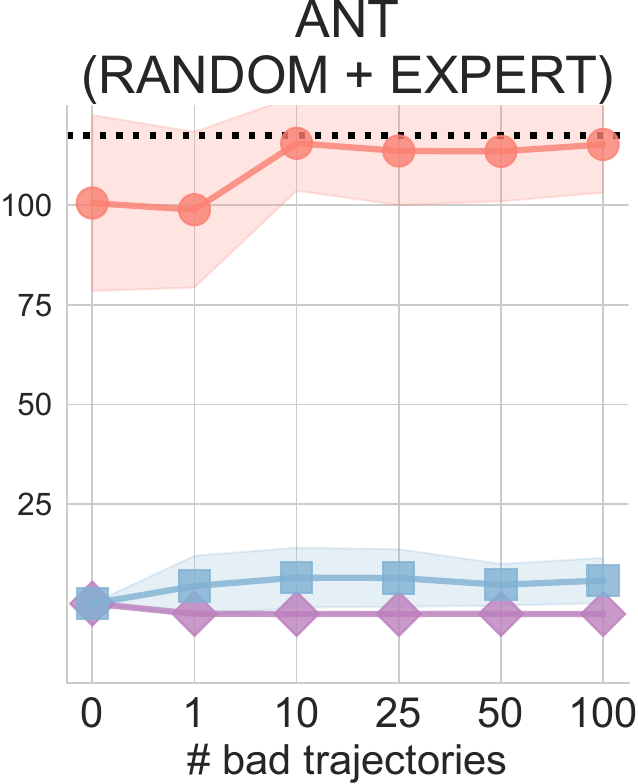}{}
    \showImage[0.225]{0}{0}{0}{0}{figures/dif_bad_size_hopper_random_expert}{}
    \showImage[0.225]{0}{0}{0}{0}{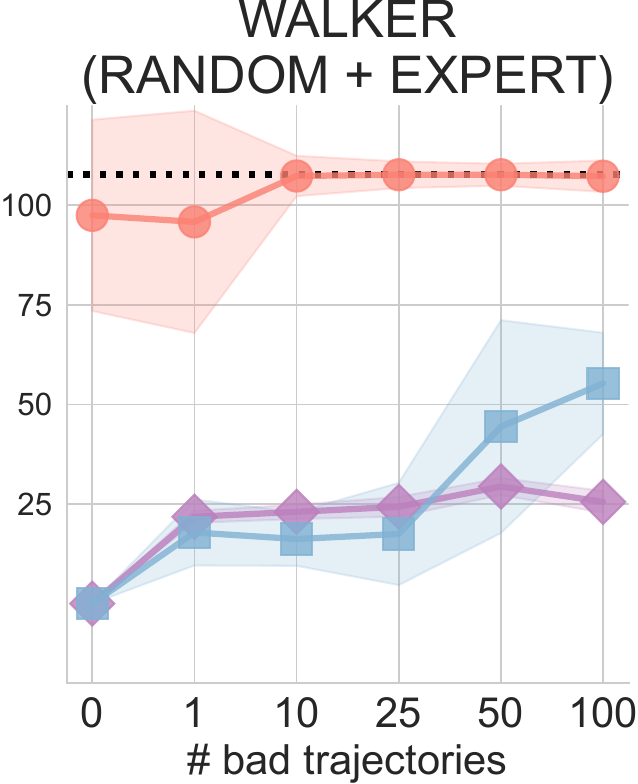}{}
    \vspace{-0.2cm}
    
    \rotatebox[origin=c]{90}{\centering Score}
    \showImage[0.225]{0}{0}{0}{0}{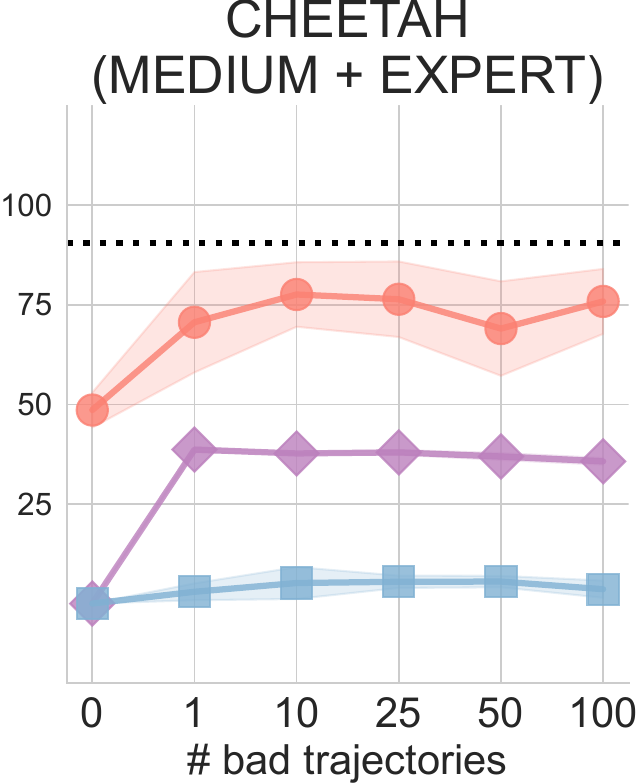}{}
    \showImage[0.225]{0}{0}{0}{0}{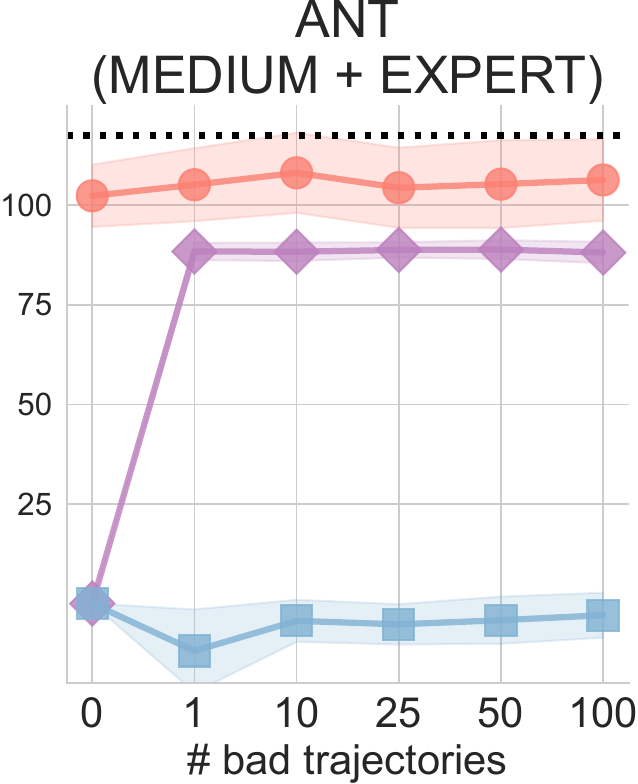}{}
    \showImage[0.225]{0}{0}{0}{0}{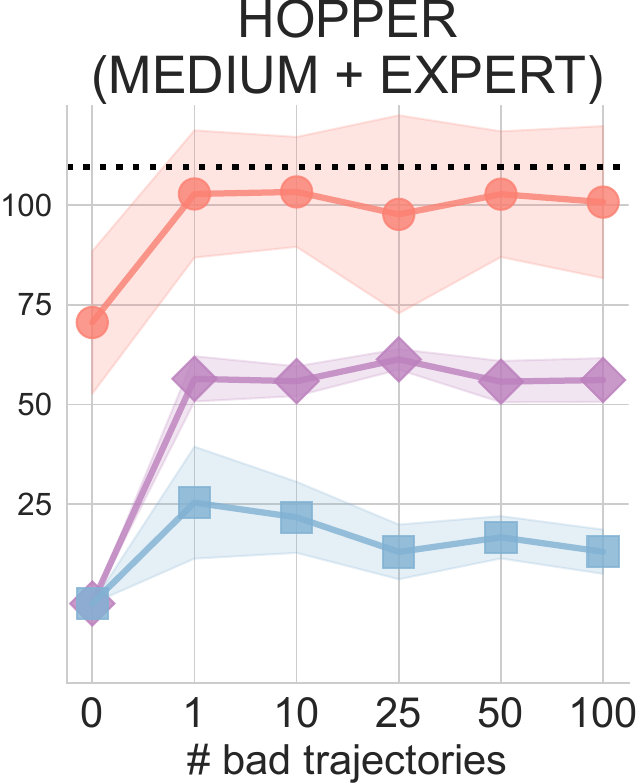}{}
    \showImage[0.225]{0}{0}{0}{0}{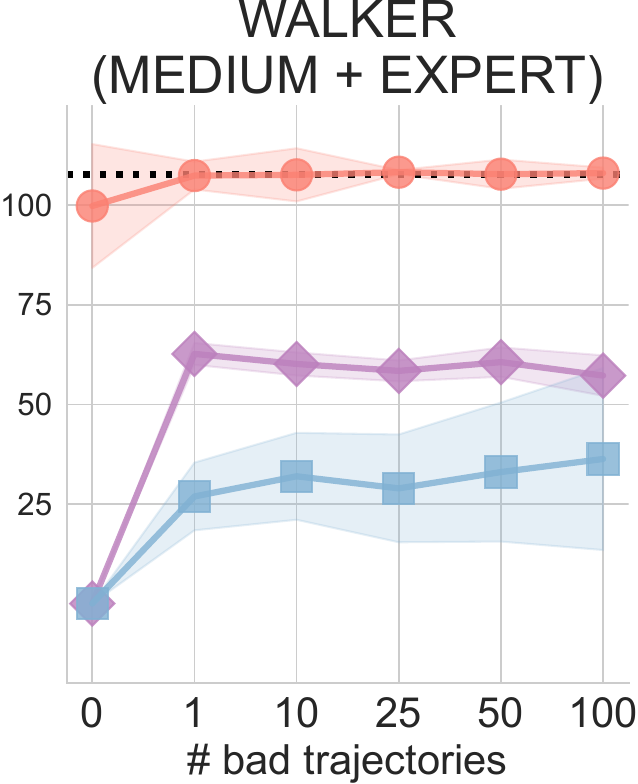}{}
    \vspace{-0.2cm}
    
    \rotatebox[origin=c]{90}{\centering Score}
    \showImage[0.225]{0}{0}{0}{0}{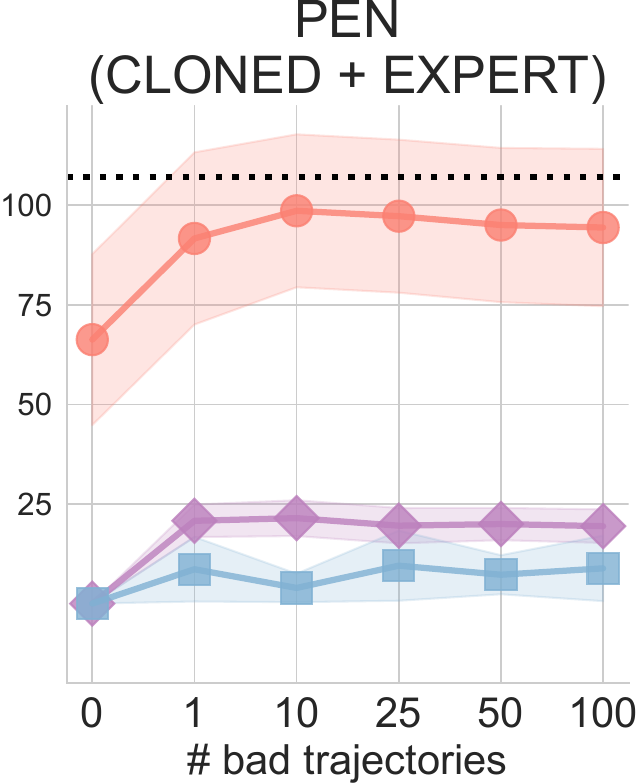}{}
    \showImage[0.225]{0}{0}{0}{0}{figures/dif_bad_size_hammer_cloned_expert}{}
    \showImage[0.225]{0}{0}{0}{0}{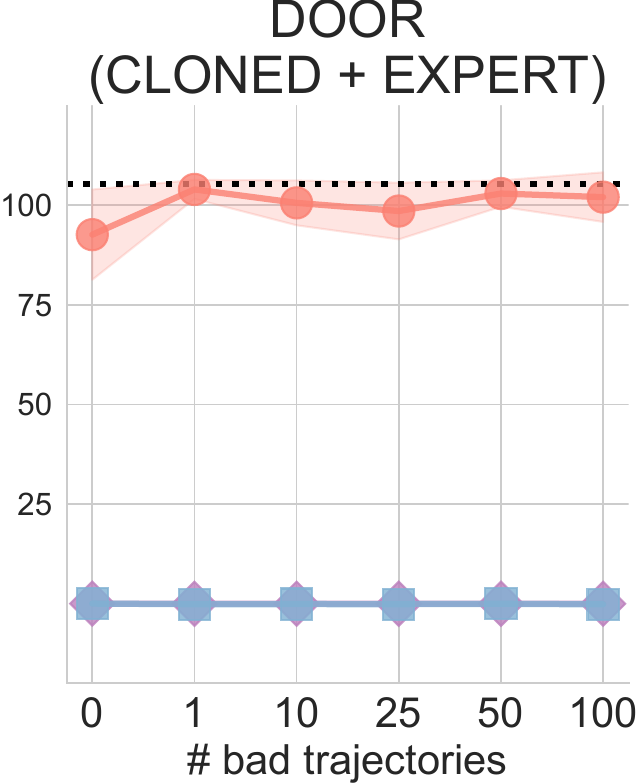}{}
    \showImage[0.225]{0}{0}{0}{0}{figures/dif_bad_size_relocate_cloned_expert}{}
    \vspace{-0.2cm}
    
    \rotatebox[origin=c]{90}{\centering Score}
    \showImage[0.225]{0}{0}{0}{0}{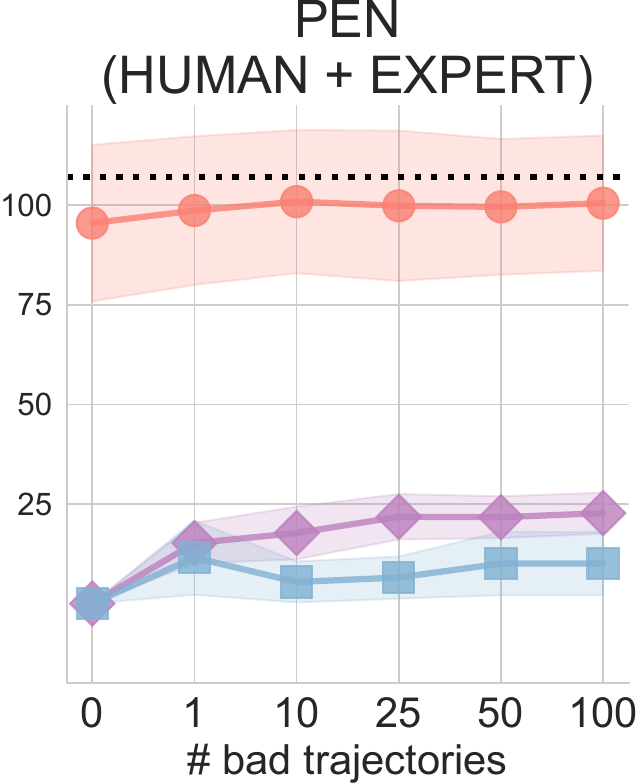}{}
    \showImage[0.225]{0}{0}{0}{0}{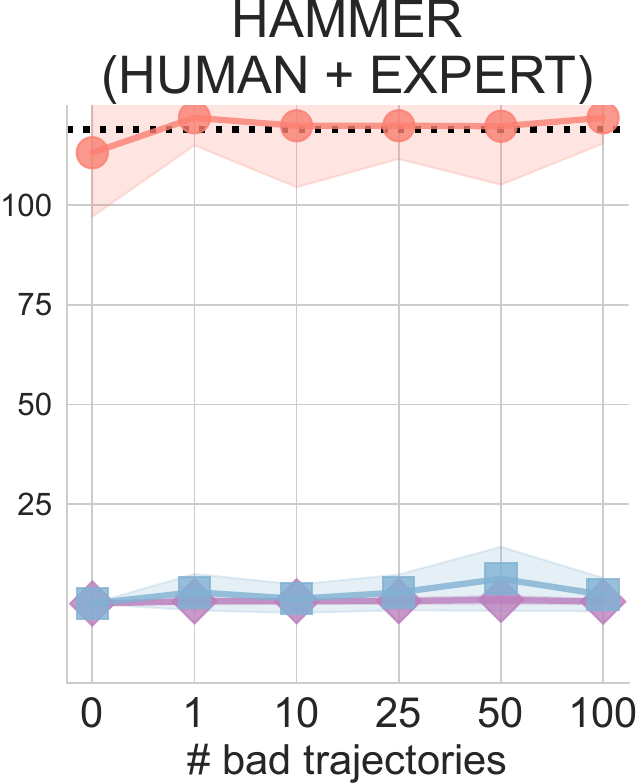}{}
    \showImage[0.225]{0}{0}{0}{0}{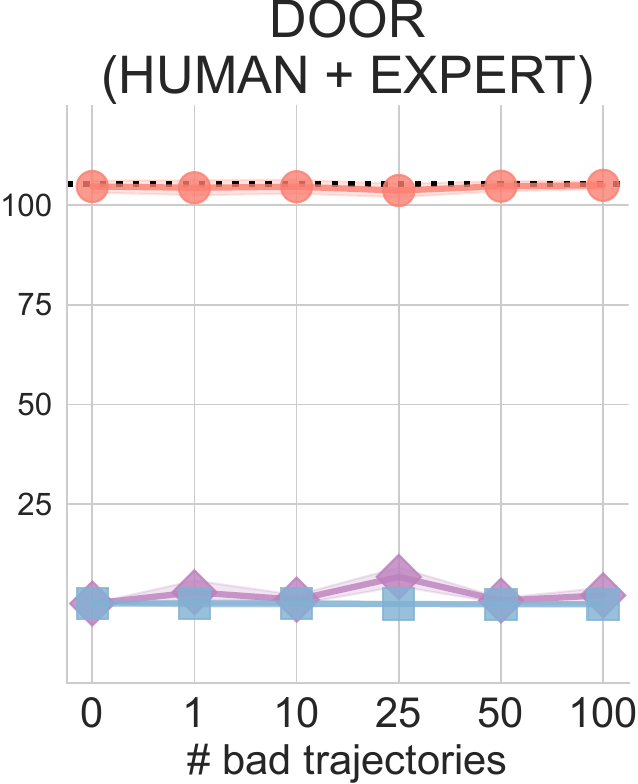}{}
    \showImage[0.225]{0}{0}{0}{0}{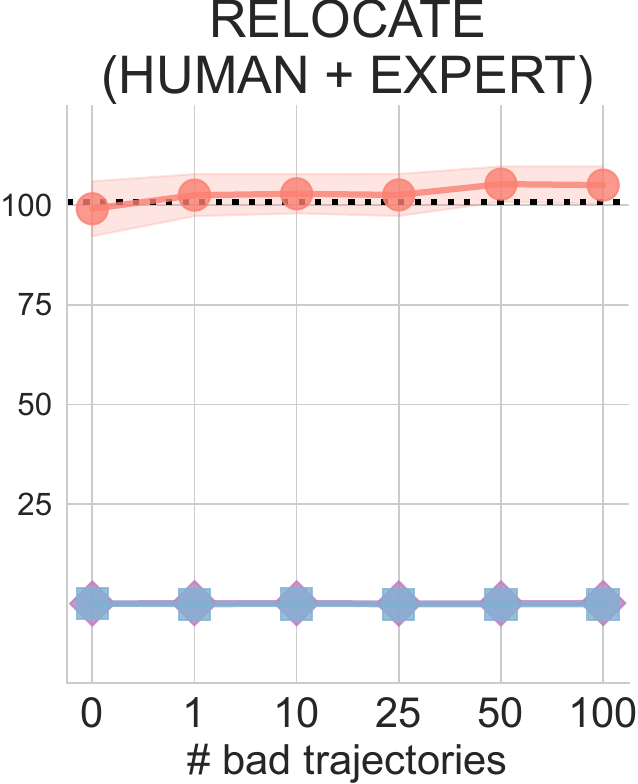}{}
    

    \caption{Full bad dataset size effect. SafeDICE and DWBC-GB do not have version that learn from 0 bad trajectory, we assign result 0.0 for them.}
    \label{fig:full_bad_size_ablation}
\end{figure}

To support the experiment in Section~\ref{sec:main_dif_bad}, we present the complete results for all MuJoCo Locomotion and Adroit manipulation tasks. In particular, we progressively increase the size of the suboptimal dataset $\cD^B$ and evaluate the impact on each algorithm's performance. The results, shown in Figure~\ref{fig:full_bad_size_ablation}, demonstrate that ContraDICE consistently outperforms all other baselines across all tasks, effectively leveraging the bad data to achieve superior performance. Notably, the results indicate that with only a single good trajectory in $\cD^G$, increasing the number of bad trajectories in $\cD^B$ to just 10 is sufficient for ContraDICE to achieve its highest performance across all tasks.

\newpage
\subsection{Impact of the Number of Expert Demonstrations in \texorpdfstring{$\cD^G$}{}}
\label{apdx:increase_good_full}
\begin{figure}[htbp]
    \centering
    
    \rotatebox[origin=c]{90}{\centering Score}
    \showImage[0.225]{0}{0}{0}{0}{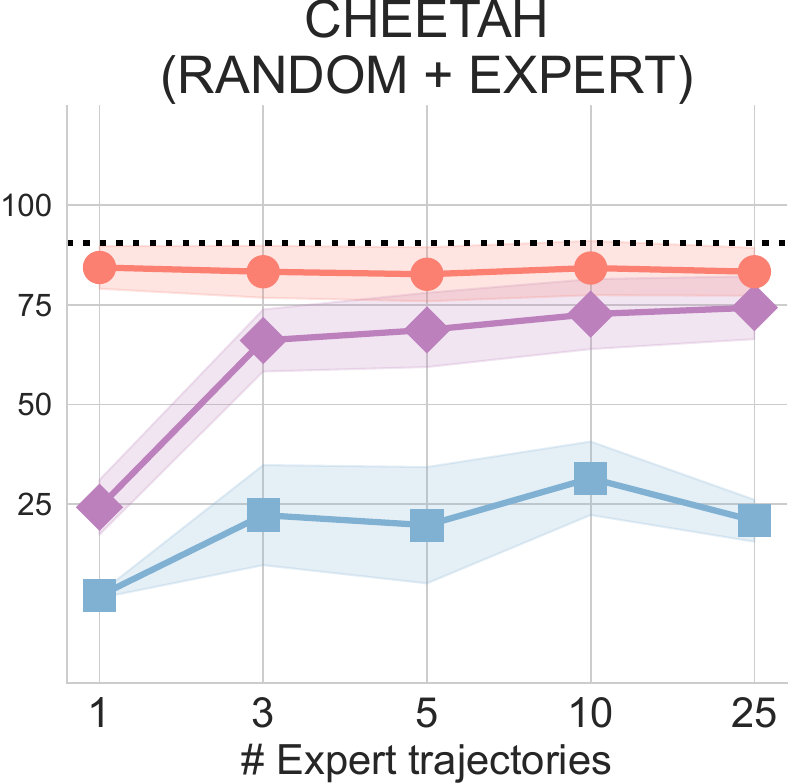}{}
    \showImage[0.225]{0}{0}{0}{0}{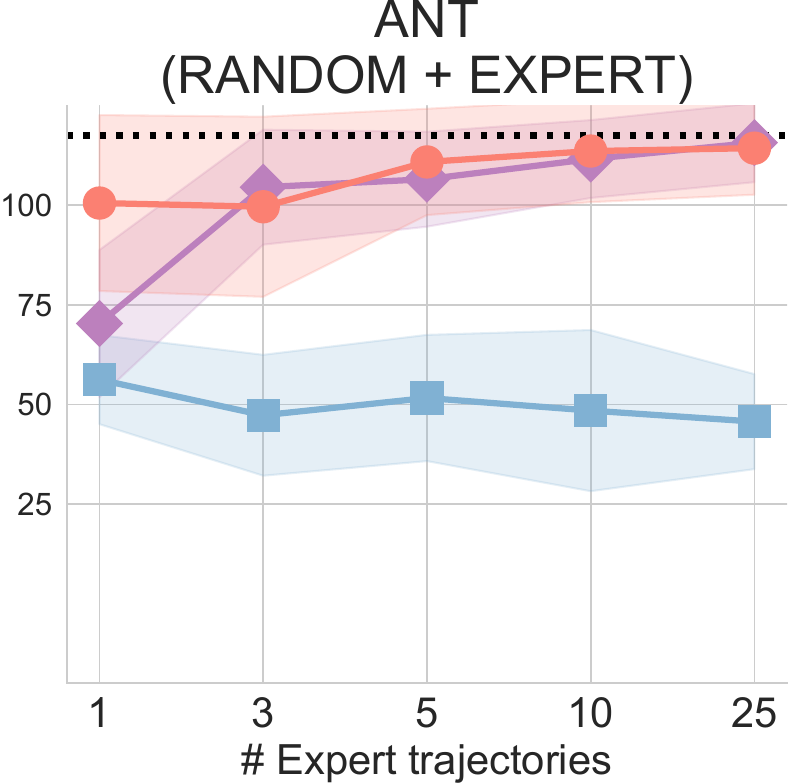}{}
    \showImage[0.225]{0}{0}{0}{0}{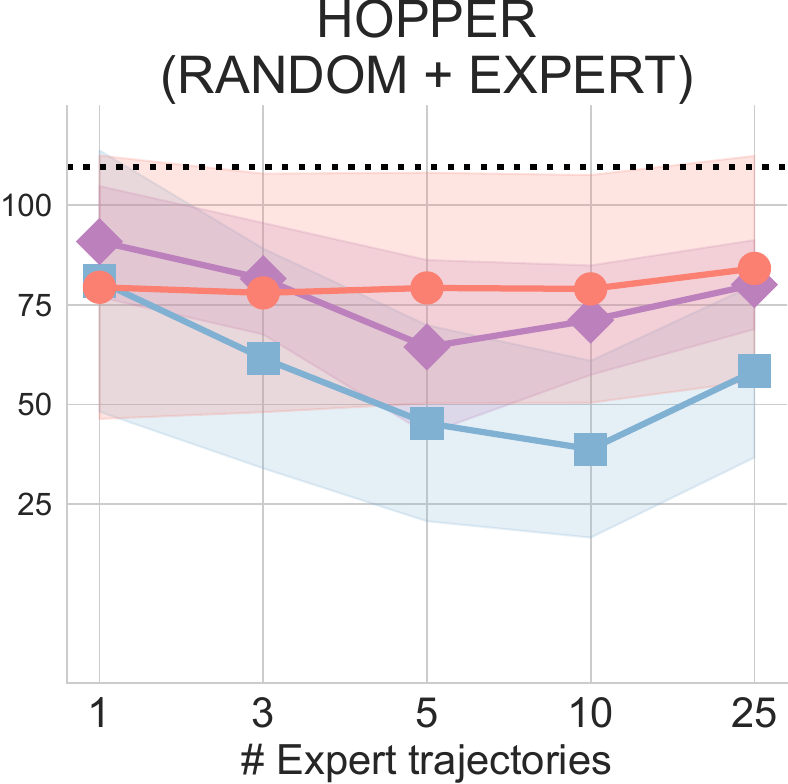}{}
    \showImage[0.225]{0}{0}{0}{0}{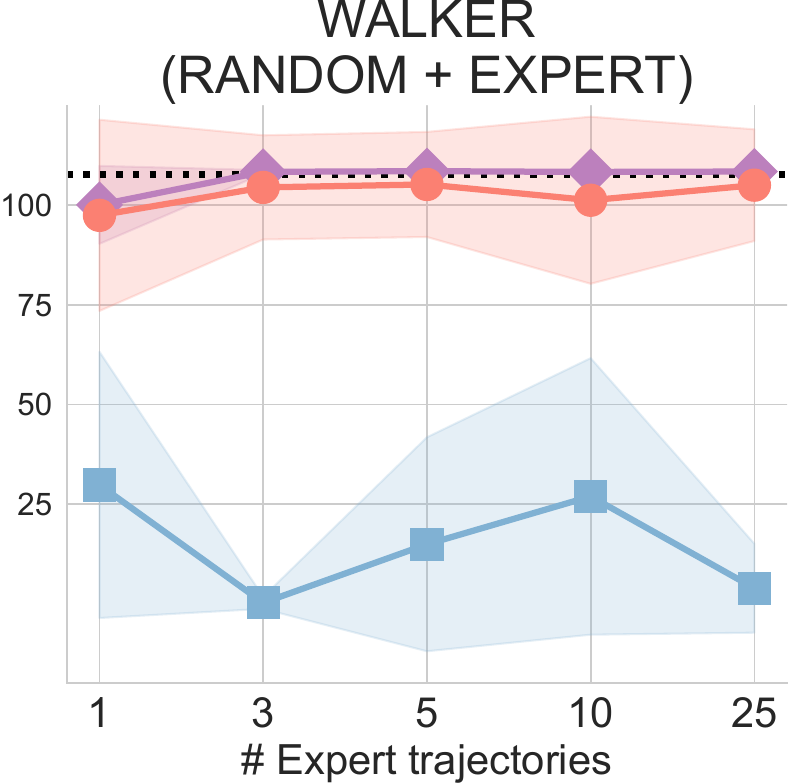}{}
    \vspace{-0.2cm}
    
    \rotatebox[origin=c]{90}{\centering Score}
    \showImage[0.225]{0}{0}{0}{0}{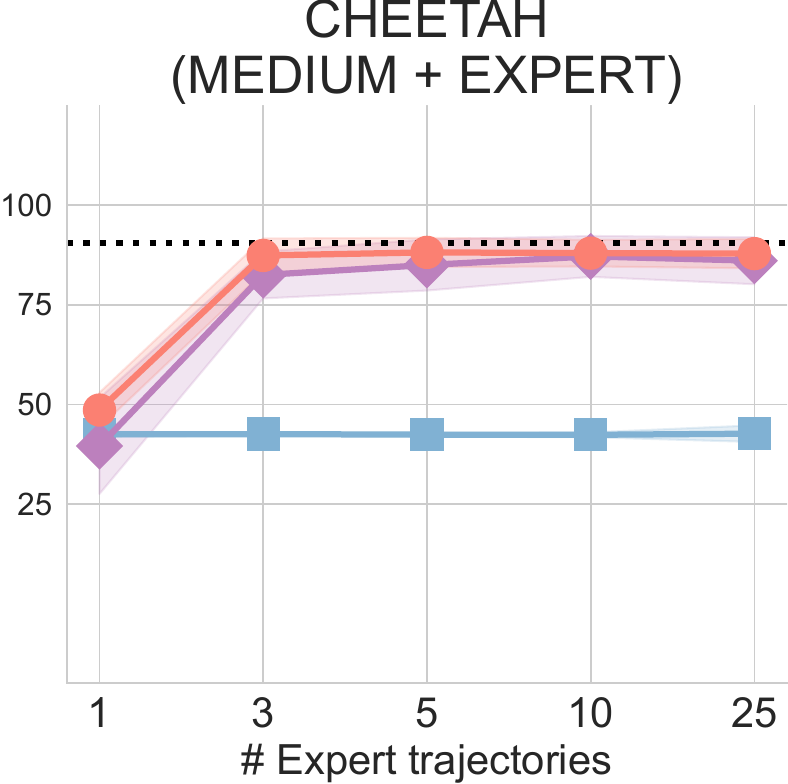}{}
    \showImage[0.225]{0}{0}{0}{0}{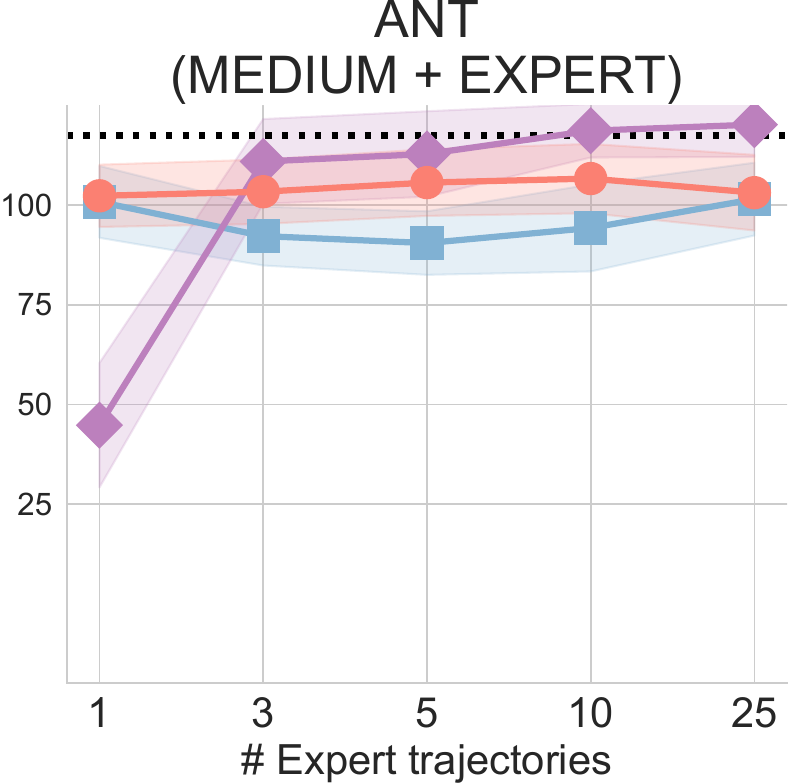}{}
    \showImage[0.225]{0}{0}{0}{0}{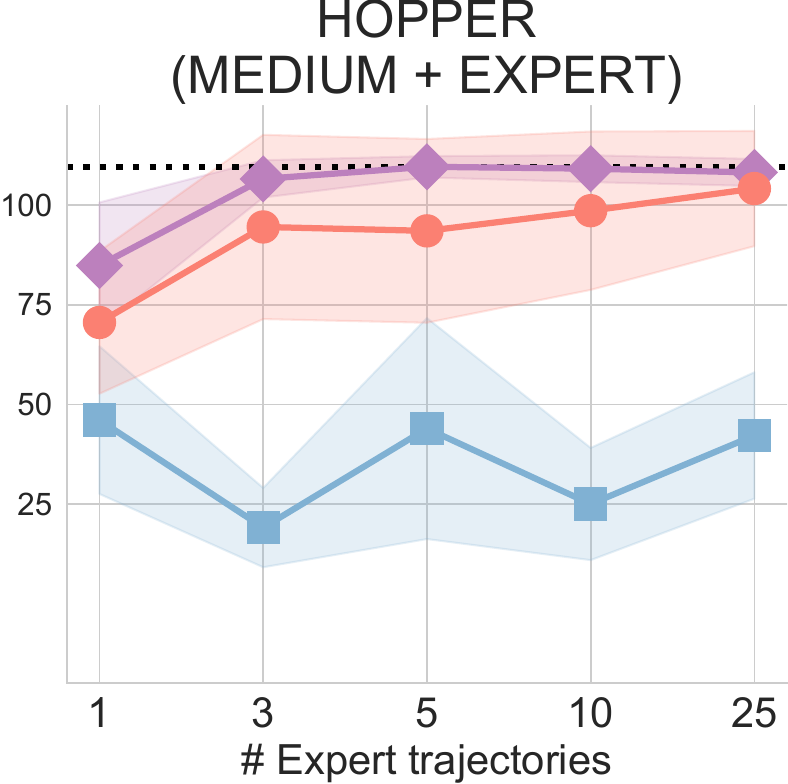}{}
    \showImage[0.225]{0}{0}{0}{0}{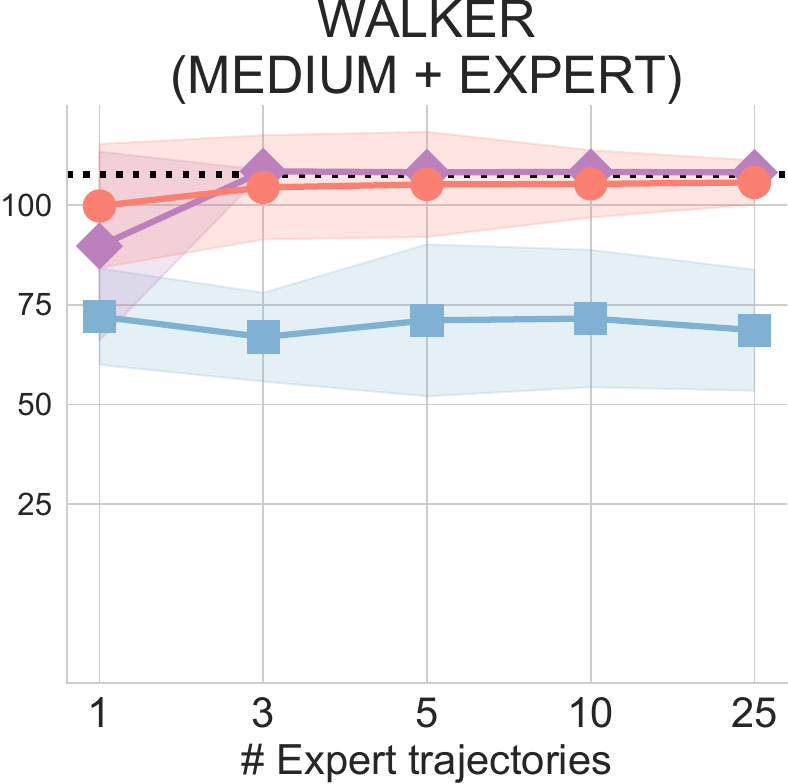}{}

  \showLegend[0.6]{0}{0}{0}{0}{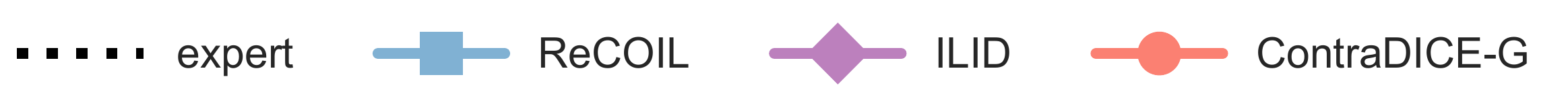}
  
    \caption{Different of good dataset size without impact from bad dataset in MuJoCo Locomotion tasks.}
    \label{fig:increase_good_ablation_mujoco}
\end{figure}

\begin{figure}[htbp]
    \centering

    \rotatebox[origin=c]{90}{\centering Score}
    \showImage[0.225]{0}{0}{0}{0}{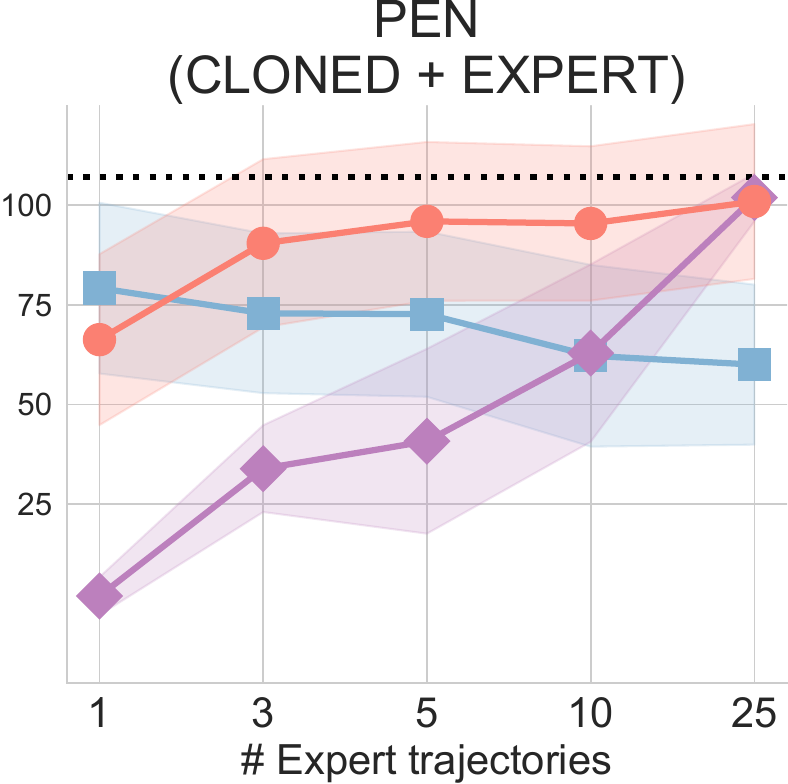}{}
    \showImage[0.225]{0}{0}{0}{0}{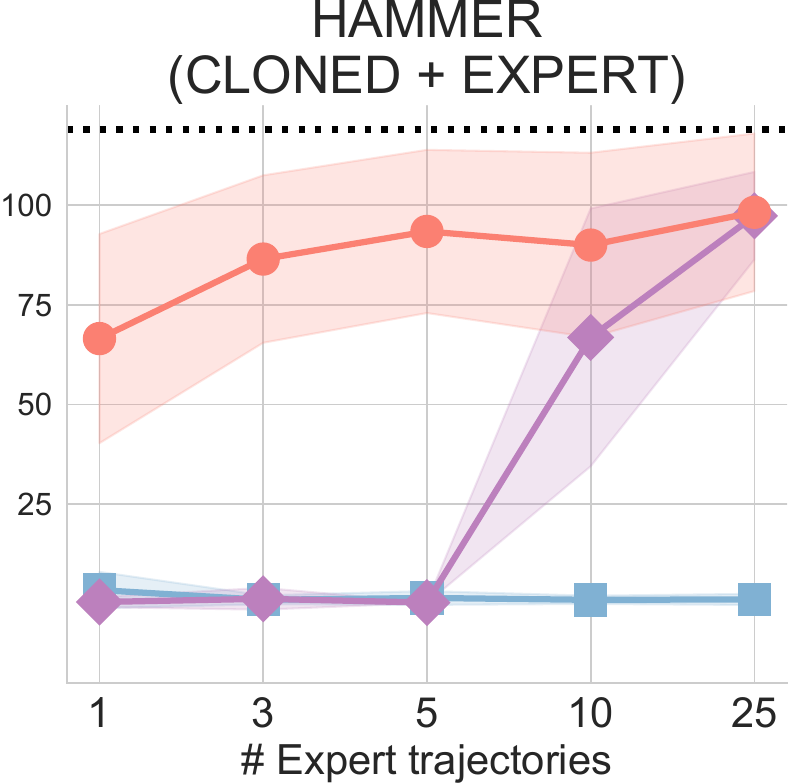}{}
    \showImage[0.225]{0}{0}{0}{0}{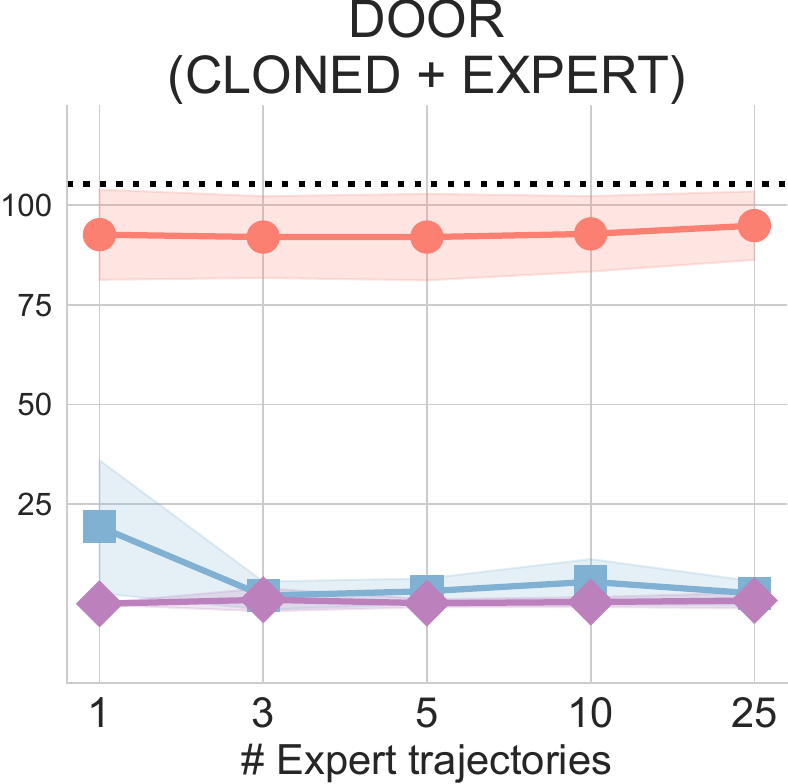}{}
    \showImage[0.225]{0}{0}{0}{0}{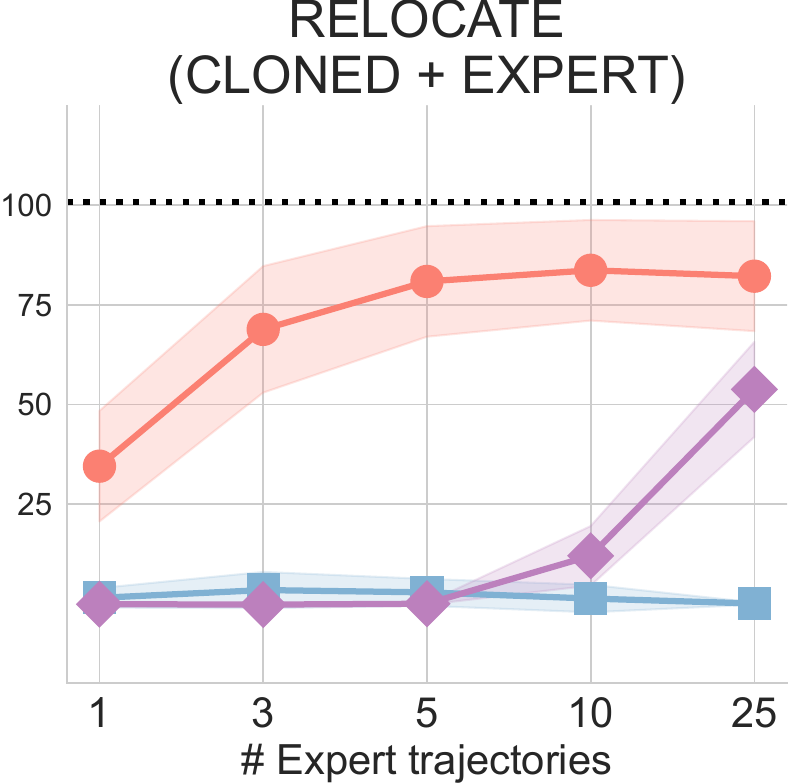}{}
    \vspace{-0.2cm}
    
    \rotatebox[origin=c]{90}{\centering Score}
    \showImage[0.225]{0}{0}{0}{0}{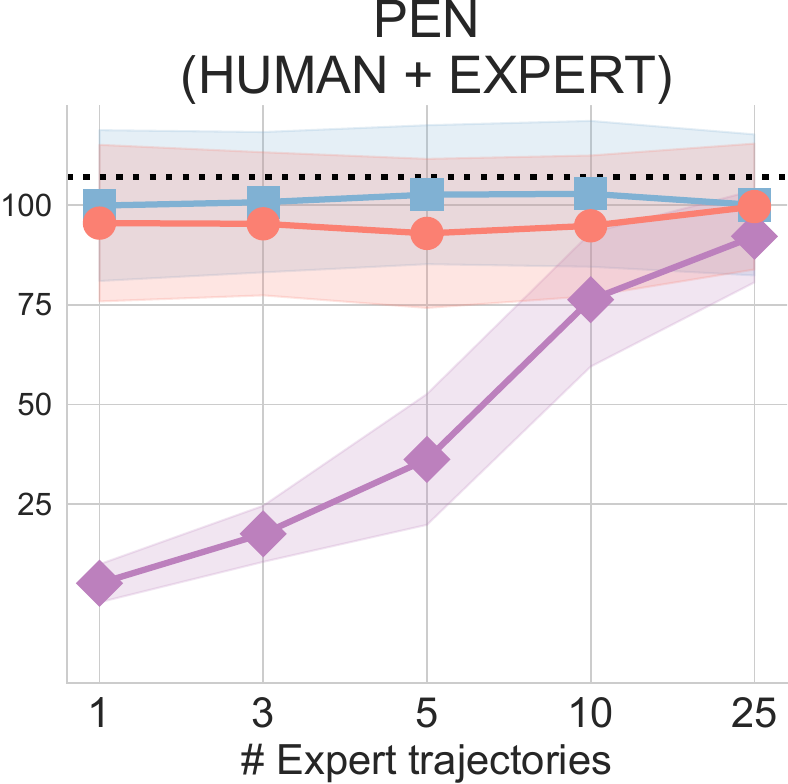}{}
    \showImage[0.225]{0}{0}{0}{0}{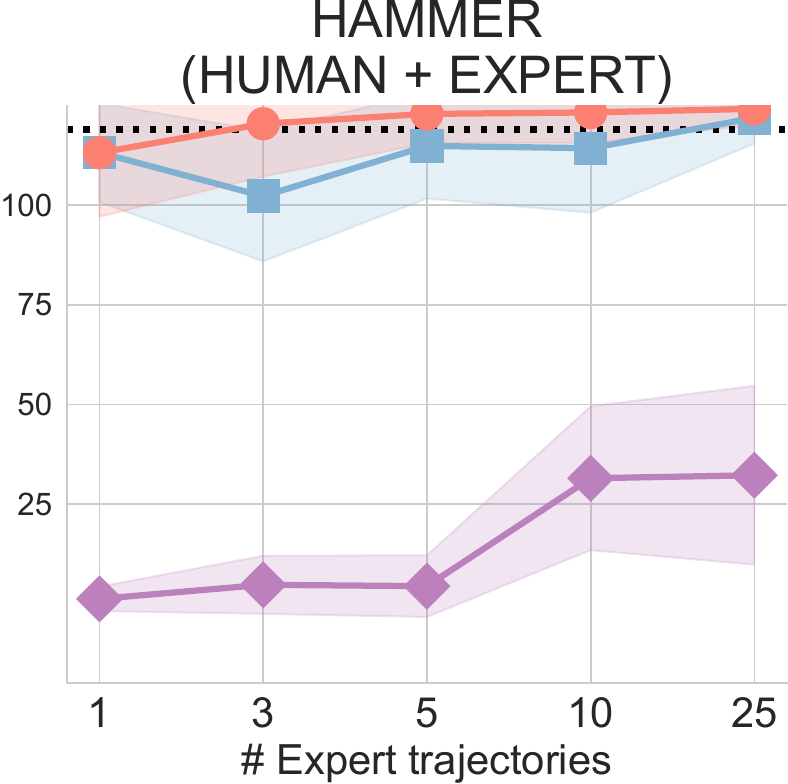}{}
    \showImage[0.225]{0}{0}{0}{0}{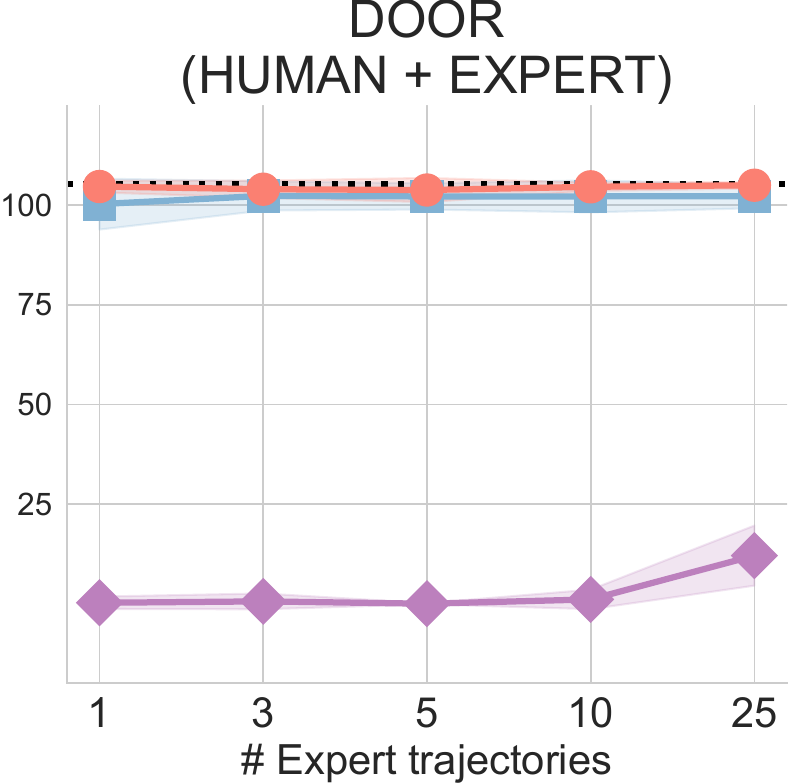}{}
    \showImage[0.225]{0}{0}{0}{0}{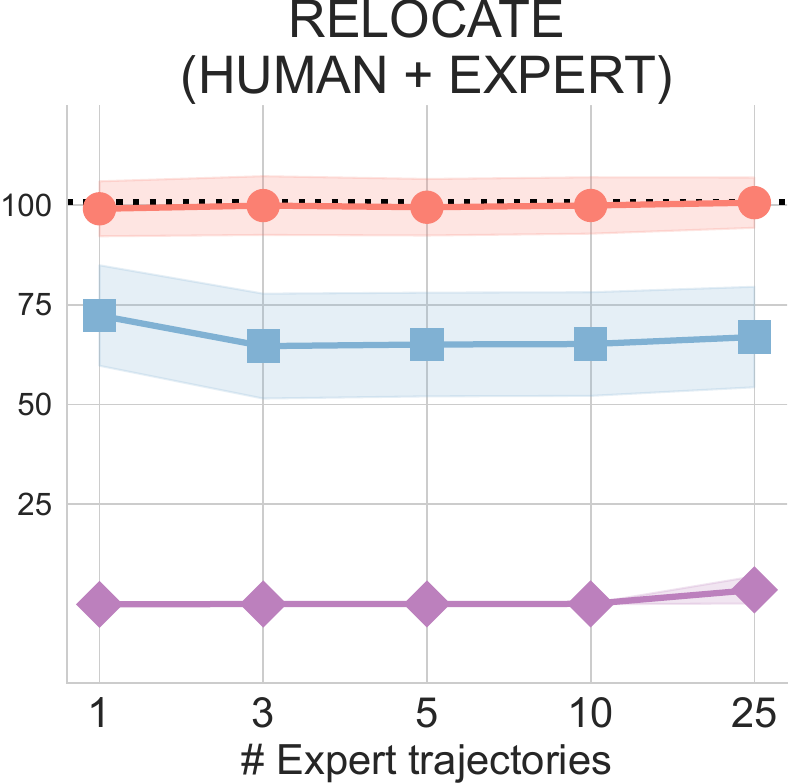}{}

  \showLegend[0.6]{0}{0}{0}{0}{figures/increase_good_legend}
  
    \caption{Different of good dataset size without impact from bad dataset in Adroit Manipulation tasks.}
    \label{fig:increase_good_ablation_androit}
\end{figure}

In this section, we investigate how many expert trajectories in the good dataset $\cD^G$ are sufficient to achieve optimal performance. To this end, the quantity of expert trajectories in $\cD^G$ was incrementally increased through the set {1,3,5,10,25}, while the composition of the unlabeled dataset ($\cD^{\Mix}$) remained fixed, as specified in Table~\ref{tab:main_tab}. The detailed results are presented in Figure~\ref{fig:increase_good_ablation_mujoco} and~\ref{fig:increase_good_ablation_androit}.

ILID performs well on the Mujoco locomotion tasks (\textsc{cheetah, ant, hopper, walker}), but struggles in 3 out of 4 Adroit tasks (\textsc{hammer, door, relocate}). This indicates that ILID requires a sufficient number of expert trajectories to achieve stable expert performance, which is not met in the more complex Adroit tasks. In contrast, ReCOIL appears unable to effectively leverage the good data, as its performance does not improve significantly with more expert trajectories. Overall, ContraDICE demonstrates consistently strong performance, \textbf{requiring only 3 to 5 expert trajectories} to achieve near-optimal results in all tasks.

**\textbf{Discussion on the Use Cases of ILID and ContraDICE:}**
Through this experiment, we observe that in the Mujoco tasks, ILID can outperform ContraDICE-G when the size of the good dataset is sufficiently large. This highlights a limitation of ContraDICE, where the policy extraction objective is defined as $\max_\pi \left\{\sum_{(s,a)\sim \cD^U}  \exp(\frac{1}{\beta} Q(s,a)) \log \pi(a|s) \right\}$. This objective uses data from the union dataset $\cD^U$, which may assign high weights to poor-quality transitions, potentially harming training.

In contrast, ILID only retains transitions that are connected to good data and explicitly discards irrelevant or undesirable transitions (refer to the implementation details of ILID for more information). This targeted filtering strategy enables ILID to avoid the negative effects of poor transitions and scale more effectively with increasing amounts of good data.

These observations suggest a potential direction for improving ContraDICE by incorporating similar data filtering mechanisms. Specifically, enhancing ContraDICE to better isolate high-quality transitions could help it perform competitively with ILID in scenarios where the good dataset is large. We leave this exploration for future work, as it requires a careful study of how to construct an optimal dataset using Q-based methods.

In summary, ILID is a strong approach that scales well with the quality and size of the expert dataset. Practitioners may prefer discriminator-based methods like ILID when sufficient high-quality expert data is available, while ContraDICE remains a robust choice in settings where such data is limited and scalalbe with bad dataset.

\subsection{Discussion: How Many Bad Trajectories in \texorpdfstring{$\cD^B$}{} Are Sufficient to Replace a Good Trajectory in \texorpdfstring{$\cD^G$}{} for ContraDICE?}

Based on the previous experiments:
\begin{itemize}
\item Section~\ref{apdx:increase_bad_full} addresses the question: How does the size of the bad dataset $\cD^B$ affect the performance of ContraDICE?
\item Section~\ref{apdx:increase_good_full} investigates an additional question: How does the size of the good dataset $\cD^G$ affect the performance of ContraDICE?
\end{itemize}

From these experiments, we derive the following observations:
\begin{itemize}
\item With only one good trajectory in $\cD^G$, adding 10 bad trajectories in $\cD^B$ is sufficient for ContraDICE to achieve its best performance.
\item Without any bad data $\cD^B$, 3 to 5 good trajectories in $\cD^G$ are enough to reach peak performance.
\end{itemize}

These results suggest that ContraDICE can efficiently utilize bad data to reduce the need for good data, with an estimated ratio of 2 to 5 bad trajectories being roughly equivalent to one good trajectory across the benchmarks studied in this paper.

\newpage
\subsection{Comparison of Advantage-weighted BC and Q-weighted BC for the Policy Extraction}
\begin{figure}[htbp]
    \centering
    \rotatebox[origin=c]{90}{\centering Score}
    \showImage[0.225]{0}{0}{0}{0}{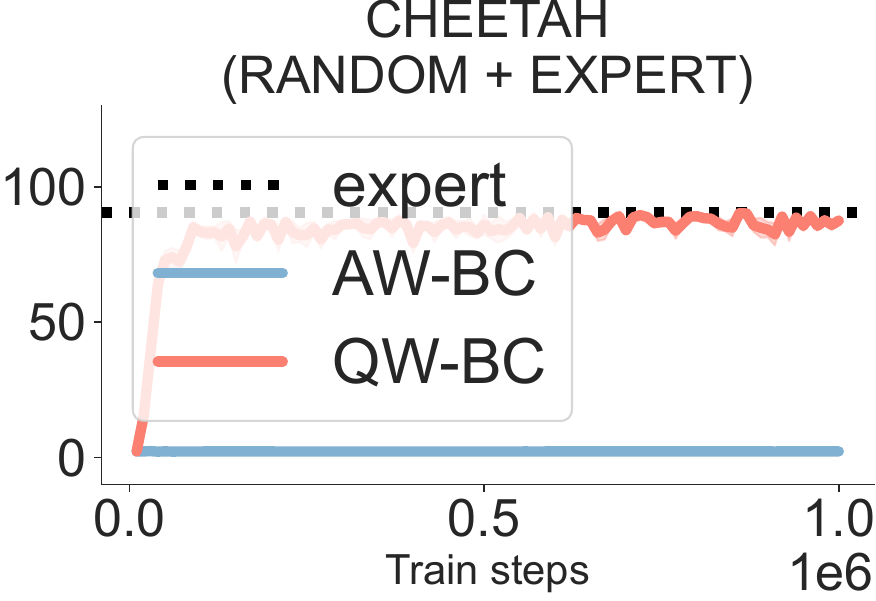}{}
    \showImage[0.225]{0}{0}{0}{0}{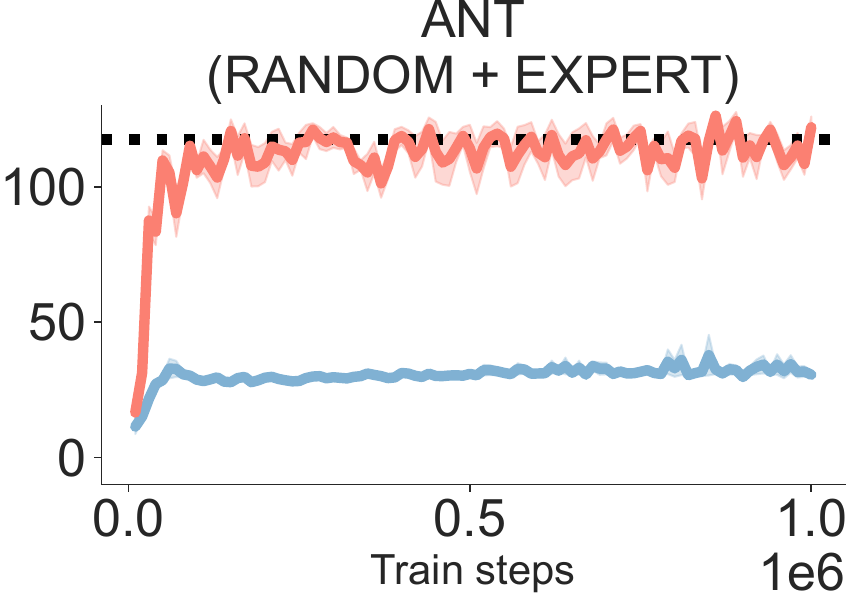}{}
    \showImage[0.225]{0}{0}{0}{0}{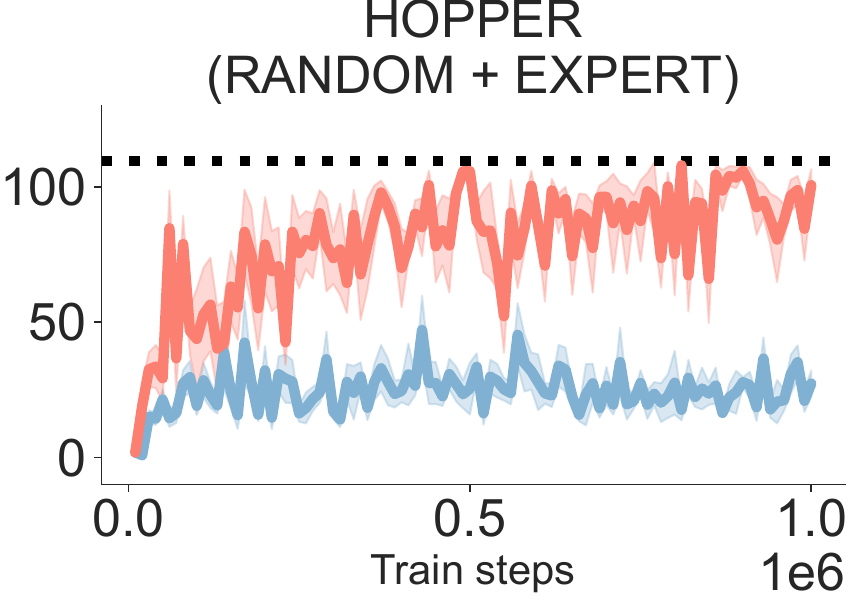}{}
    \showImage[0.225]{0}{0}{0}{0}{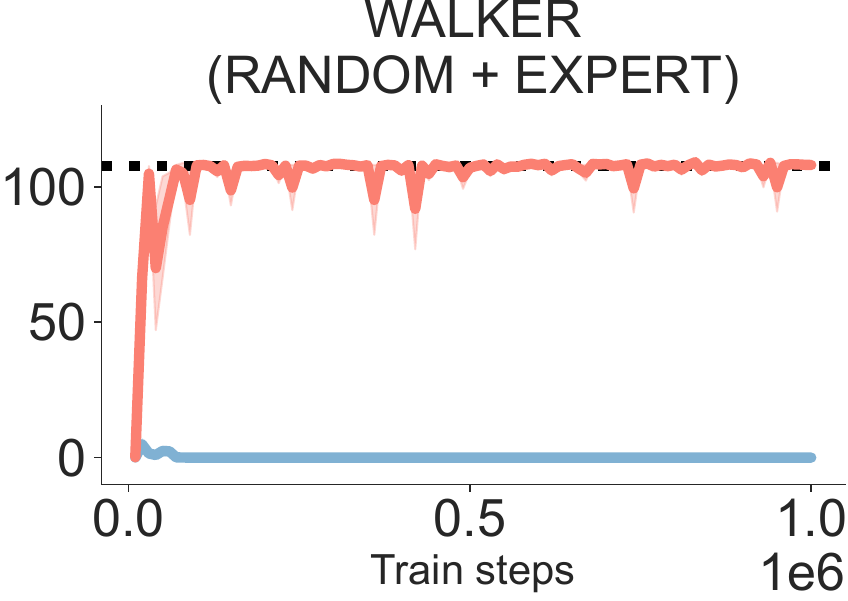}{}
    \vspace{0.5cm}
    
    \rotatebox[origin=c]{90}{\centering Score}
    \showImage[0.225]{0}{0}{0}{0}{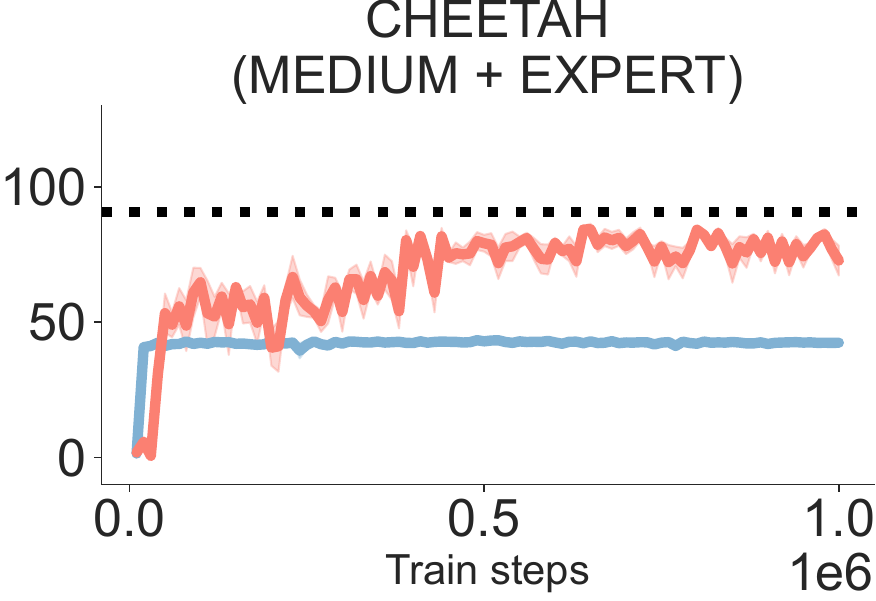}{}
    \showImage[0.225]{0}{0}{0}{0}{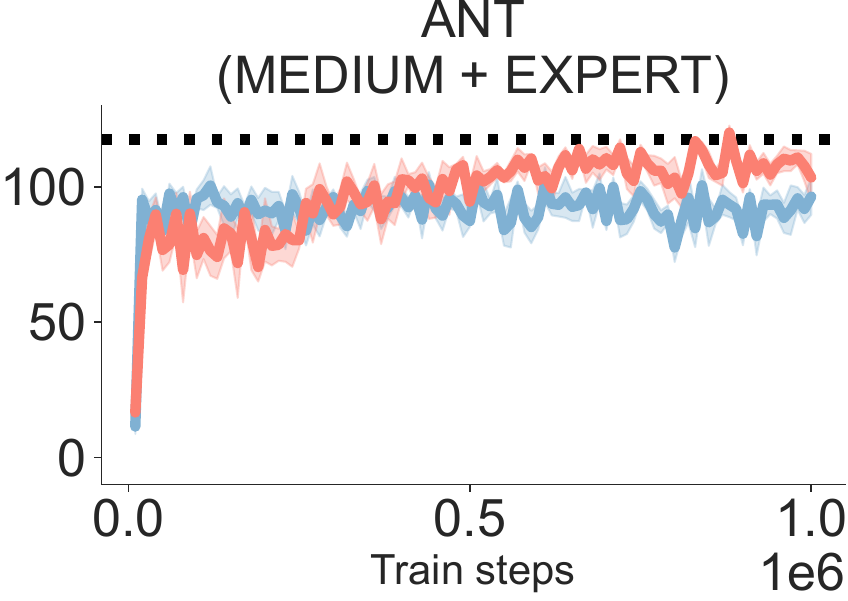}{}
    \showImage[0.225]{0}{0}{0}{0}{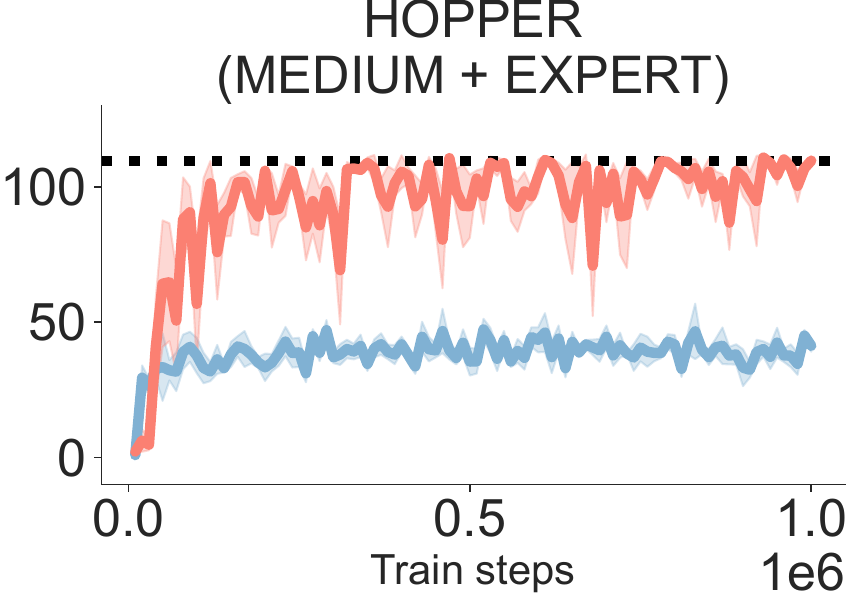}{}
    \showImage[0.225]{0}{0}{0}{0}{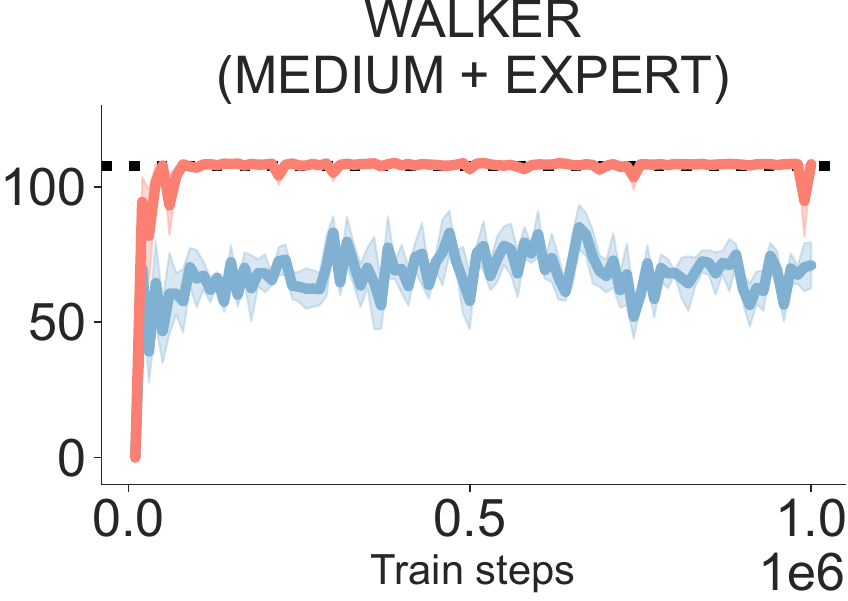}{}
    \vspace{0.5cm}
    
    \rotatebox[origin=c]{90}{\centering Score}
    \showImage[0.225]{0}{0}{0}{0}{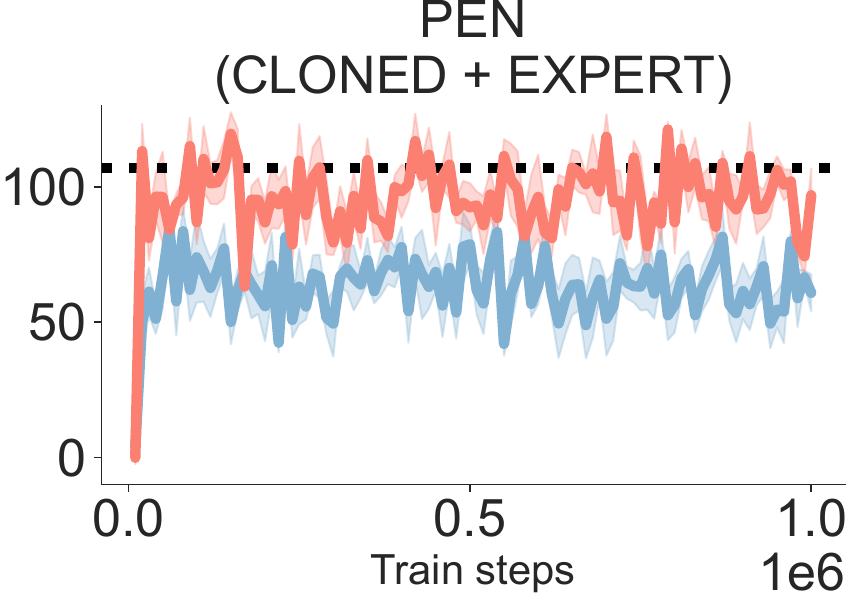}{}
    \showImage[0.225]{0}{0}{0}{0}{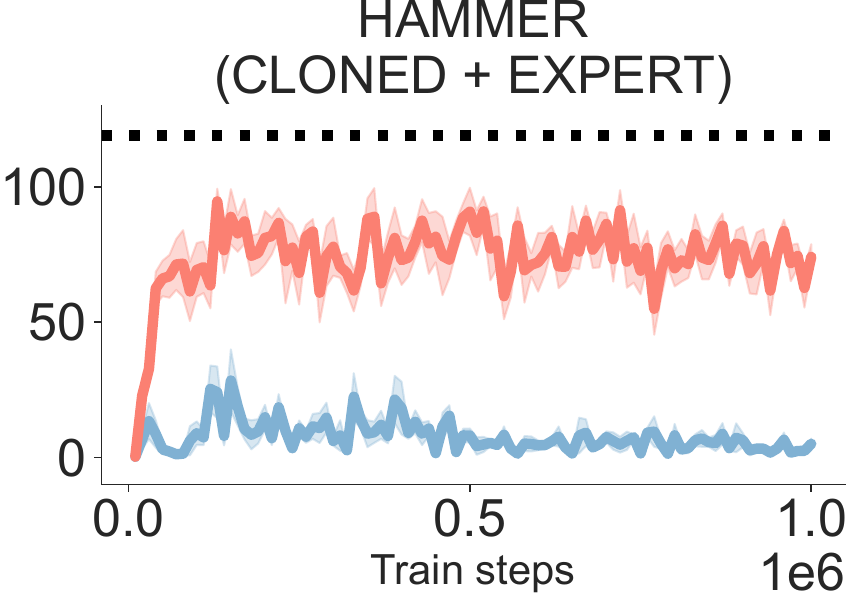}{}
    \showImage[0.225]{0}{0}{0}{0}{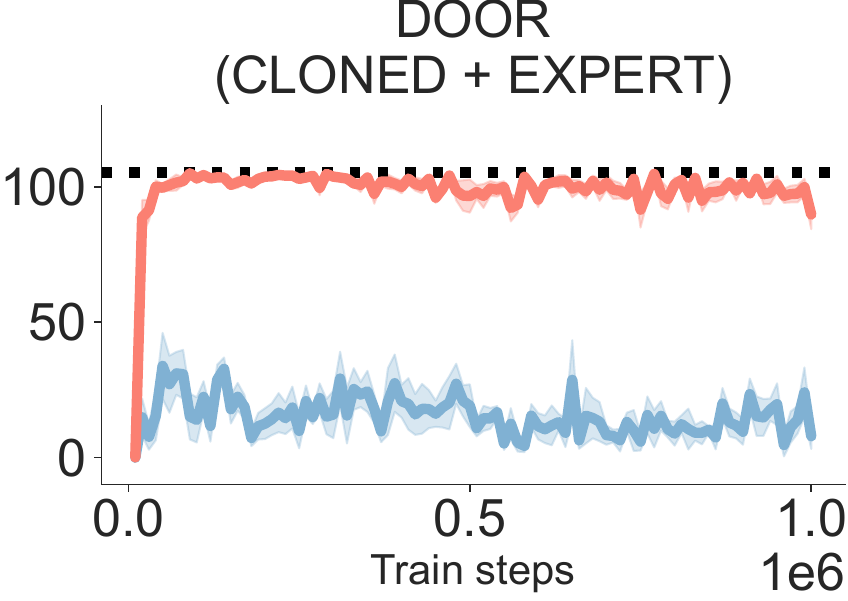}{}
    \showImage[0.225]{0}{0}{0}{0}{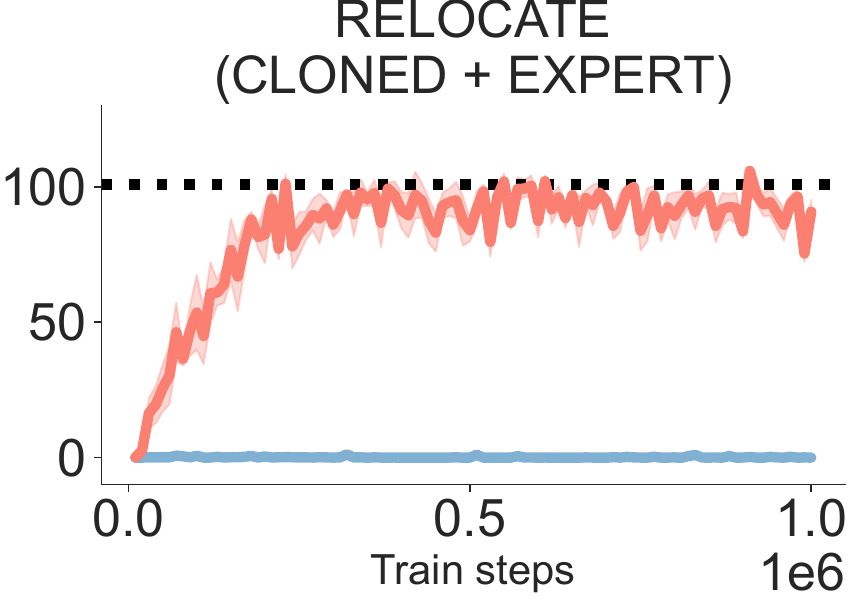}{}
    \vspace{0.5cm}
    
    \rotatebox[origin=c]{90}{\centering Score}
    \showImage[0.225]{0}{0}{0}{0}{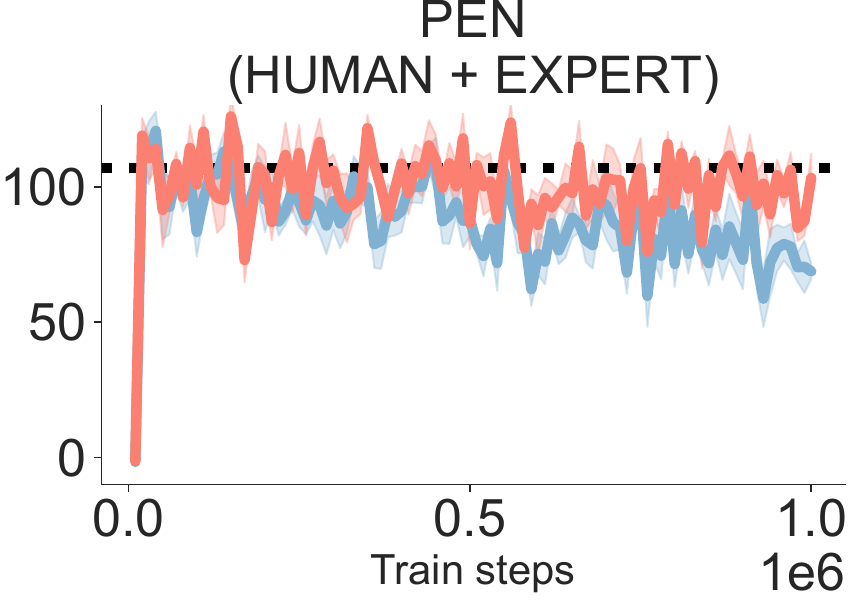}{}
    \showImage[0.225]{0}{0}{0}{0}{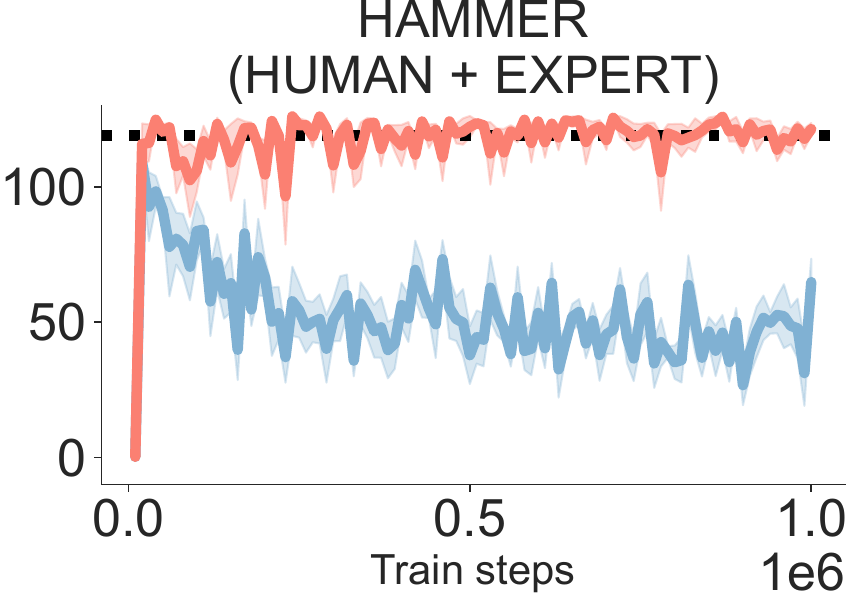}{}
    \showImage[0.225]{0}{0}{0}{0}{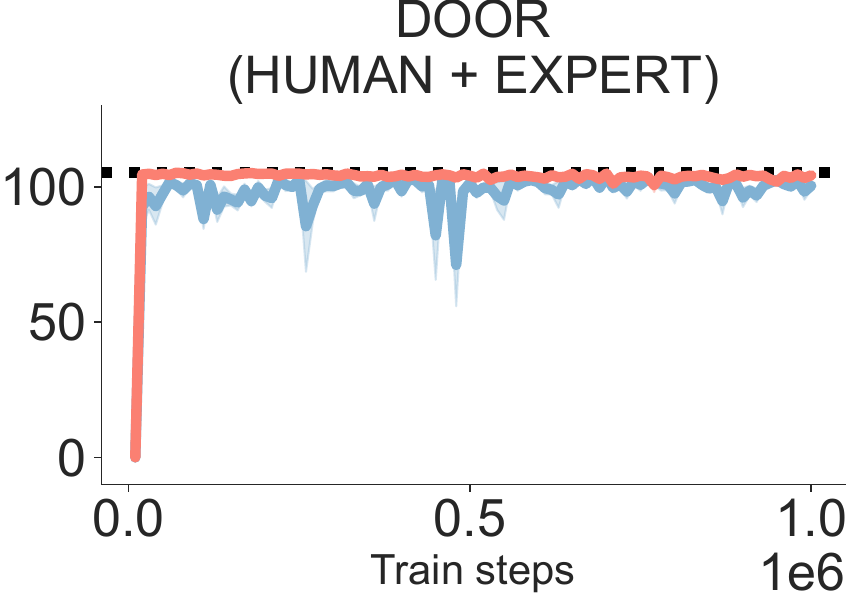}{}
    \showImage[0.225]{0}{0}{0}{0}{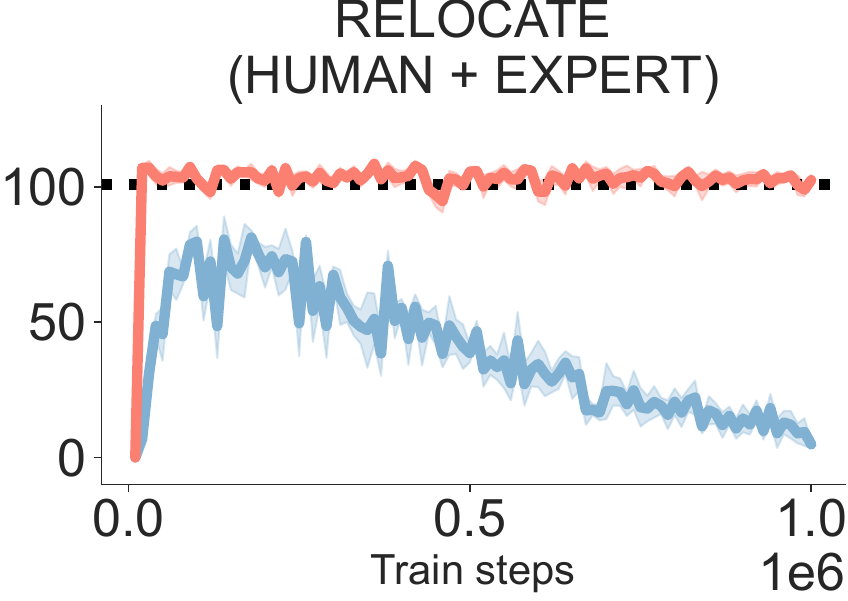}{}
    
    \caption{AW-BC and QW-BC comparison.}
    \label{fig:adv_curves}
\end{figure}

In this paper, we propose a novel policy extraction method called QW-BC (Objective~\eqref{eq:qwbc}), in contrast to prior approaches that rely on AW-BC (Objective~\eqref{eq:A-weighted-BC}). In this section, we present a comparison between QW-BC and AW-BC, as illustrated in Figure~\ref{fig:adv_curves}. Overall, QW-BC demonstrates superior policy extraction performance, attributed to its stability derived from relying on a single network estimation. In contrast, AW-BC often exhibits oscillations and instability, frequently assigning inconsistent and overly high weights to bad transitions.

\subsection{Performance Across Varying Quality Levels of the Unlabeled Dataset \texorpdfstring{$\cB^{\Mix}$}{}}
\begin{figure}[htbp]
    \centering    
    
    \rotatebox[origin=c]{90}{\centering Score}
    \showImage[0.225]{0}{0}{0}{0}{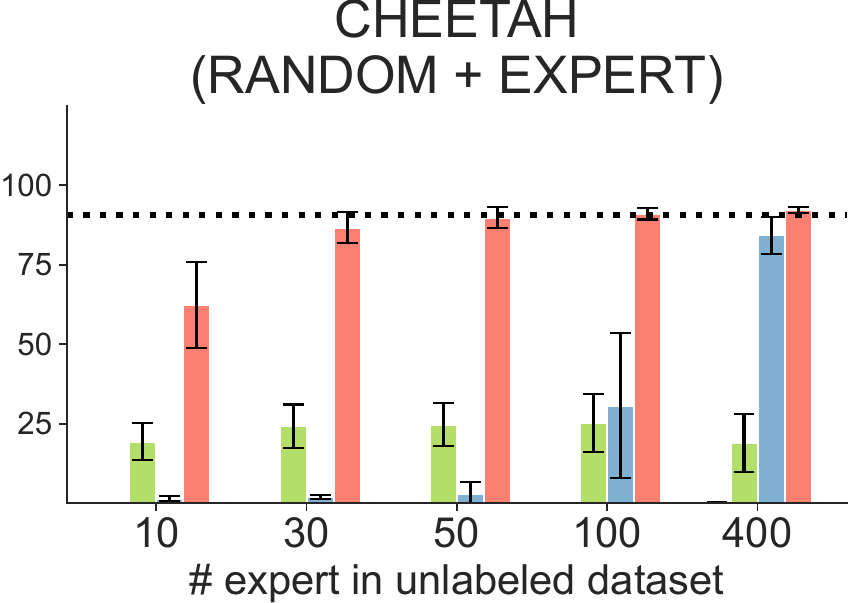}{}
    \showImage[0.225]{0}{0}{0}{0}{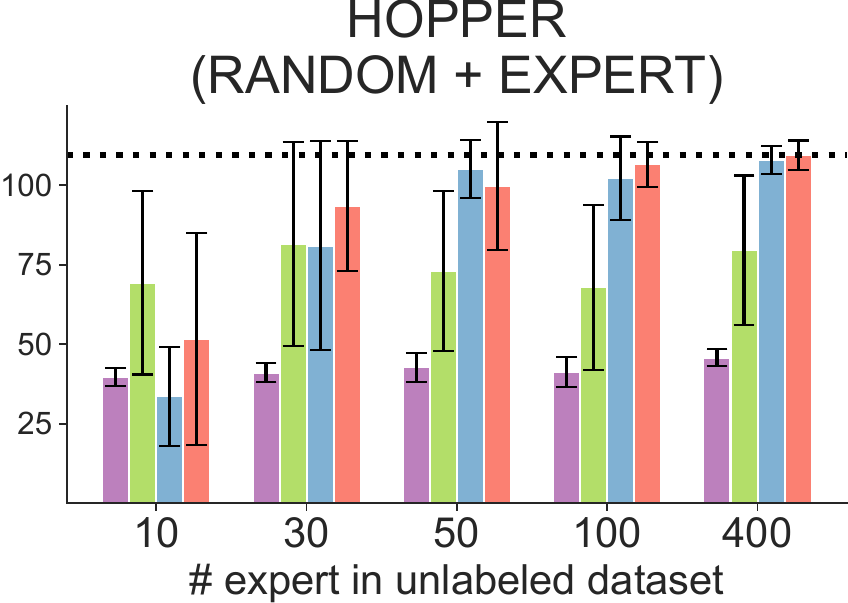}{}
    \showImage[0.225]{0}{0}{0}{0}{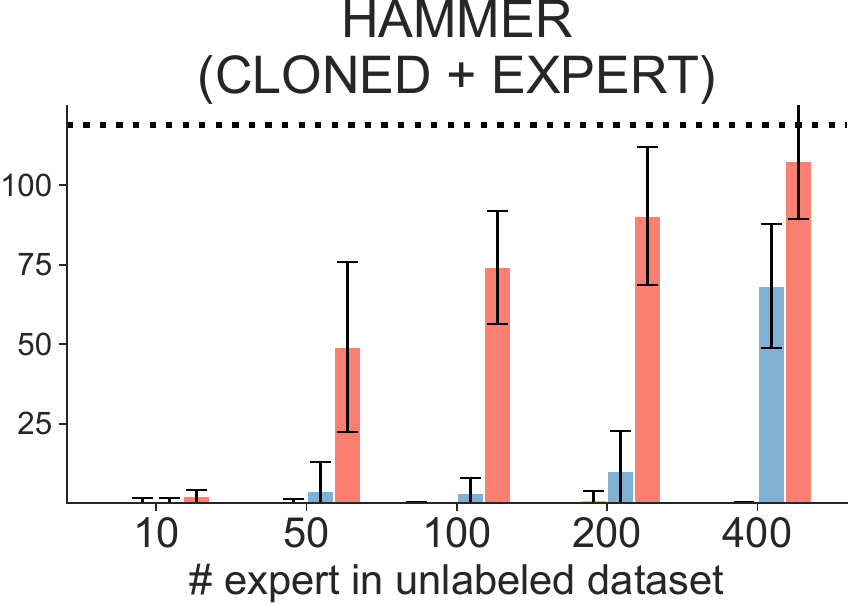}{}
    \showImage[0.225]{0}{0}{0}{0}{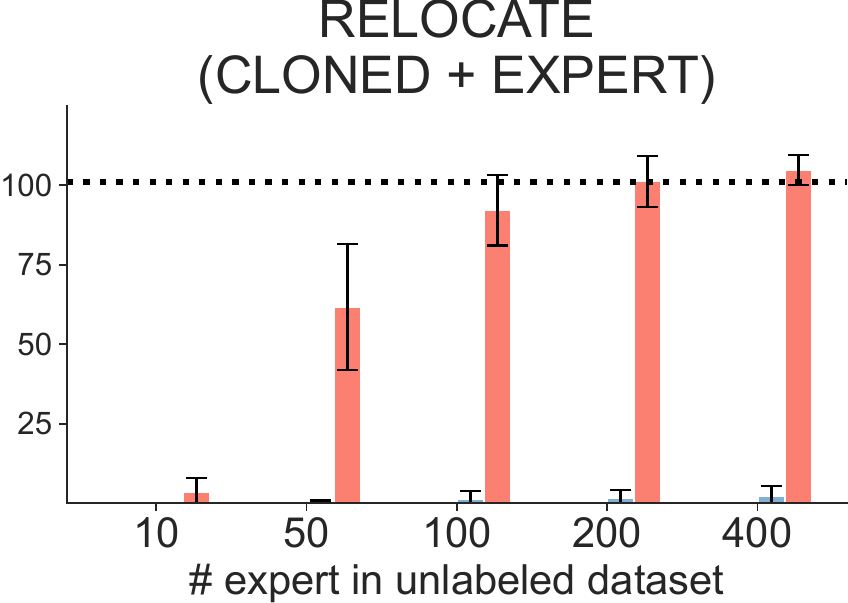}{}
    \rotatebox[origin=c]{90}{\centering Score}
    \showImage[0.225]{0}{0}{0}{0}{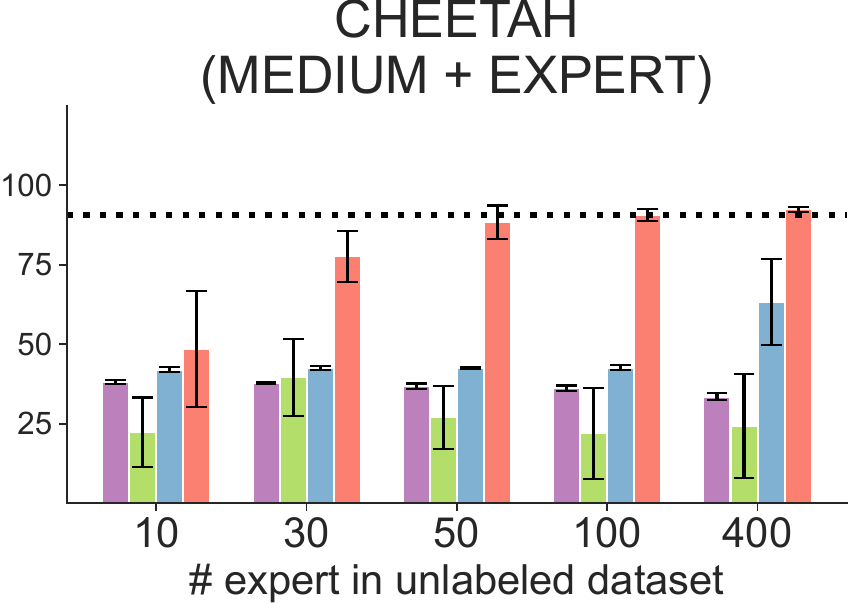}{}
    \showImage[0.225]{0}{0}{0}{0}{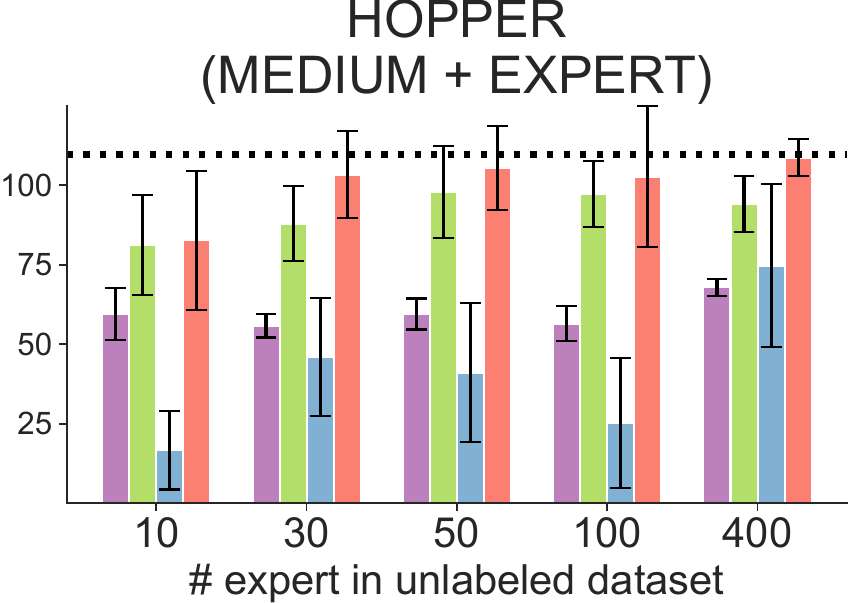}{}
    \showImage[0.225]{0}{0}{0}{0}{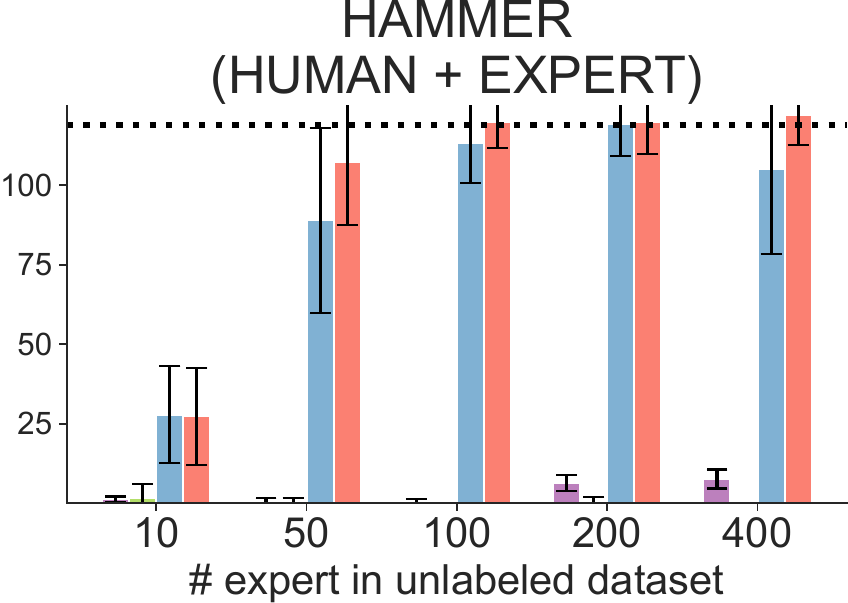}{}
    \showImage[0.225]{0}{0}{0}{0}{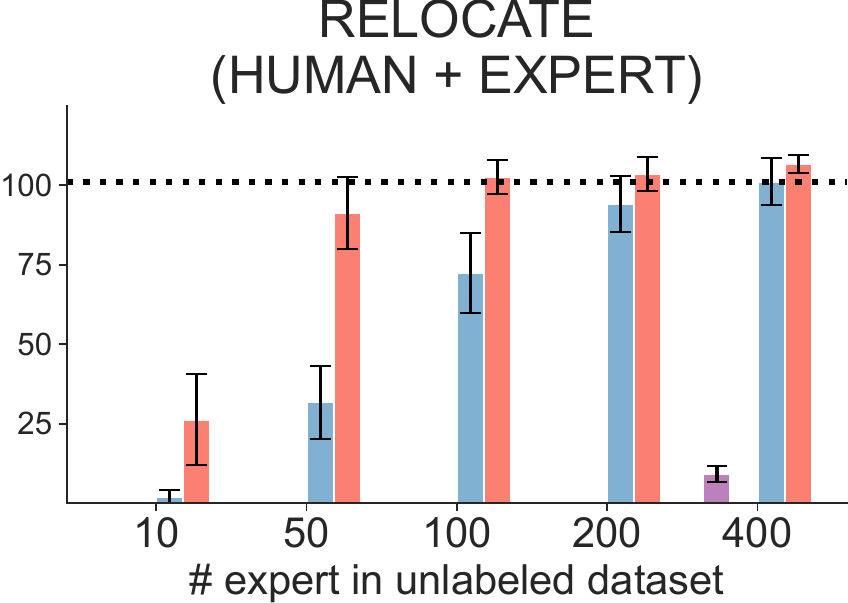}{}
    \showLegend[0.6]{0}{0}{0}{0}{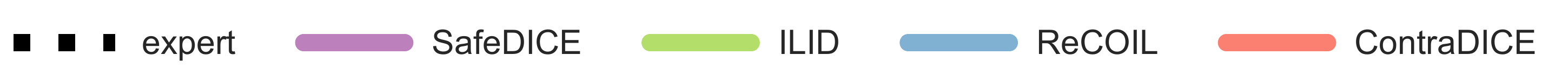}
    
    \caption{\textbf{Effect of Unlabeled Dataset Quality on Performance:} We evaluate the effect of increasing the number of expert trajectories in the unlabeled dataset $\cD^{\Mix}$. The results are calculated from 5 different training seeds, reported in normalized score. Our method outperforms SafeDICE, ILID and ReCOIL across both locomotion and manipulation tasks, achieving near-expert performance on most environments even with a small number of expert demonstrations.}
    \label{fig:unlabelled_quality}
\end{figure}

The performance of all methods is influenced by the quality of the unlabeled dataset $\cD^{\Mix}$. To evaluate the robustness of our method under varying dataset quality, we conduct experiments with different amounts of expert trajectories combined with the full set of undesirable trajectories in the unlabeled dataset. We compare our approach against ILID and ReCOIL—which leverage $\cD^G$ and $\cD^{\Mix}$—as well as SafeDICE, which learns from $\cD^B$ and $\cD^{\Mix}$. The detailed results of this study are presented in Figure~\ref{fig:unlabelled_quality}. 

In the Mujoco locomotion tasks, increasing the quality of the unlabeled dataset has minimal effect on SafeDICE and ILID, and both methods continue to underperform on the Adroit hand manipulation tasks regardless of the number of expert trajectories included. In contrast, ReCOIL shows improved performance as the quality of the unlabeled dataset increases, successfully learning 4 out of 8 tasks across both locomotion and manipulation domains. Overall, our method achieves near-expert performance on 7 out of 8 tasks while requiring significantly lower-quality unlabeled datasets $\cD^{\Mix}$, demonstrating its superior data efficiency and robustness.

\subsection{Adaptations and Experiments with  \texorpdfstring{$\alpha>1$}{}}
\label{sec:large_alpha_ablation}
From our objective function~\eqref{eq:main-obj}, we introduce a hyperparameter $0 \leq \alpha < 1$, which controls the weighting of the bad data objective—this corresponds to question \textbf{(Q3)}. To evaluate the sensitivity of our method to $\alpha$, we conduct experiments by varying its value and observing its impact on final performance. Specifically, we perform a full sweep over $\alpha \in \{0, 0.1, 0.2, \ldots, 0.9\}$ to illustrate how this key hyperparameter influences learning outcomes.

Interestingly, we observe that in some cases, settings with $\alpha \geq 1$ yield favorable performance, suggesting that avoiding bad data may, at times, be more critical than imitating good data. However, directly applying $\alpha \geq 1$ in our original formulation violates convexity conditions.

\begin{figure}[htbp]
    \centering    
    \showImage[0.8]{0}{0}{0}{0}{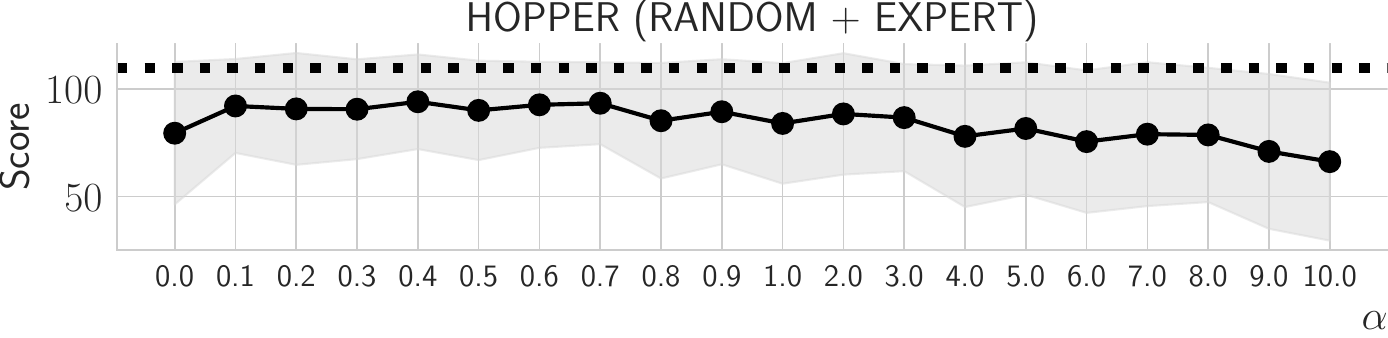}{}
    \showImage[0.8]{0}{0}{0}{0}{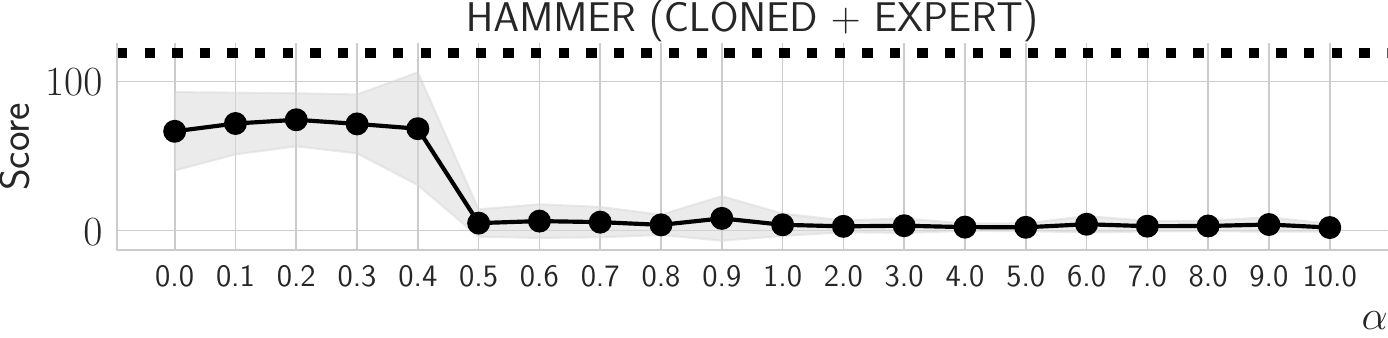}{}
    \showImage[0.8]{0}{0}{0}{0}{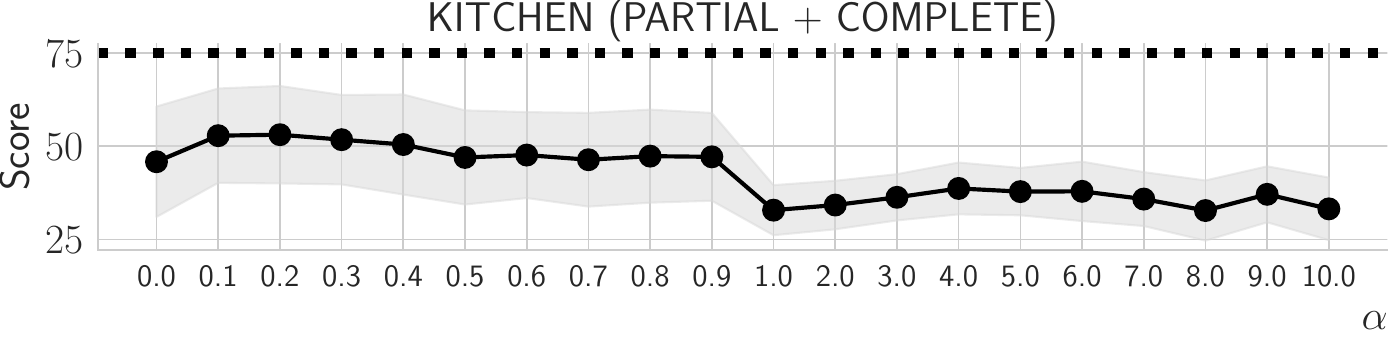}{}
    
    \caption{Performance of large $\alpha\geq1$.}
    \label{fig:large_alpha_ablation_study}
\end{figure}

To address this, we propose a naive modification of Objective~\eqref{eq:L(Q|V)} that accommodates $\alpha \geq 1$ while preserving practical applicability. The revised objective is defined as:
\begin{align}
\widetilde{L}(Q \mid V) &=
    (1 - \gamma) \, \mathbb{E}_{s \sim p_0} \left[ V(s) \right]
    - \mathbb{E}_{(s, a) \sim d^U} \left[\exp\left( \Psi(s, a)\right) \left(Q(s,a) - \gamma \mathbb{E}_{s'}[V(s')]\right) \right],
    \label{eq:large_alpha_L(Q|V)}
\end{align}
which enables empirical investigation into the high-$\alpha$ regime while sidestepping theoretical limitations. The experiment results are provided in Figure~\ref{fig:large_alpha_ablation_study}. Overall, $\alpha \geq 1$ does not provide good performance, which raises the limitation of the naive adaptation.

\newpage
\subsection{Comparison Between \texorpdfstring{$L(Q, \pi)$}{} and the Surrogate \texorpdfstring{$\widetilde{L}(Q, \pi)$}{}}
As shown in Proposition~\ref{prop:4.3}, the original objective \( L(Q \mid V) \) (Equation~\eqref{eq:main-obj-Q-pi}) is transformed into a modified version \( \widetilde{L}(Q \mid V) \) (Equation~\eqref{eq:L(Q|V)}). This experiment investigates the performance differences between the two objectives.

To improve the stability of the original objective \( L(Q \mid V) \), we need to address the issue of exponential terms producing extremely large values, which can lead to numerical instability. A practical approach is to clip the input to the exponential function to a bounded range \([ \text{minR}, \text{maxR} ]\), resulting in the following formulation:

\begin{align}
    L(Q, \pi) = &(1 - \gamma) \, \mathbb{E}_{s \sim p_0} \left[ V_Q^\pi(s) \right] \nonumber\\
&+ (1-\alpha)\mathbb{E}_{(s, a) \sim d^U} \left[ \exp\left(\left( \frac{\Psi(s, a) - \mathcal{T}^\pi[Q](s, a)}{1 - \alpha} \right)\textcolor{red}{.\text{clip}(\text{minR}, \text{maxR})}\right) \right]   ,
\end{align}
where \(\text{minR} = -7\) and \(\text{maxR} = 7\) in our experiments.

\begin{figure}[htbp]
    \centering
    \rotatebox[origin=c]{90}{\centering Score}
    \showImage[0.225]{0}{0}{0}{0}{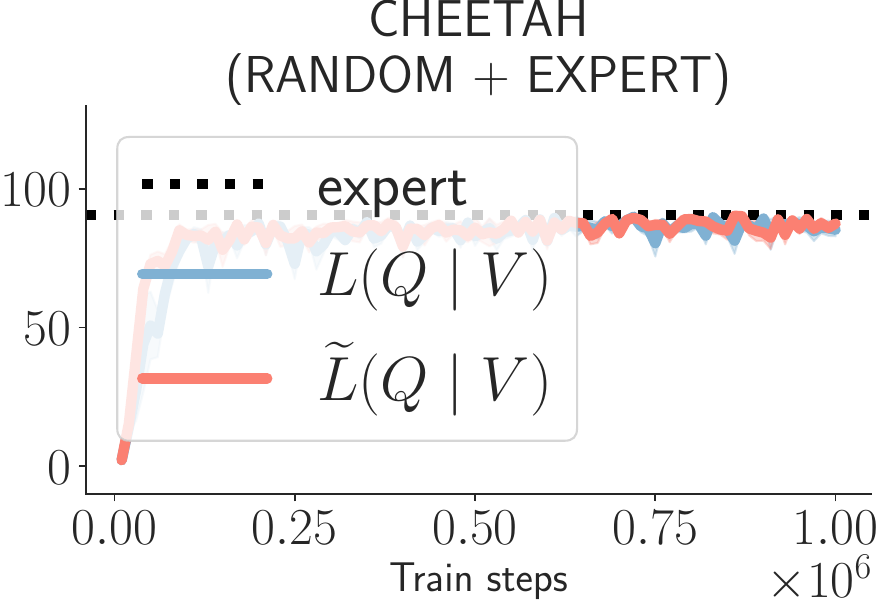}{}
    \showImage[0.225]{0}{0}{0}{0}{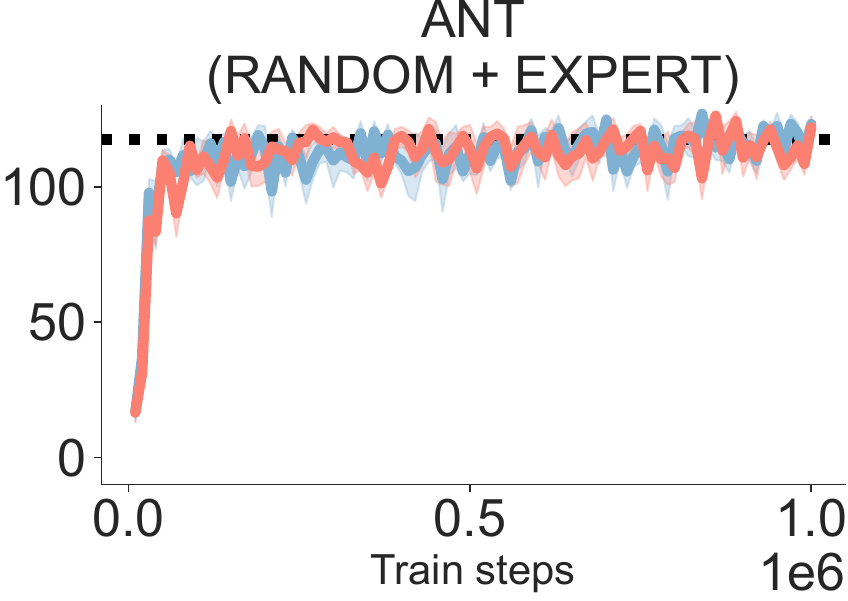}{}
    \showImage[0.225]{0}{0}{0}{0}{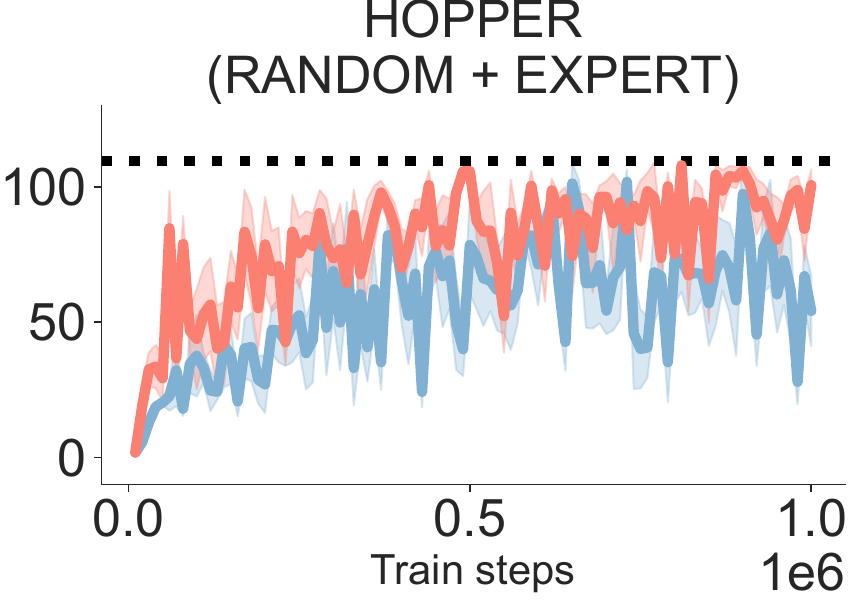}{}
    \showImage[0.225]{0}{0}{0}{0}{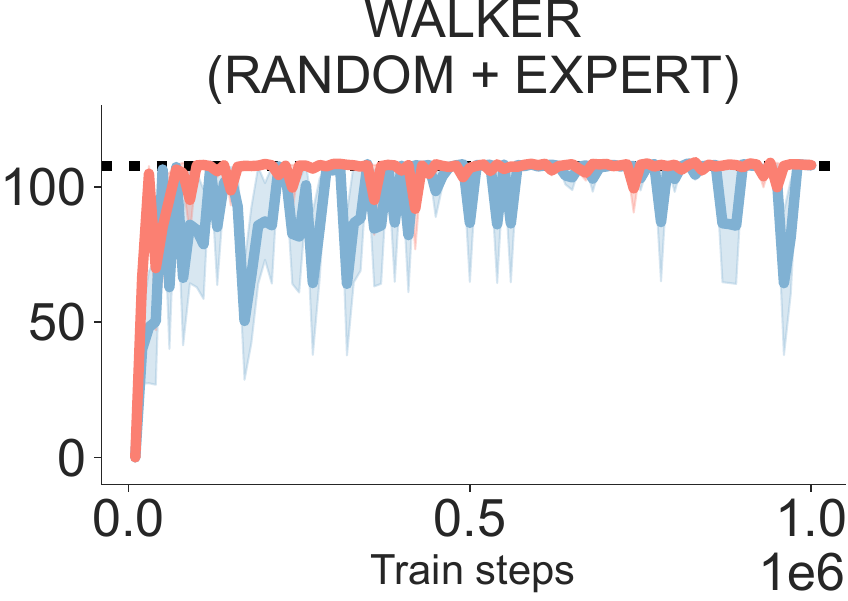}{}
    \vspace{0.5cm}
    
    \rotatebox[origin=c]{90}{\centering Score}
    \showImage[0.225]{0}{0}{0}{0}{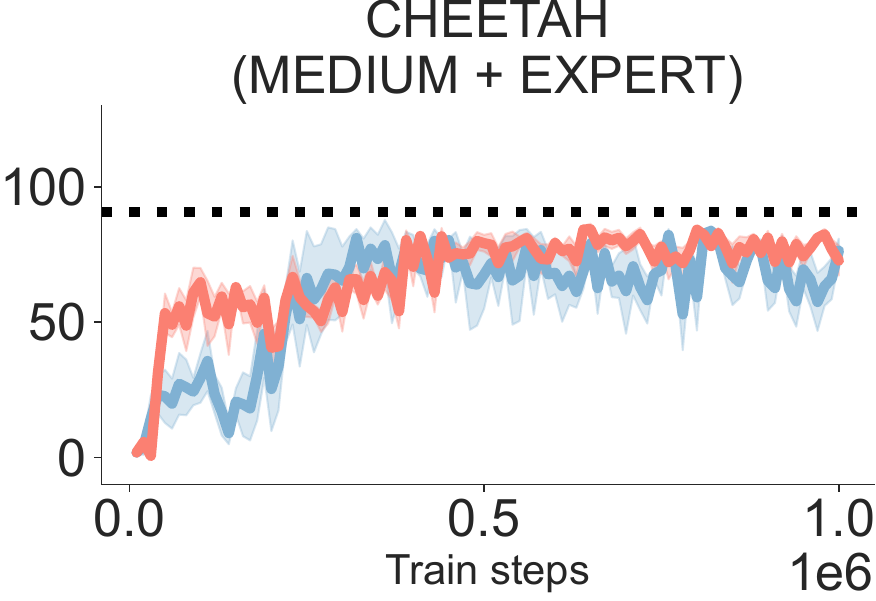}{}
    \showImage[0.225]{0}{0}{0}{0}{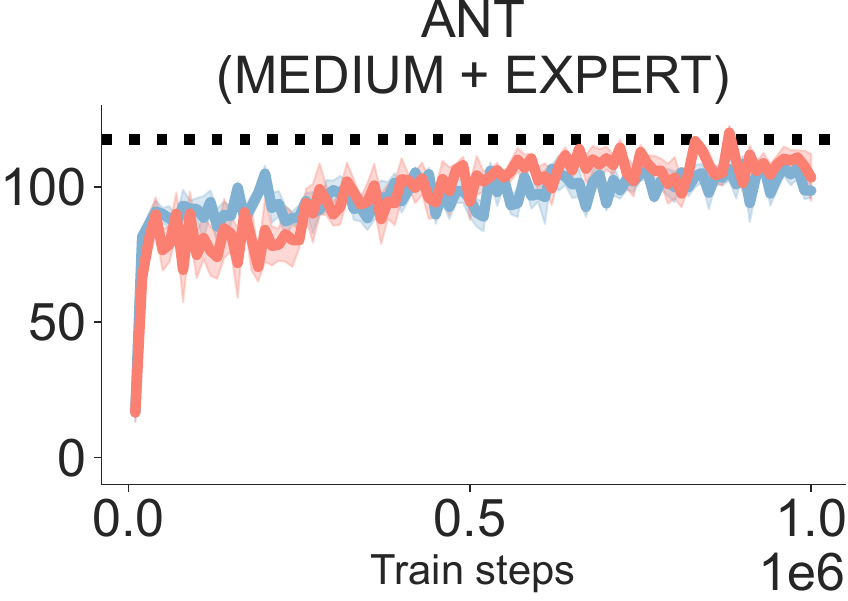}{}
    \showImage[0.225]{0}{0}{0}{0}{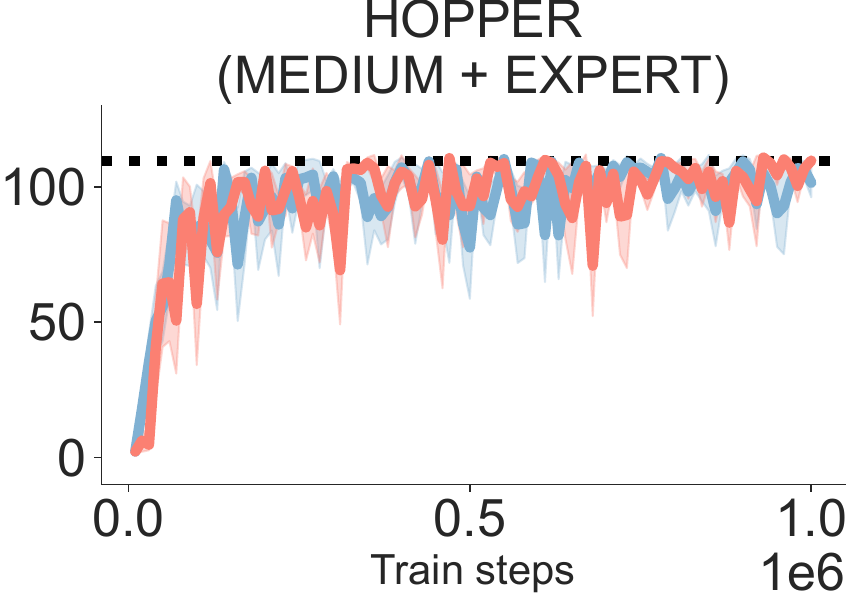}{}
    \showImage[0.225]{0}{0}{0}{0}{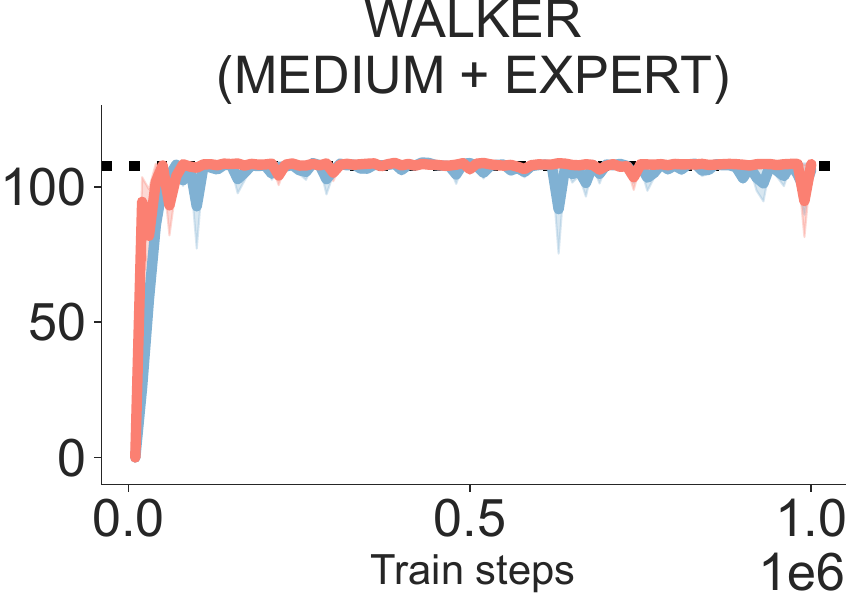}{}
    \vspace{0.5cm}
    
    \rotatebox[origin=c]{90}{\centering Score}
    \showImage[0.225]{0}{0}{0}{0}{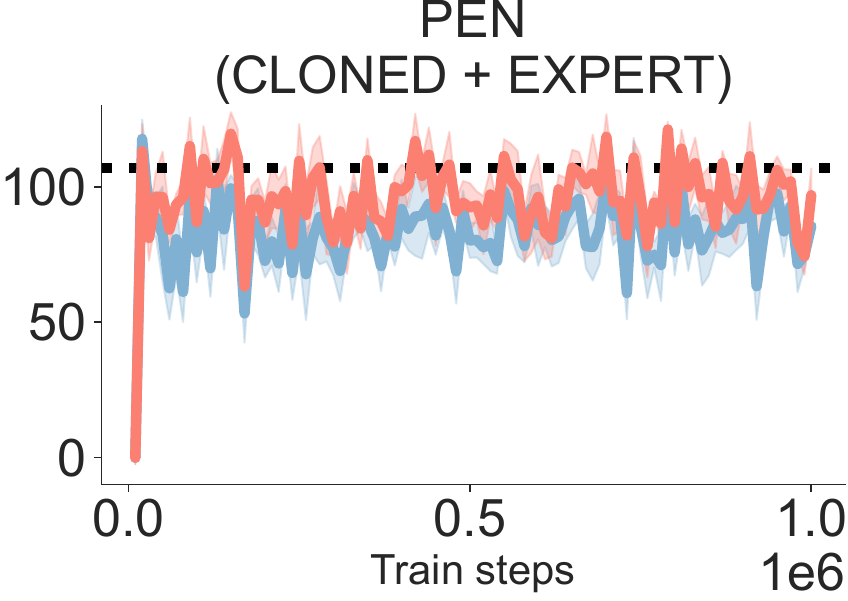}{}
    \showImage[0.225]{0}{0}{0}{0}{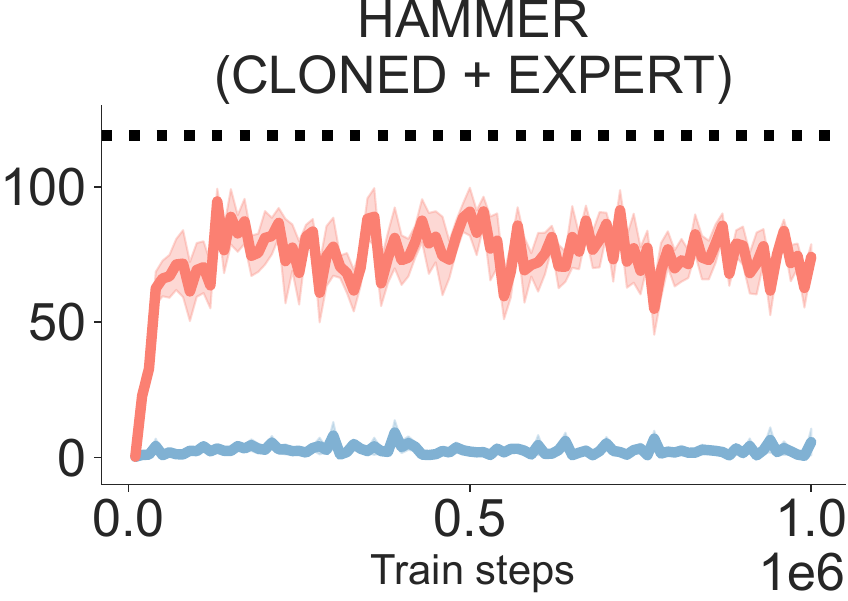}{}
    \showImage[0.225]{0}{0}{0}{0}{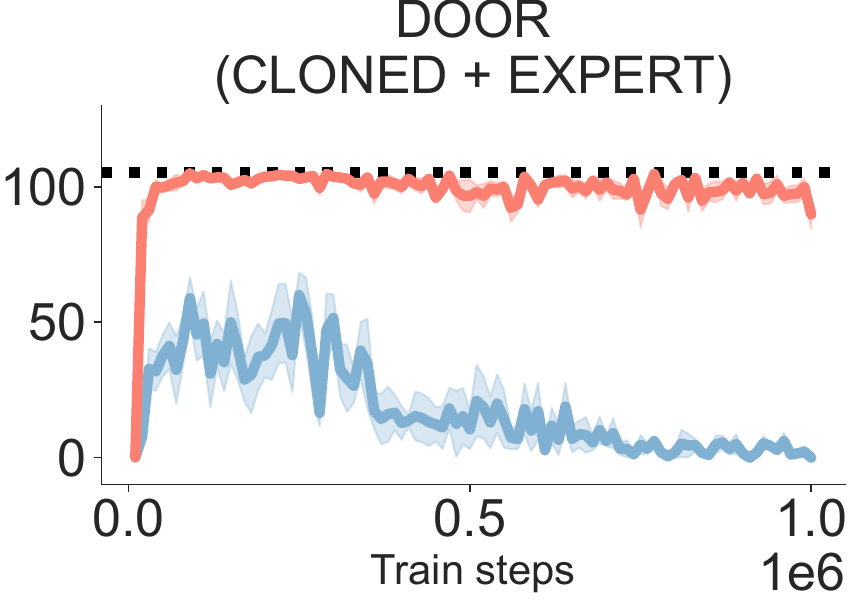}{}
    \showImage[0.225]{0}{0}{0}{0}{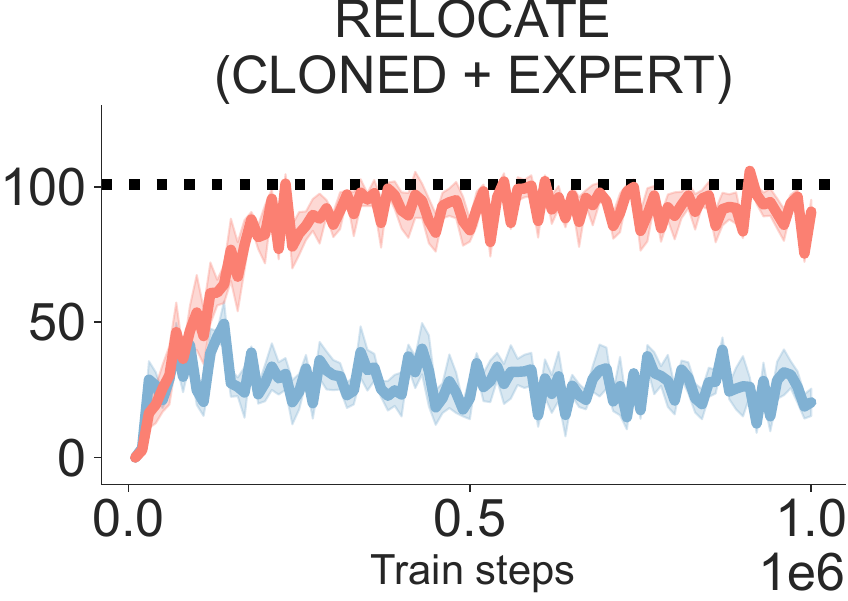}{}
    \vspace{0.5cm}

    \rotatebox[origin=c]{90}{\centering Score}
    \showImage[0.225]{0}{0}{0}{0}{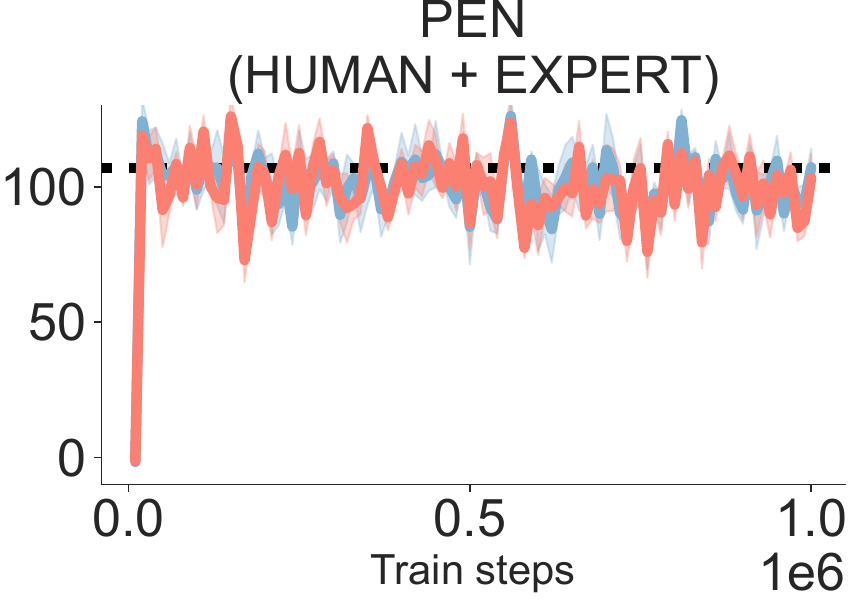}{}
    \showImage[0.225]{0}{0}{0}{0}{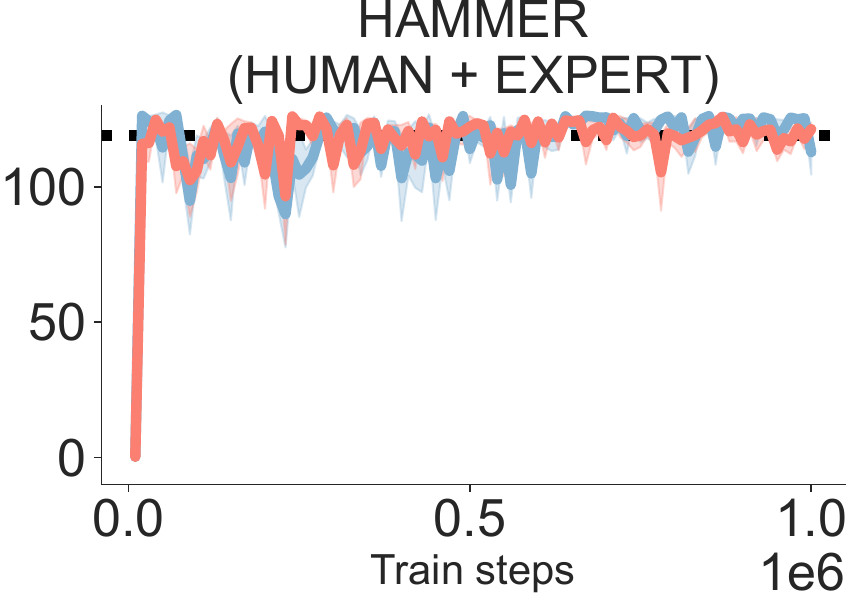}{}
    \showImage[0.225]{0}{0}{0}{0}{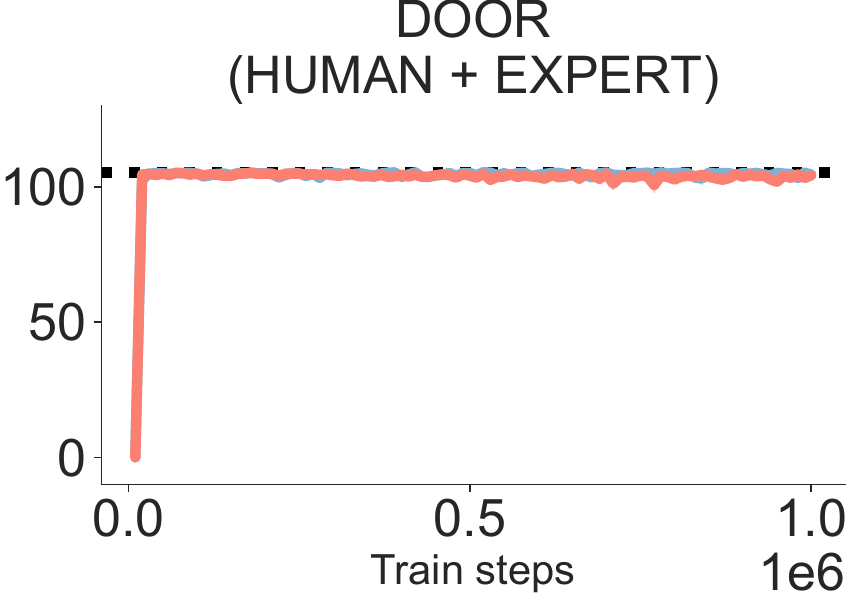}{}
    \showImage[0.225]{0}{0}{0}{0}{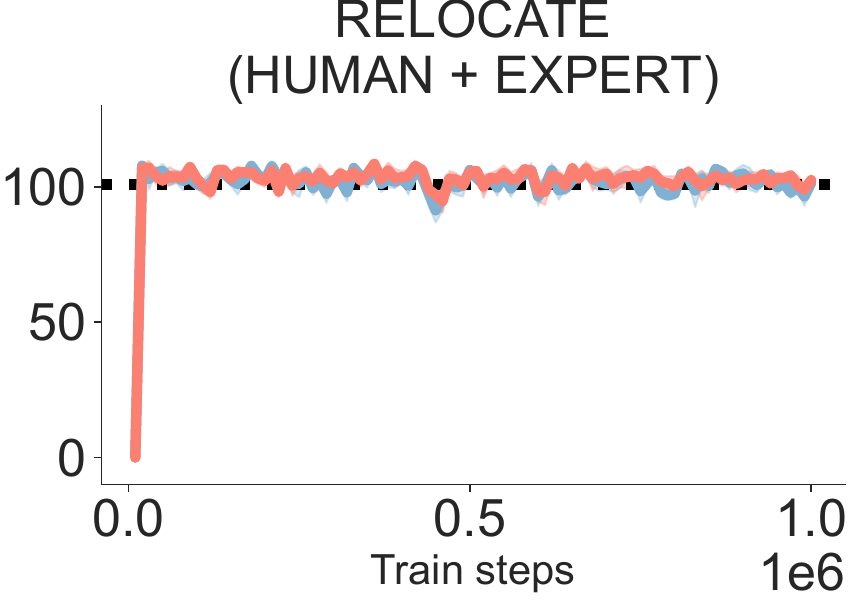}{}
    
    \caption{Exponetial ablation study.}
    \label{fig:exp_ablation}
\end{figure}

The results of this ablation study are presented in Figure~\ref{fig:exp_ablation}, illustrating the performance impact of this stability-enhancing modification. In general, the clipping technique effectively mitigates the instability caused by the exponential term, successfully preventing $NaN$ errors during training. However, this modification also leads to a drop in performance and, in some tasks, causes the method to fail to learn effectively.

\newpage
\subsection{Sensitivity Analysis of  \texorpdfstring{$\beta$}{}}
\label{sec:beta_ablation}

In this section, we explore how different values of the $\beta$ parameter affect performance. The experiment results are provided in Table~\ref{tab:beta_ablation_study}. The results show that while $\beta$ significantly influences outcomes, performance remains consistent over a wide range of $\beta$ values, implying that minimal tuning effort is needed for this hyperparameter.

\begin{table}[htbp]
\vskip 0.15in
\begin{center}
\begin{small}
\resizebox{\textwidth}{!}{
\begin{tabular}{lllllllll}
\toprule
\multirow{2}{*}{Task}&\multirow{2}{*}{unlabeled $\cD^{\Mix}$}&&&\multicolumn{3}{c}{\textbf{$\beta$ value}}\\
\cmidrule(lr){3-9}
& &  1& 3& 5& 10& 15& 20&30\\
\midrule
\multirow{2}{*}{\textsc{cheetah}}&\textsc{random}+\textsc{expert}&  $2.25_{\pm0.0}$& $2.25_{\pm0.0}$& $2.25_{\pm0.0}$& $2.24_{\pm0.0}$& $83.2_{\pm5.3}$& \highlight{85.8$_{\pm2.1}$}&$84.3_{\pm1.4}$\\
 & \textsc{medium}+\textsc{expert}& $42.4_{\pm0.2}$&  $42.9_{\pm0.3}$& $53.9_{\pm8.8}$& \highlight{83.1$_{\pm4.9}$}& $80.1_{\pm2.6}$& $78.7_{\pm2.3}$&$76.7_{\pm5.2}$\\
\midrule
 \multirow{2}{*}{\textsc{ant}}& \textsc{random}+\textsc{expert}& $39.5_{\pm7.3}$&  $69.3_{\pm6.5}$& $60.9_{\pm28.7}$& $115.6_{\pm4.6}$& \highlight{118.0$_{\pm2.1}$}& $114.5_{\pm1.7}$&$116.0_{\pm2.1}$\\
 & \textsc{medium}+\textsc{expert}& $91.0_{\pm1.1}$&  $90.6_{\pm1.7}$& $93.7_{\pm1.5}$& $104.8_{\pm3.9}$& \highlight{106.5$_{\pm2.4}$}& $101.1_{\pm3.3}$&$95.1_{\pm1.3}$\\
\midrule
 \multirow{2}{*}{\textsc{hopper}}& \textsc{random}+\textsc{expert}& $4.7_{\pm0.4}$&  $5.2_{\pm0.9}$& $7.2_{\pm1.3}$& $7.9_{\pm1.9}$& $20.4_{\pm9.7}$& $67.4_{\pm7.9}$&\highlight{94.4$_{\pm6.3}$}\\
 & \textsc{medium}+\textsc{expert}& $52.1_{\pm1.5}$&  $46.0_{\pm1.0}$& $85.8_{\pm11.6}$& $96.3_{\pm8.1}$& $96.9_{\pm12.5}$& \highlight{99.6$_{\pm4.1}$}&$98.0_{\pm5.7}$\\
\midrule
 \multirow{2}{*}{\textsc{walker}}& \textsc{random}+\textsc{expert}& $2.9_{\pm2.6}$&  $3.5_{\pm2.9}$& $6.4_{\pm4.6}$& $32.5_{\pm27.7}$& $105.7_{\pm4.5}$& $106.2_{\pm2.0}$&\highlight{107.5$_{\pm1.1}$}\\
 & \textsc{medium}+\textsc{expert}& $68.3_{\pm3.7}$&  $65.8_{\pm3.2}$& $53.4_{\pm3.6}$& $104.9_{\pm2.5}$& $108.1_{\pm0.1}$& \highlight{108.2$_{\pm0.2}$}&\highlight{108.2$_{\pm0.1}$}\\
\bottomrule
\end{tabular}
}
\end{small}
\end{center}
\caption{Performance of ContraDICE in different $\beta$ value in MuJoCo locomotion tasks.}
\label{tab:beta_ablation_study}
\vskip -0.1in
\end{table}

\end{document}